\def\year{2021}\relax
\documentclass[letterpaper]{article} % DO NOT CHANGE THIS

\usepackage{aaai21}  % DO NOT CHANGE THIS

\usepackage{times}  % DO NOT CHANGE THIS
\usepackage{helvet} % DO NOT CHANGE THIS
\usepackage{courier}  % DO NOT CHANGE THIS
\usepackage[hyphens]{url}  % DO NOT CHANGE THIS
\usepackage{graphicx} % DO NOT CHANGE THIS
\usepackage{natbib}  % DO NOT CHANGE THIS AND DO NOT ADD ANY OPTIONS TO IT
\usepackage{caption} % DO NOT CHANGE THIS AND DO NOT ADD ANY OPTIONS TO IT
\usepackage{subfigure}
\usepackage{textcomp,booktabs,threeparttable}
\usepackage{amsfonts}

\usepackage{amsmath}

\newcommand{\RR}[2]{\mathbb{R}^{#1 \times #2}}

\newcommand{\rank}[1]{\mbox{rank}\left(#1\right)}

\newcommand{\ST}{\mathcal{S}}
\newcommand{\SVT}{\mathcal{D}}

\newcommand{\st}{\mbox{s.t.~}}

\newcommand{\refEq}[1]{Eq.(\ref{#1})}

\newcommand{\refFig}[1]{Figure~\ref{#1}}

\def\bfu{{\mathbf{u}}}
\def\bfv{{\mathbf{v}}}

\def\bfD{{\mathbf{D}}}
\def\bfE{{\mathbf{E}}}

\def\bfM{{\mathbf{M}}}

\def\bfX{{\mathbf{X}}}

\def\half{\frac{1}{2}~}

\title{Exploring Common and Individual Characteristics of Students via Matrix Recovering}
\author {
	Zhen Wang,\textsuperscript{\rm 1,2}
	Ben Teng, \textsuperscript{\rm 1,3}
	Yun Zhou \textsuperscript{\rm 1,3}
	Hanshuang Tong, \textsuperscript{\rm 1,3}
	Guangtong Liu \textsuperscript{\rm 1,3}
	 \\
}
\affiliations {
	\textsuperscript{\rm 1} Aixuexi Education Group \\
	\textsuperscript{\rm 2} wangzhenYSU@163.com\\ \textsuperscript{\rm 3} \{tengben0905,zhouyun.nudt,tonghanshuang.thu, feidieliuliuliu\}@gmail.com \\
}

\begin{document}
%% remove the box 
\maketitle

\begin{abstract}
Balancing group teaching and individual mentoring is an important issue in education area. The nature behind this issue is to explore common characteristics shared by multiple students and individual characteristics for each student. Biclustering methods have been proved successful for detecting meaningful patterns with the goal of driving group instructions based on students' characteristics. However, these methods ignore the individual characteristics of students as they only focus on common characteristics of students. In this article, we propose a framework to detect both group characteristics and individual characteristics of students simultaneously. We assume that the characteristics matrix of students' is composed of two parts: one is a low-rank matrix representing the common characteristics of students; the other is a sparse matrix representing individual characteristics of students. Thus, we treat the balancing issue as a matrix recovering problem. The experiment results show the effectiveness of our method. Firstly, it can detect meaningful biclusters that are comparable with the state-of-the-art biclutering algorithms. Secondly, it can identify individual characteristics for each student simultaneously. Both the source code of our algorithm and the real datasets are available upon request.

\end{abstract}

\section{Introduction}
A growing collection of educational data contributes to the research of modeling student characteristics. For instance, by exploiting the exercising records of students with knowledge tracing or knowledge diagnosis methods, researchers can track the change of each student’s knowledge acquisition during their exercising activities, and output a student-knowledge mastery matrix, whose element presents the performance level of knowledge mastery for each student \cite{NIPS2015_5654,liu2019ekt,tong2020exercise}. 

Based on the characteristics of students, instructors can offer each student specific interventions to improve their performance \cite{Lin2016Instructional}. However, it is too time-consuming for instructors to complete these tasks by hand, since a large number of students  usually vary greatly in learning rates and knowledge levels. One feasible approach is conducting a clustering anaylsis on the student-knowledge mastery matrix to solve this challenge, which is the practice of placing students of similar characteristics in the same group \cite{m.a2015clustering,DBLP:journals/tsmc/RomeroV10}. After discovering student groups according to students' common characteristics, personalized learning systems can be built and adaptive contents can be offered to promote effective group learning by instructors.

% instructors can build a personalized learning system to promote effective group learning and provide adaptive contents to the groups.

To date, serveral studies have been published to cluster students into meaningful groups based on their characteristics with a goal of driving group instructions \cite{amershi2009combining,dutt2017a,DBLP:conf/its/MojaradEMB18,DBLP:conf/its/HenriquesFC19}. These methods can be mainly divided into two major categories: traditional clustering methods and biclustering methods. Traditional clustering methods, such as hierarchical agglomerative clustering, K-means and model-based clustering, identify groups of students with similar characteristics in a global way, that is they simply group students according to all available values (all knowledge mastery levels in this article), thus being unable to identify local patterns. While biclustering algorithms, whose particularity is that partitioning is done in two dimensions of a matrix yielding to clustering according to students and their characteristics, allow the discovery of local patterns. \citep{DBLP:conf/its/MojaradEMB18} is one of typical algorithms based on traditional clustering methods to group students. Mean-shift clustering is used to select a number of clusters, and then k-mean clustering is applied to identify distinct student proﬁles. \cite{DBLP:conf/its/HenriquesFC19} is one recent application of biclustering method in educational data. The authors use pattern-based biclustering approach to detect non-trivial, yet potentially relevant educational and statistically significant patterns from the performance of students' data. Their results confirm the unique role of biclustering in finding relevant patterns of students' performance.

Despite the advances of researches on student grouping, one important problem that students maybe do not have the same characteristics in some aspects even though they're in the same group is always ignored. Balancing group teaching and individual mentoring is not trivial in the education area. To our knowledge, there has been no existing computional method to solve this issue. In this article, we propose a novel method to solve this problem. Specifically, we use the method to identify both group characteristics and individual characteristics of students simultaneously in this article. Under the assumption that common characteristics shared by multiple students form a low-rank matrix and individual characteristics for each student form a sparse matrix, we model the problem as a matrix recovering problem. Then we use an iterative algorithm to solve it.

To demonstrate the performance of the proposed method, we conduct comparison experiments using both synthesized datasets and real educational datasets. Simulation results show that the proposed method achieves comparable performances compared with existing methods in many settings. And experiment results on two real datasets demonstrate the effectiveness of our method.

Overall, the salient features of the method described in this article can be summarized as follows: 
\begin{itemize}
	\item it inherits the advantage of biclustering that local pattern can be discovered.
	\item the individual characteristics for each student can be detected meanwhile. 
	\item statistical evaluation is employed to filter the spurious biclusters, which guarantees the statistical significant of the results.
\end{itemize}

\section{Method}
In this paper, we propose a new framework for exploring common and individual characteristerics of students.
%, which is based on matrix recovering and statistical validation. 
\begin{figure}[h]
	\centering
	\includegraphics[width=1.1 \linewidth]{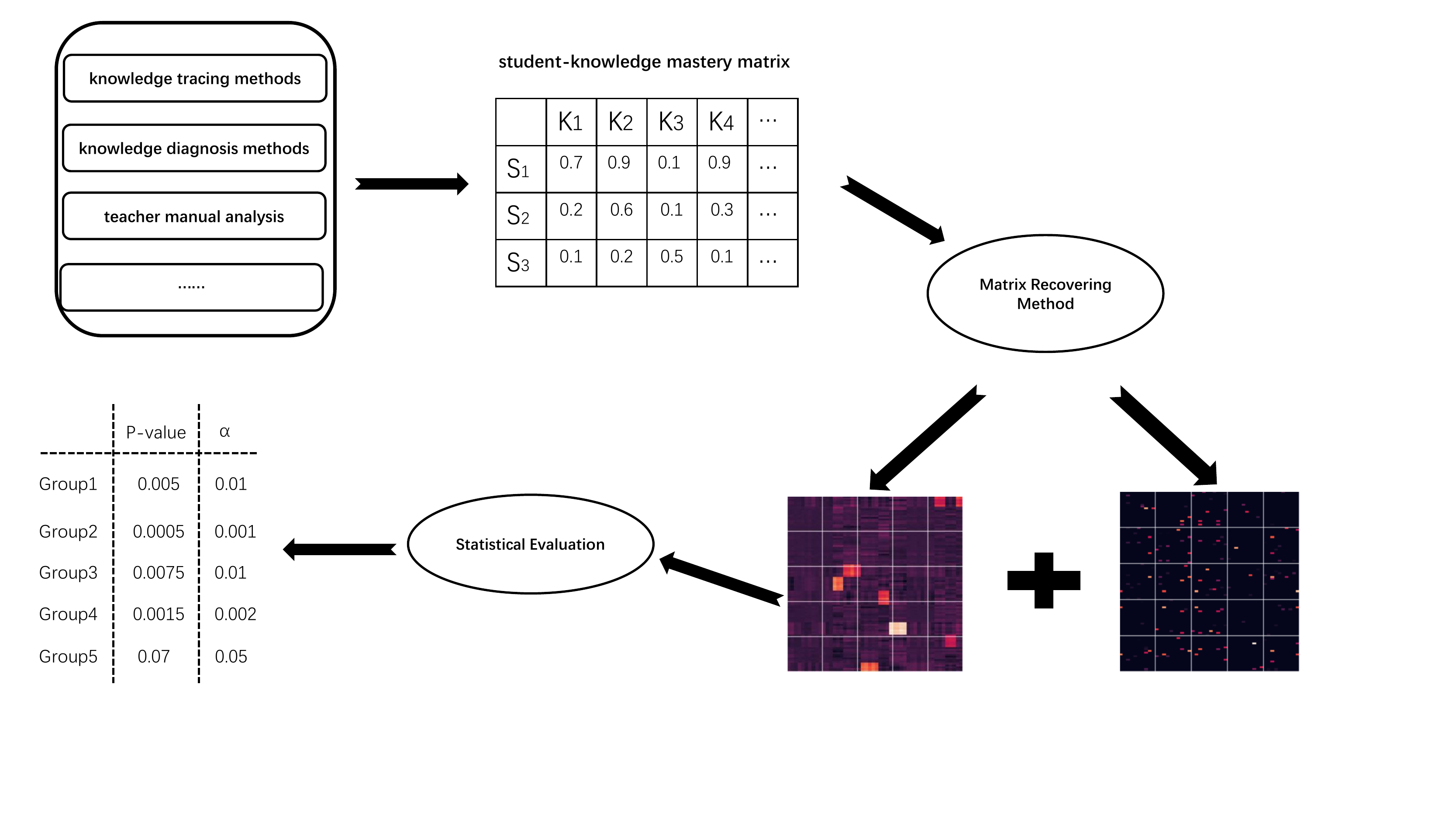}
	\caption{The framework of our method. It contains two steps: matrix recovering and statistical evalulation. Following above two steps, the method proposed can be very robust and accurate in the detection of common characteristics and individual characteristics for students.} 
	\label{fig1}
\end{figure}

\refFig{fig1} gives the flow of the overall framework. There are two key steps in the framework. In the first step, we perform matrix recovering method on the students' characteristics from different sources, such as knowledge tracing methods and manual analysis by instructors. After that, statistical evaluation is used to detect robust biclusters. The result validation step is aimed to guarantee the statistic significance for each discovered clusters, as stated in \cite{DBLP:journals/datamine/HenriquesM18}. In the following, we will explain these two steps in detail.

\begin{figure}[h]
	\centering
	\includegraphics[width=1\linewidth]{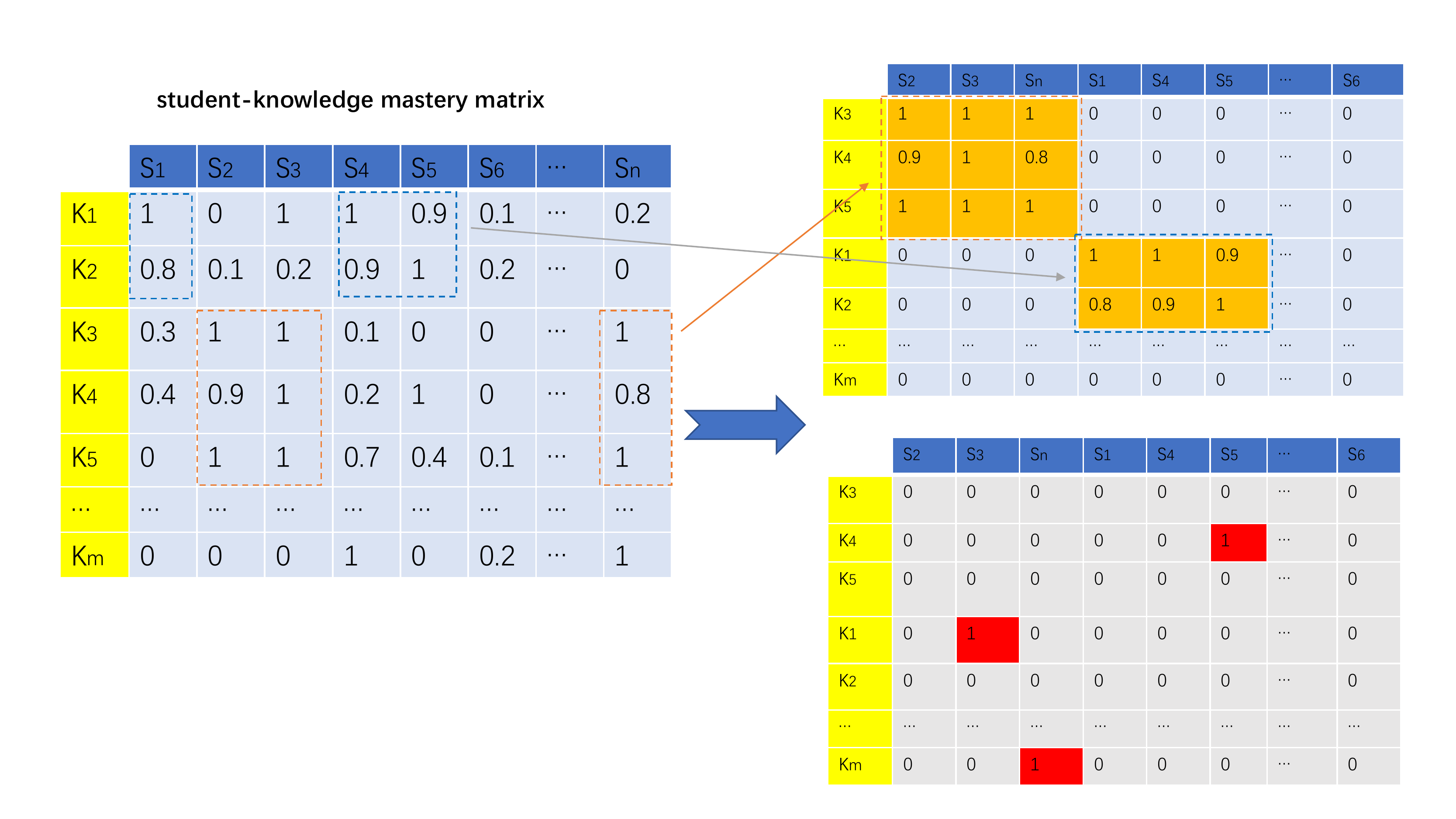}
	\caption{An example of matrix recovering for educational data. Base on the assumption that common characteristics form a low-rank matrix and individual characteristics form a sparse matrix, we can decompose the input matrix into two matrices.} 
	\label{fig2}
\end{figure}

\subsection{Matrix recovering}
The goal of the first step is to recover low-rank component and sparse component from original input matrix, respectively. \refFig{fig2} shows an example result of matrix recovering for educational data. In this subsection, we first present the mathematical formulation of matrix recovering problem. And then the solution to solve it is provided. Finally, we give some suggestions to select appropriate parameters.

\subsubsection{Formulation}
Mathematically, we can express the characteristics of students as a matrix $\bfD$ $\in$ $\RR{n}{m}$,
where each element $d_{ij}$ of the matrix is a value representing the mastery of knowledge for each student, and $n$ and $m$ are the numbers of students and knowledge topics in this article, respectively. Because our goal is to balance group teaching and individual mentoring, we need to be able to explore common characteristics shared by multiple students and individual characteristics for each student. As common characteristics of students can be represented as biclusters and individual characteristics for each student can be assumed as randomly distributed and sparse in the matrix, we can treat the balancing issue as a  problem of recovering a low-rank matrix $\bfX$ and a sparse matrix $\bfE$ from the orginal characteristics matrix $\bfD$.

Naturally, the following matrix decomposition model is proposed to detect two types of characteristics from input with noise:
\begin{align}\label{eq:decomp}
	\bfD = \bfX + \bfE + \epsilon,
\end{align}

In Eq. \eqref{eq:decomp}, $\bfX$ refers to the common characteristeris component. $\bfE$ refers to the individual component and $\epsilon$ is a noise component.

We consider the following minimization problem to achieve the decomposition:
\begin{align}\label{eq:nonrelax}
	\min_{\bfX,\bfE,\epsilon} ~ &{\half}\|\epsilon\|_F^2 + \alpha \rank{\bfX} + \beta \|\bfE\|_0 \nonumber \\
	\st ~ &\bfD = \bfX + \bfE + \epsilon,
\end{align}
where $\|\epsilon\|_F$ is the Frobenius norm, $\rank{\bfX}$ is the rank of matrix $\bfX$ and $\|\bfE\|_0$ is the $\ell_0$-norm. We can get a penalized maximum likelihood estimate with respect to the variables $\bfX,\bfE,\epsilon$ through
solving Eq. \eqref{eq:nonrelax}.

%where $\|\epsilon\|_F=\sqrt{\sum_{i,j}{\epsilon_{ij}^2}}$ is the Frobenius norm and $\|\bfE\|_0$ is the $\ell_0$-norm that counts the number of nonzero values in $\bfE$. The solution to \refEq{eq:nonrelax} will give a penalized maximum likelihood estimate with respect to the variables $\bfX,\bfE,\epsilon$.

Since the model proposed in \refEq{eq:nonrelax} is NP-hard, the convex relaxation approach is used to effectively recover $\bfX$ and $\bfE$. Specifically, the $\rank{\cdot}$ is replaced by the nuclear norm and the $\ell_0$-norm is replaced by the $\ell_1$-norm. The nuclear norm is defined as the sum of the singular values of $\bfX$, whic is the tightest convex surrogate to the rank operator \cite{fazel2002matrix} and has been widely used for low-rank matrix recovery \cite{candes2011robust}. The $\ell_1$-norm is defined as $\|\bfX\|_{1}=\sum_{i,j}{|X_{ij}|}$. The $\ell_1$ relaxation has proven to be a powerful technique for sparse signal recovery \cite{tropp2006just}.

Thus, we can solve the following problem, instead of directly solving \refEq{eq:nonrelax}:
{\footnotesize
	\begin{align}\label{eq:rpla}
		\mathcal{F}(X,E) = \min_{\bfX,\bfE}{~~\frac{1}{2}\|\bfD-\bfX-\bfE\|_F^2+\alpha\|\bfX\|_*+\beta\|\bfE\|_1}.
\end{align}}

\refEq{eq:rpla} is a convex problem so that the global optimal solution is unique. This means that the results of our method is stable. 

\subsubsection{Solution}\label{sec:alg}
We can solve the optimization problem of \refEq{eq:rpla} by alternatively solving the following two sub-problems until convergence:
\begin{align}
	\hat\bfX & \leftarrow \arg\min_{\bfX}\mathcal{F}(\bfX,\hat\bfE) \label{eq:admm_1}\\
	\hat\bfE & \leftarrow \arg\min_{\bfE}\mathcal{F}(\hat\bfX,\bfE) \label{eq:admm_2}.
\end{align}
The theoretical proof for the convergence can be found in \cite{boyd2010distributed}.
The problem in \refEq{eq:admm_1} can be reduced to
{\small
	\begin{align} \label{eq:B-step}
		\min_{\bfX}{~~\frac{1}{2}\|\bfD-\hat\bfE-\bfX\|_F^2+\alpha\|\bfX\|_*},
\end{align}}
which becomes a nuclear-norm regularized least-squares problem and has the following closed-form solution \cite{cai2010singular},
\begin{align}\label{eq:svt_solution}
	\hat\bfX = \SVT_{\alpha}\left(\bfD-\hat\bfE\right),
\end{align}
where $\SVT_{\lambda}$ refers to the singular value thresholding (SVT)
\begin{align}\label{eq:svt}
	\SVT_{\lambda}(\bfM) = \sum_{i=1}^{r}{(\sigma_i-\lambda)_+\bfu_i\bfv_i^T}.
\end{align}
Here, $(x)_+ = \max(x,0)$. $\{\bfu_i\}$, $\{\bfv_i\}$, and $\{\sigma_i\}$ are the left singular vectors, the right singular vectors, and the singular values of $\bfM$, respectively.

The problem in \refEq{eq:admm_2} can be rewritten as
\begin{align}\label{eq:E-step}
	\min_{\bfE}\frac{1}{2}\|\bfD-\hat\bfX-\bfE\|_F^2+\beta\|\bfE\|_1.
\end{align}
It admits a closed-form solution
\begin{align}
	\hat\bfE = \ST_{\beta}\left(\bfD-\hat\bfX\right),
\end{align}
where $\ST_{\beta}(\bfM)_{ij}=\mbox{sign}(M_{ij})(M_{ij}-\beta)_+$ refers to the elementwise soft-thresholding operator \cite{boyd2010distributed}.

%\begin{algorithm}
%	\caption{The algorithm to solve \refEq{eq:rpla}.}\label{alg:rpla}
%	\begin{algorithmic}[1]
%		\algsetup{linenodelimiter=.}
%		\STATE {\bf Input:} $\bfD$
%		\STATE Initialize all variables to be zero.
%		\REPEAT
%		\STATE Update $\bfX$ by solving \refEq{eq:B-step} via singular value thresholding.
%		\STATE Update $\bfE$ by solving \refEq{eq:E-step} via soft thresholding.
%		\UNTIL{convergence}
%		\STATE {\bf Output:} $\hat{\bfX}$ and $\hat{\bfE}$
%	\end{algorithmic}
%\end{algorithm}

\subsubsection{Parameter selection}
Two parameters need to be estimated in the first step. In this article, we give some suggestions to select proper estimations via the analysis of the size of the input matrix $(n,m)$ and the standard variation of the noise $\sigma$ \cite{candes2011robust,zhou2010stable}.

Firstly, we estimate an intermediate variable $\sigma$ from the data by the median-absolute-deviation estimator \cite{meer1991robust}
\begin{align}
	\hat\sigma = 1.48~\mbox{median}\left\{|\bfD-\mbox{median}(\bfD)|\right\}.
\end{align}

For parameter $\alpha$, it serves as a threshold in the SVT step in \refEq{eq:svt} so that it should be large enough to threshold out the noise but not too large to over-shrink the signal \cite{zhou2010stable}. A proper value is $\alpha=(\sqrt{n}+\sqrt{p})\sigma$, which is the expected $\ell_2$-norm of a $n\times p$ random matrix with entries sampled from $\mathcal{N}(0,\sigma^2)$. In practice, we can adjust $\alpha$ around this value to fit real data.

%First we choose $\alpha$ following the same way as in \cite{candes2010matrix,zhou2010stable}. When we fix $\bfE = 0$ in \refEq{eq:rpla}, the solution $\hat\bfX$ of \refEq{eq:rpla} is equal to the singular value thresholding version of $\bfD$ with threshold $\alpha$. Similarly, when we fix $\bfX = 0$ in \refEq{eq:rpla}, the solution $\hat\bfE$ is equal to the entry-wise shrinkage version of $\bfD$ with threshold $\beta$. Thus, we can choose $\alpha$ to be the smallest value such that the minimizer of \refEq{eq:rpla} is likely to be $\hat\bfX = \hat\bfE = 0$. In this way, $\alpha$ is large enough to threshold away the noise, but not too large to over-shrink the original matrices. 
%
%Since $\epsilon$ is modeled by i.i.d. Gaussian distribution with a zero mean, we estimate $\alpha=(\sqrt{n}+\sqrt{p})\sigma$, which is the expected $\ell_2$-norm of a $n\times p$ random matrix with entries sampled from $\mathcal{N}(0,\sigma^2)$.

For parameter $\beta$, there is a relative weight $\lambda=\beta/\alpha$ that balances the two terms in $\alpha\|\bfX\|_*+\beta\|\bfE\|_1$ and consequently controls the rank of $\bfX$ and the sparsity of $\bfE$.
It has been proved that $\lambda=1/\sqrt{m}$ gives a large probability of recovering $\bfX$ and $\bfE$ under their assumed conditions \cite{candes2011robust}.
% and the value of $\lambda$ can be adjusted slightly to obtain the best results in specific applications. 
In practice, we should set different values of $\lambda$ to keep sufficient characteristics in $\bfX$ in specific applications.

\subsection{Statistical evaluation}
When the low-rank matrix $X$ is recovered, we can get the biclusters in two ways. The first way is to perform biclustering methods on the low-rank matrix, rather than on the original matrix. The other one is similar to Spectra Biclustering \cite{kluger2003spectral}, by applying traditional clustering methods on the low-rank matrix to cluster students and knowledge topcis, respectively. In order to compare our framework with biclustering methods in the experiments section, we use the latter one to detect biclusters in this article. 
%In this article, we adopt the latter one. In general, we cluster the students and knowledge topics into rank(X) categories, respectively. Thus, we can get rank($X$) * rank($X$) biclusters.
However, as small biclusters can have high levels of homogeneity by chance, some ones in the detected biclusters are spurious and insignificant. In order to filter false positive biclusters and control the false discovery rate, statistical evaluation are needed.

In this article, we adopt the method BSig proposed in \cite{DBLP:journals/datamine/HenriquesM18} to evaluate the statistical significance of detected biclusters and reject those biclusters with high $P$-values. The BSig method provides the unprecedented possibility to minimize the number of false positive biclusters without incurring on false negatives. It first approximates a null model of the target educational data and then appropriately tests each bicluster in accordance with its underlying coherence. Finally, we reject those biclusters with $P-$values higher than Bonferroni correction thresholds.

\section{Experiments}
To test the method proposed in this article comprehensively, we conduct several comparison experiments both on synthetic and real datasets with those biclustering algorithms drawn from important studies in the biclustering literature. In detail, we compare the following four typical biclustering methods on synthetic datasets: 

\begin{enumerate}
	\item {\textbf{FABIA \cite{DBLP:journals/bioinformatics/HochreiterBHMMKKSLTBGSC10}.}} This algorithm is based on factor analysis and it is a multiplicative model that assumes non-Gaussian signal distributions with heavy tails. 
	\item {\textbf{LAS \cite{shabalin2009finding}.}} This method searches for patterns from input matrix by locally maximizing a Bonferroni based significance score iteratively.
	\item {\textbf{ISA \cite{10.1093/bioinformatics/bth166}.}} Iterative Signature Algorithm first randomly selects columns and rows, and then evaluates and updates them through iterative steps until convergence.
	\item {\textbf{Spectra Biclustering \cite{kluger2003spectral}.}} Spectral biclustering method assumes that the input data matrix has a hidden checkerboard structure and uses singular value decomposition to find it.
\end{enumerate}

For these methods, we use the default settings of them in biclustlib \cite{padilha2017}, a python library of biclustering algorithms.
Meanwhile, we use six evaluation metrics in the experiments to compare the biclustering results: liu wang match score \cite{10.1093/bioinformatics/btl560}, prelic recover score \cite{prelic2006systematic}, prelic relevance score \cite{prelic2006systematic}, csi \cite{campello2010generalized}, cluster error \cite{patrikainen2006comparing} and fabia consensus score \cite{DBLP:journals/bioinformatics/HochreiterBHMMKKSLTBGSC10}. They give scores from different views by comparing the predicted biclusters against the ground-truth ones. Larger scores assigned by these metrics indicate better performances of the biclustering methods. The details of these metrics are referred to the corresponding papers.

Additionally, as our method can identify sparse signals at the same time, we validate the performance of identifying sparse signals using precision, recall and F1-score. 

All the experiments are tested on the ThinkPad T480 Laptop computer with 1.80GHz CPU and 16G main memory.

\subsection{Experiments with synthetic data}
Firstly, we test the effectiveness of our method on four synthetic biclustering type: constant, shift, scale and shift \& scale. The biclustering datasets are generated based on the procedure proposed by \cite{DBLP:journals/bib/ErenDKC13}. The details of the settings for each data type are listed in Table 1. After generating biclustering data, we add sparse signals to the matrix. Specifically, a sparse matrix $E$ is generated through the following way: each element of this sparse matrix independently takes on value 0 with probability 1 - $p_s$, and value 6 with probability $p_s = 0.01$. Finally, we shuffle the resulted matrix.

\begin{table}
	\centering
	\caption[\textbf{The detail of settings for each synthetic data.}]{\textbf{The detail of settings for each synthetic data.} }
	\begin{tabular}{p{3cm}|p{5cm}}
		\hline
		%\multirow{1}{}{}
		\textbf{type} &\textbf{setting }\\ \hline
		Constant biclusters  &	nrows=300, ncols=50, nclusts=5, bicluster\_signals=5, nclustcols=8, nclustrows=5, noise=1, shuffle=True\\ \hline
		Shift biclusters  &	nrows=300, ncols=50, nclusts=5, bicluster\_noise=[0.01]*5, noise=1, base\_loc=1, shift\_loc=1, shift\_scale=3, shuffle=True \\ \hline
		Scale biclusters  &	nrows=300, ncols=50, nclusts=5, bicluster\_noise=[0.01]*5, base\_loc=1, scale\_scale=3, scale\_loc=0, shuffle=True \\ \hline
		Shift-scale biclusters  &	nrows=300, ncols=50, nclusts=5, bicluster\_noise=[0.01] * 5, base\_scale=1, scale\_scale=2, shift\_scale=3, shuffle=True \\ \hline
	\end{tabular}
\end{table}

\begin{figure*}[htbp]
	\centering
	\subfigure{
		\includegraphics[scale=0.23]{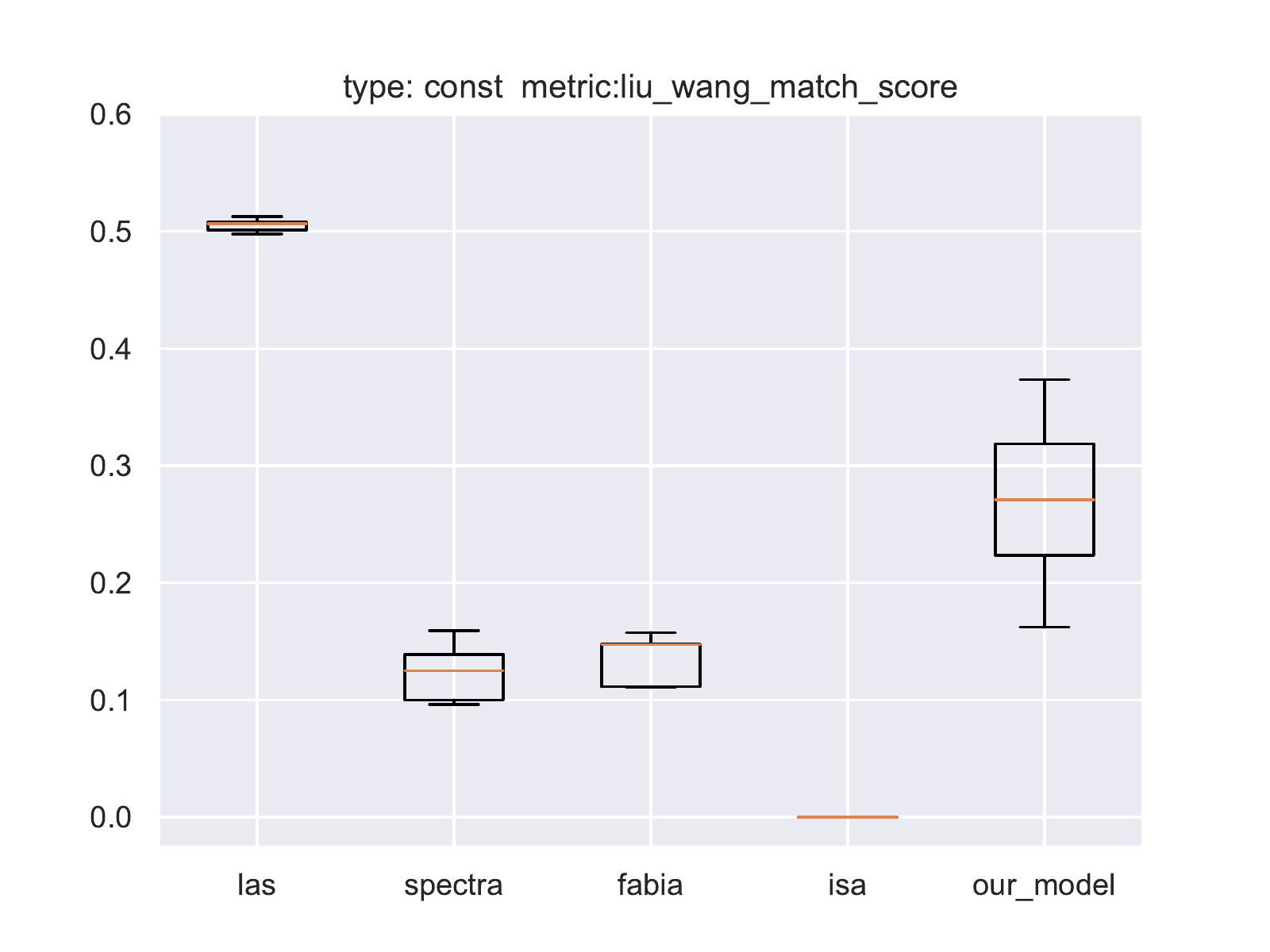}
		\label{const_1}
	}
	\subfigure{
		\includegraphics[scale=0.23]{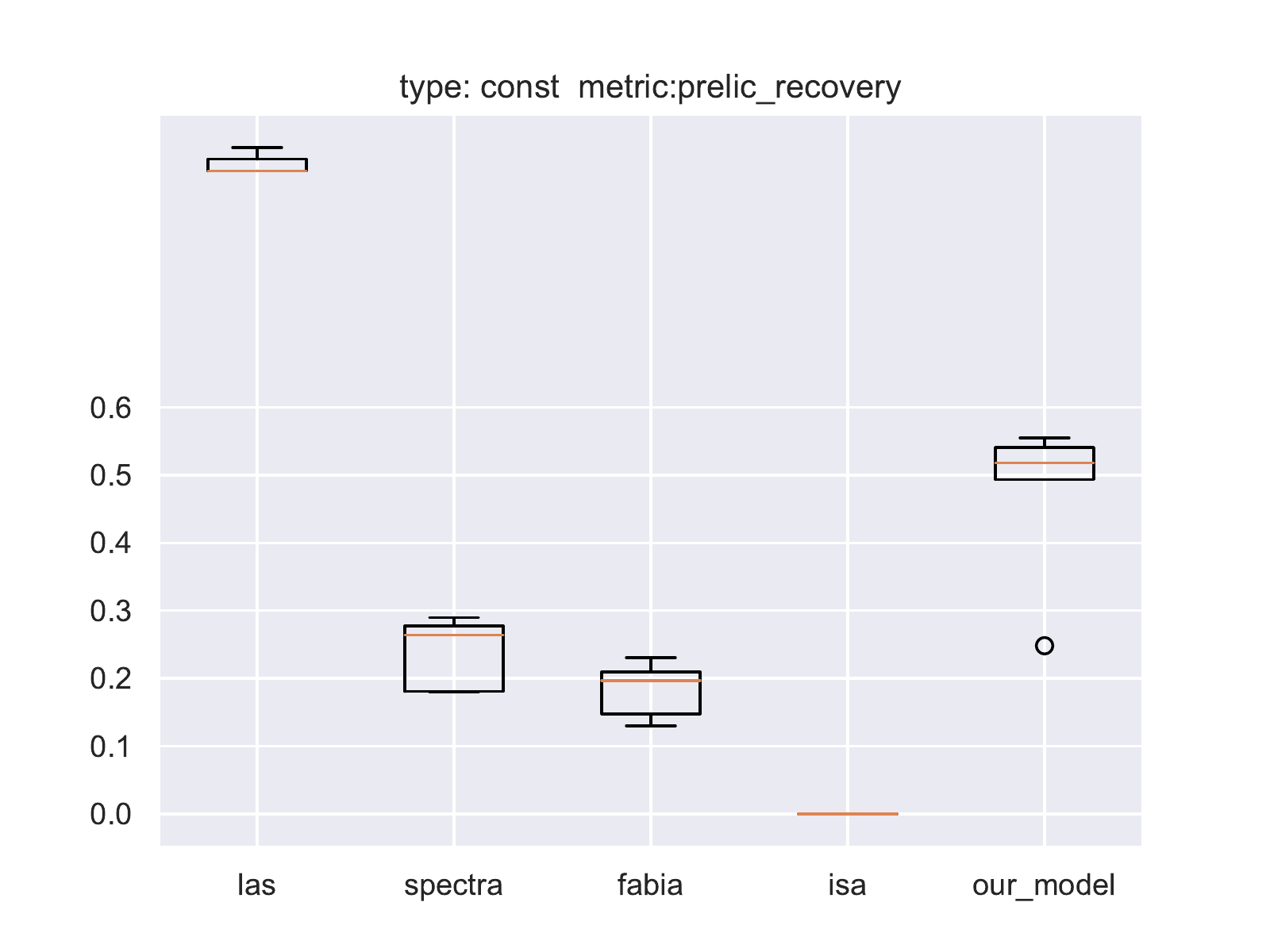}
		\label{const_2}
	}
	\subfigure{
		\includegraphics[scale=0.23]{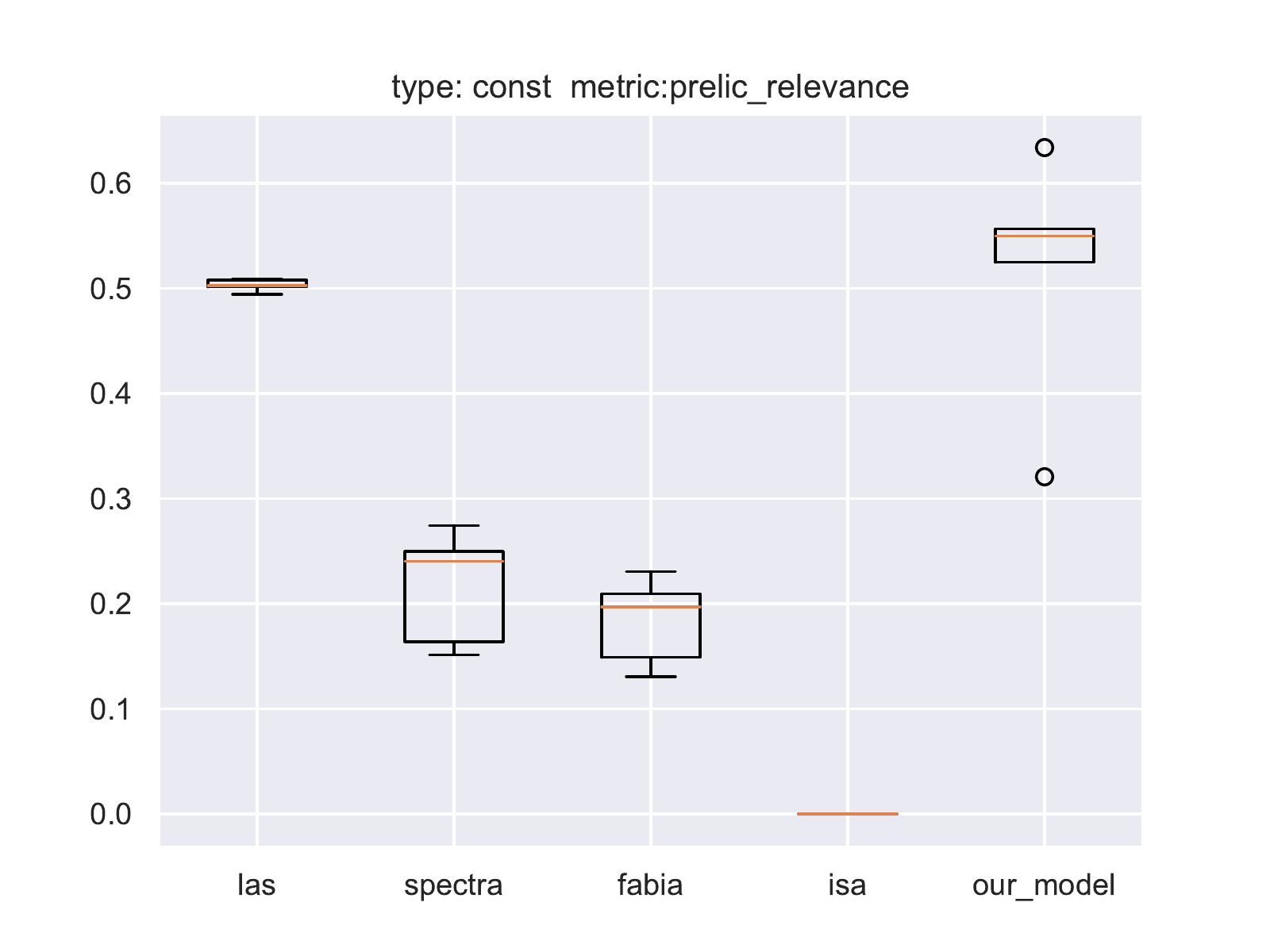}
		\label{const_3}
	}
	\subfigure{
		\includegraphics[scale=0.23]{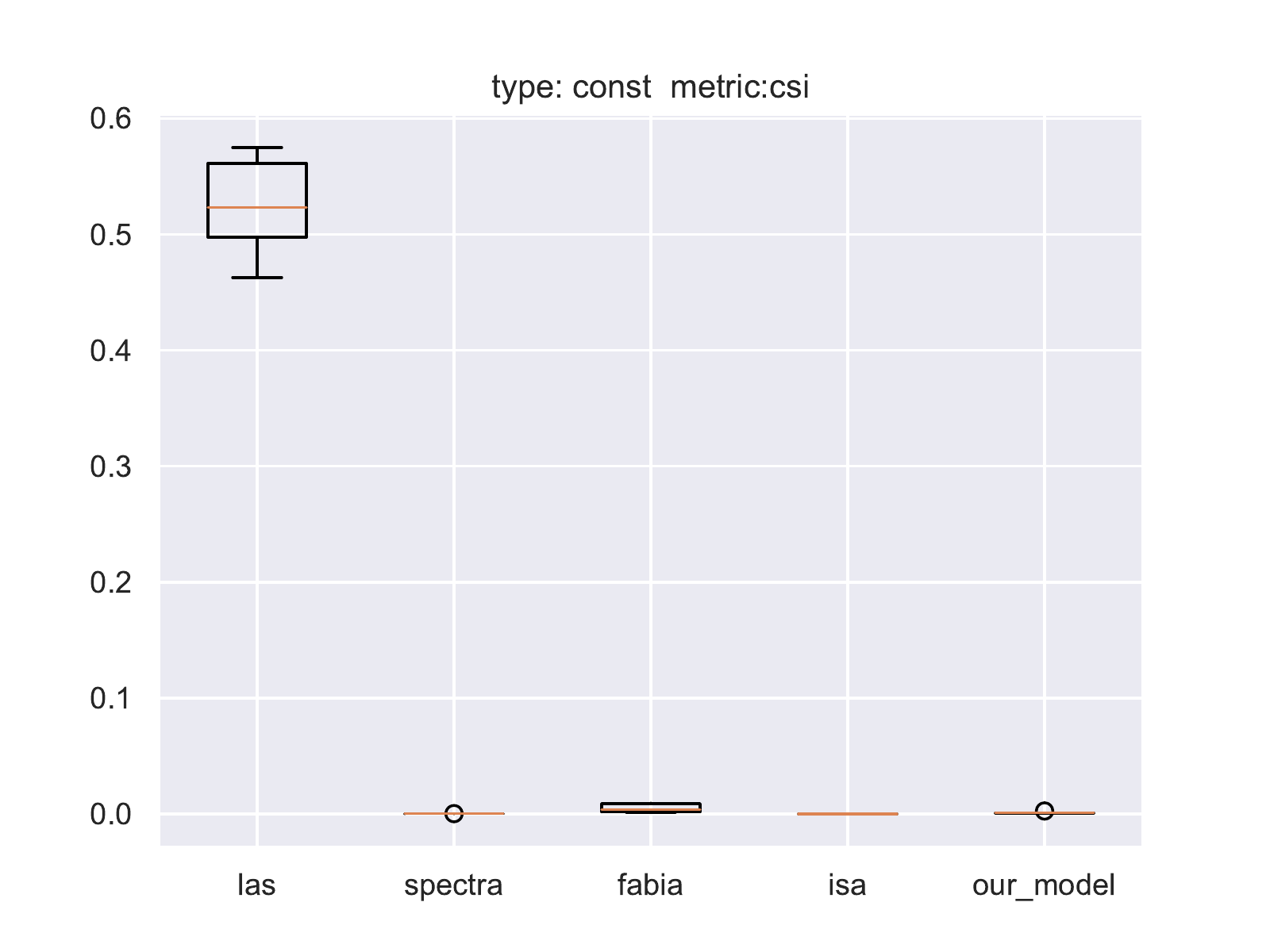}
		\label{const_4}
	}
	\quad
	\subfigure{
		\includegraphics[scale=0.23]{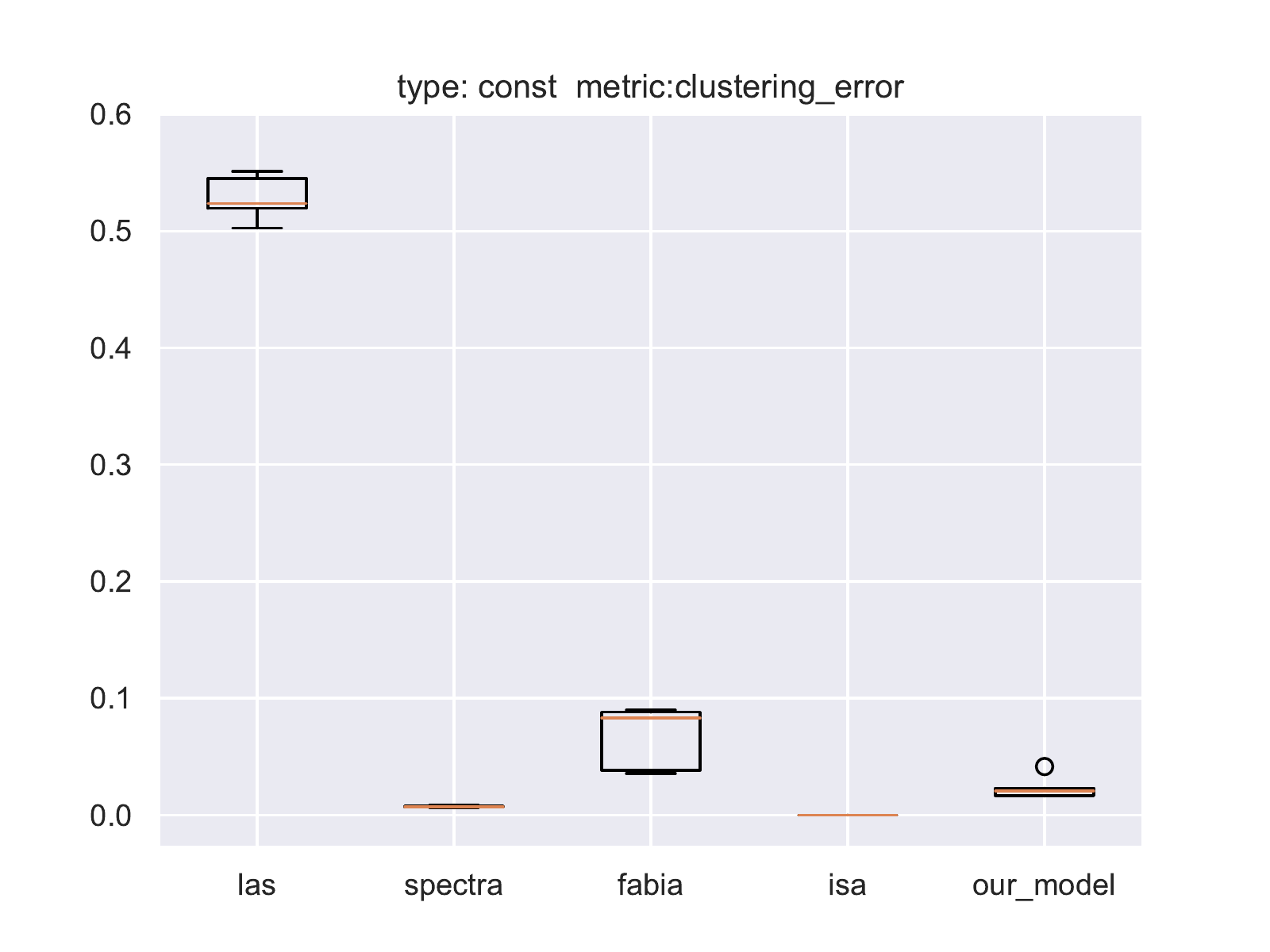}
		\label{const_5}
	}
	\subfigure{
		\includegraphics[scale=0.23]{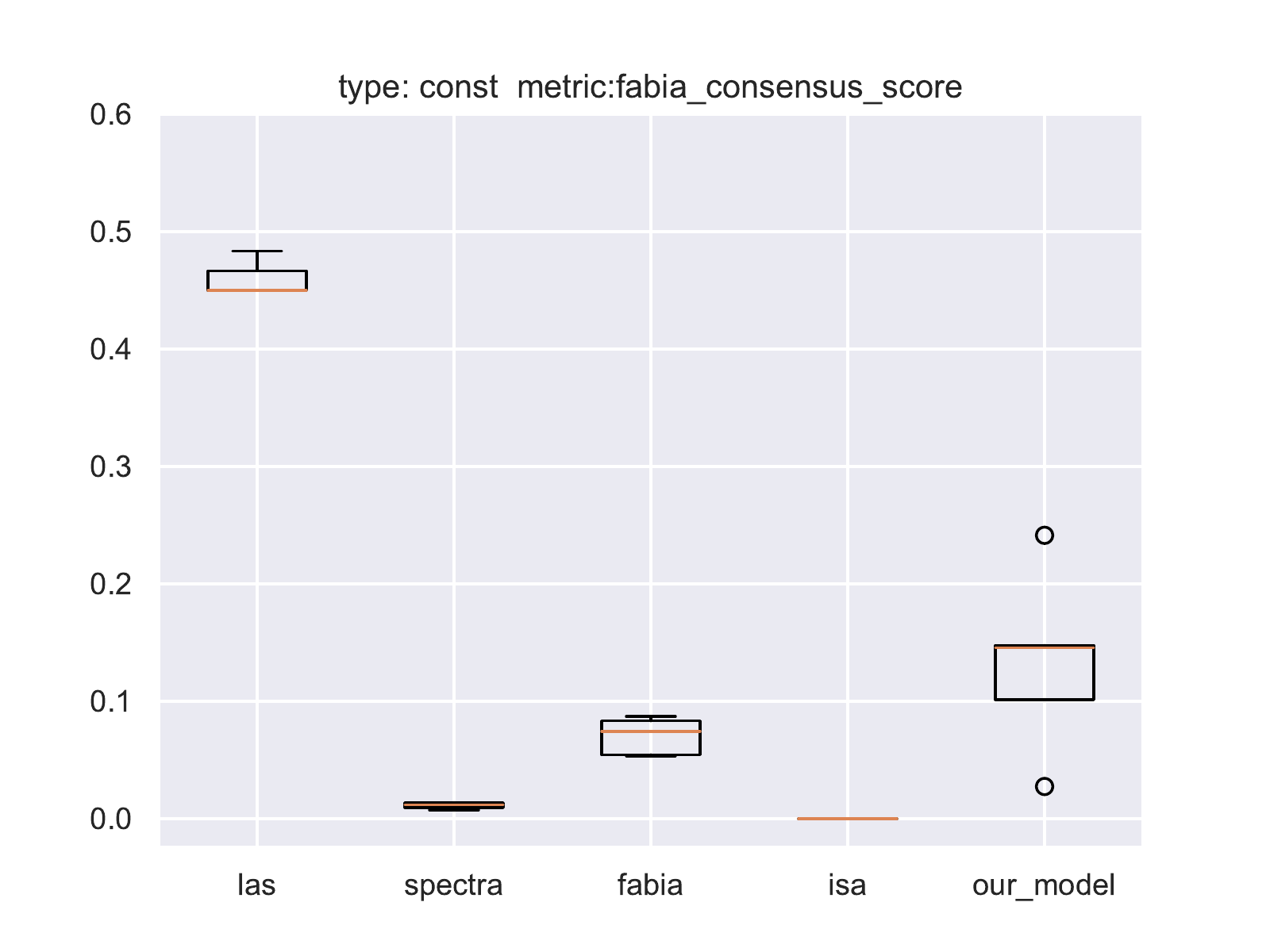}
		\label{const_6}
	}
	\subfigure{
		\includegraphics[scale=0.23]{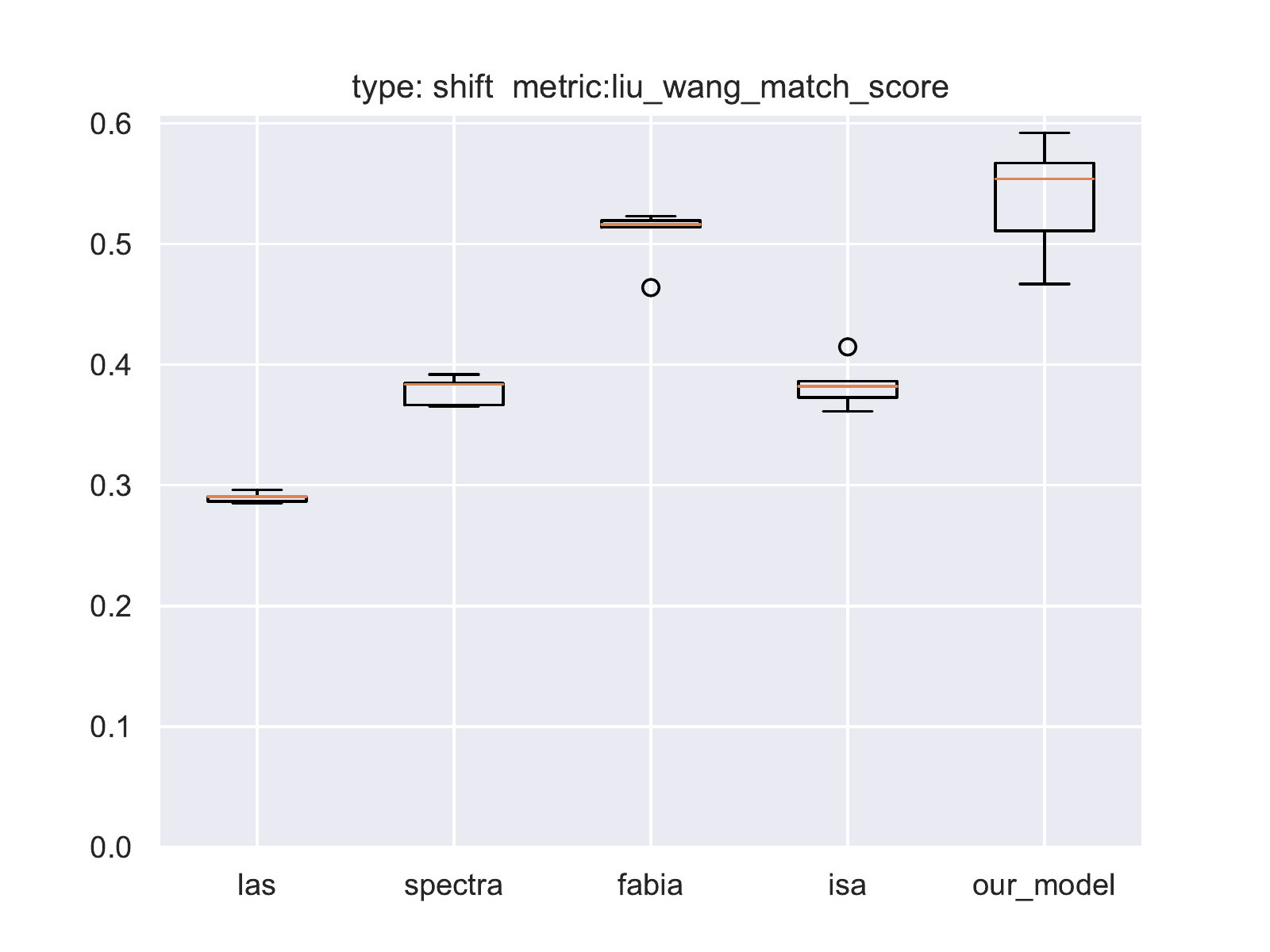}
		\label{st_1}
	}
	\subfigure{
		\includegraphics[scale=0.23]{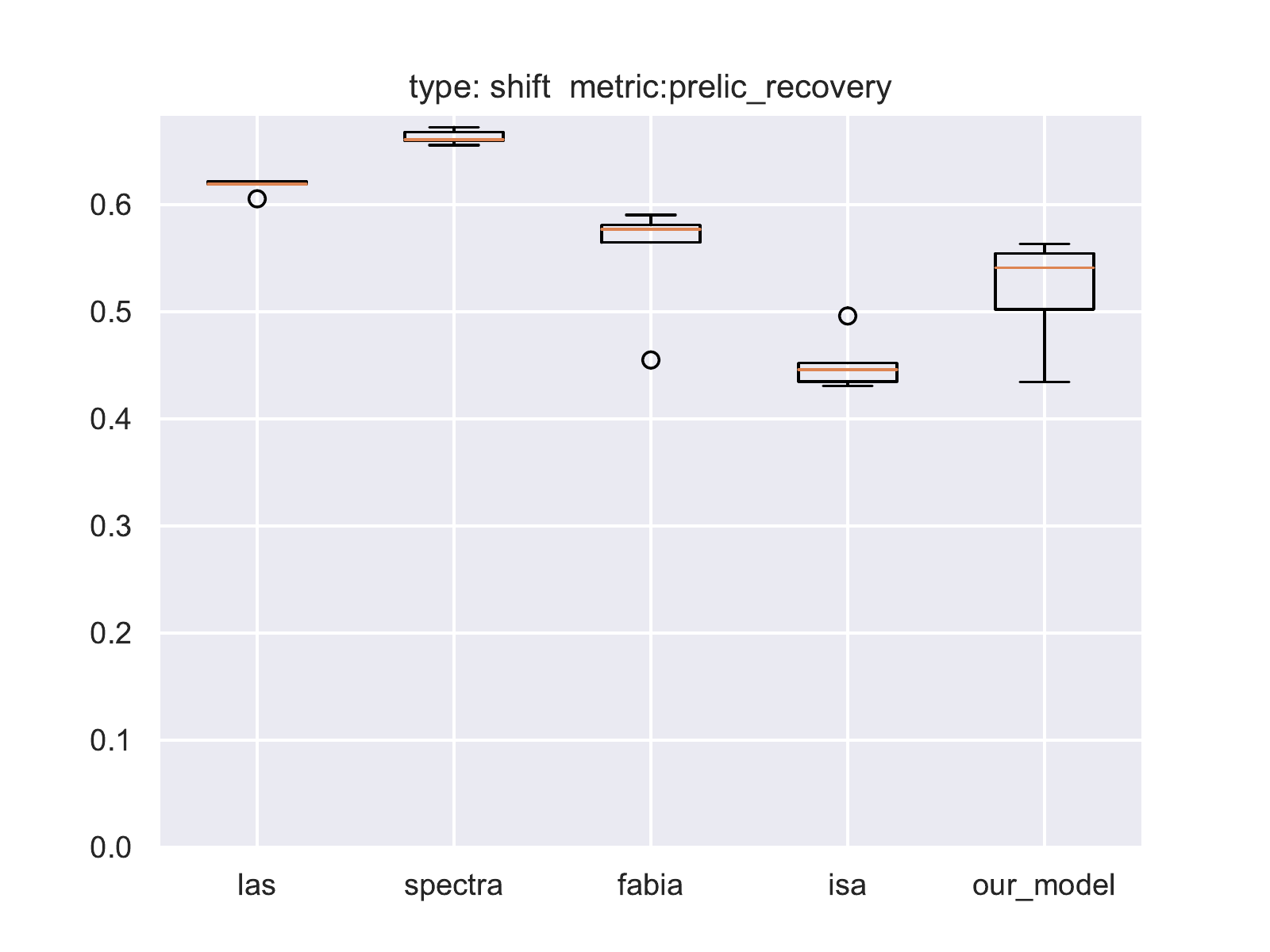}
		\label{st_2}
	}
	\quad
	\subfigure{
		\includegraphics[scale=0.23]{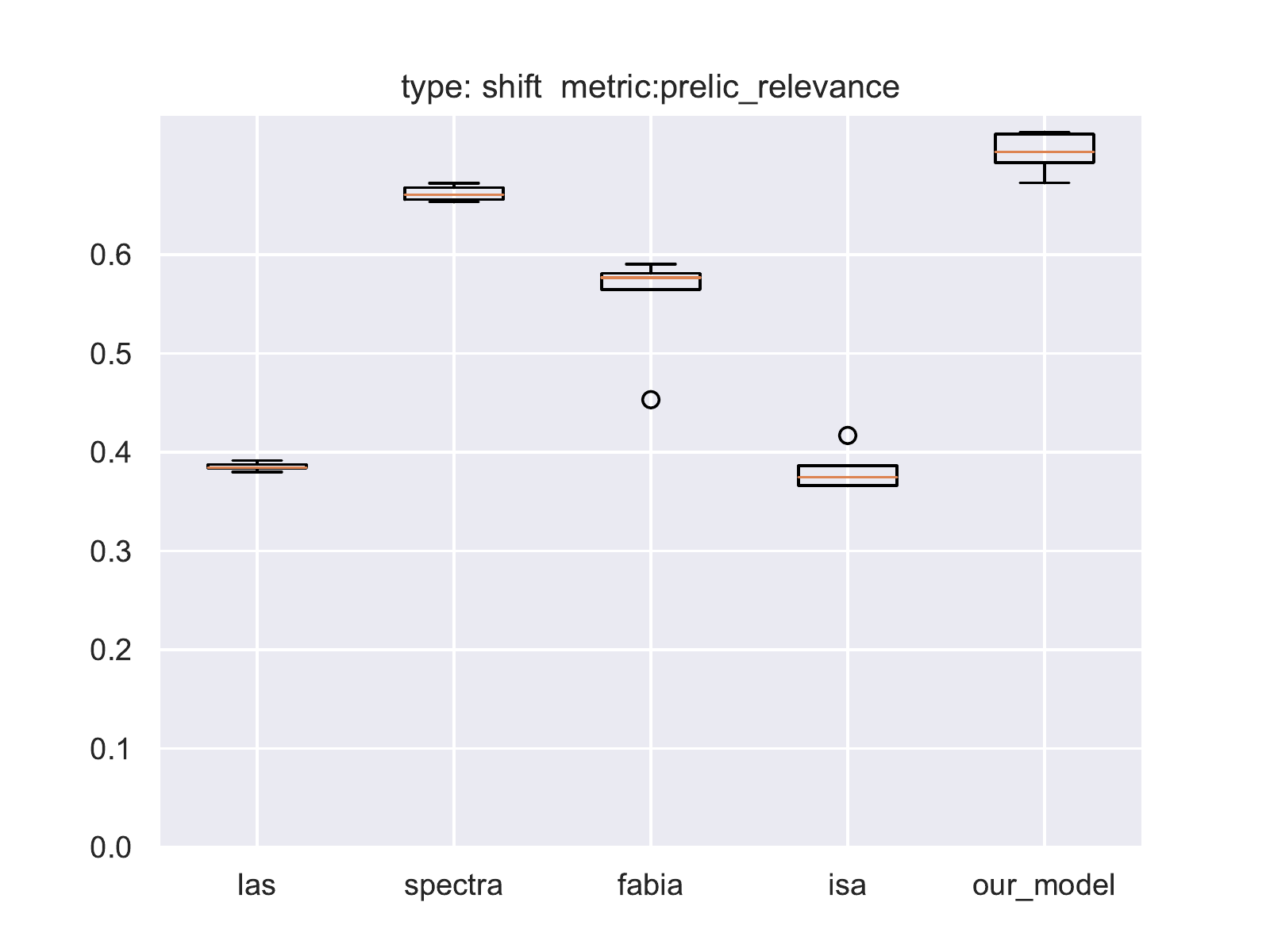}
		\label{st_3}
	}
	\subfigure{
		\includegraphics[scale=0.23]{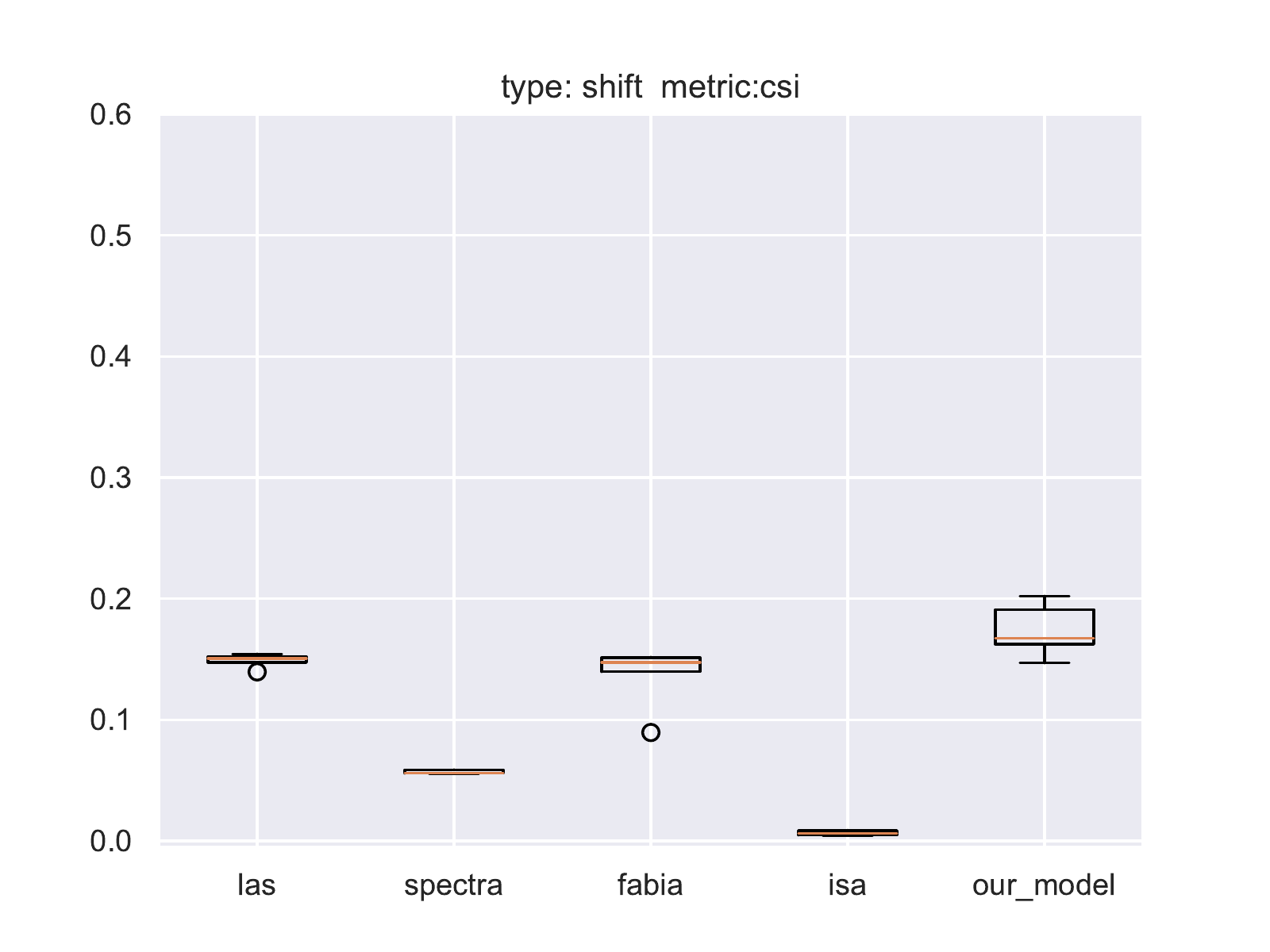}
		\label{st_4}
	}
	\subfigure{
		\includegraphics[scale=0.23]{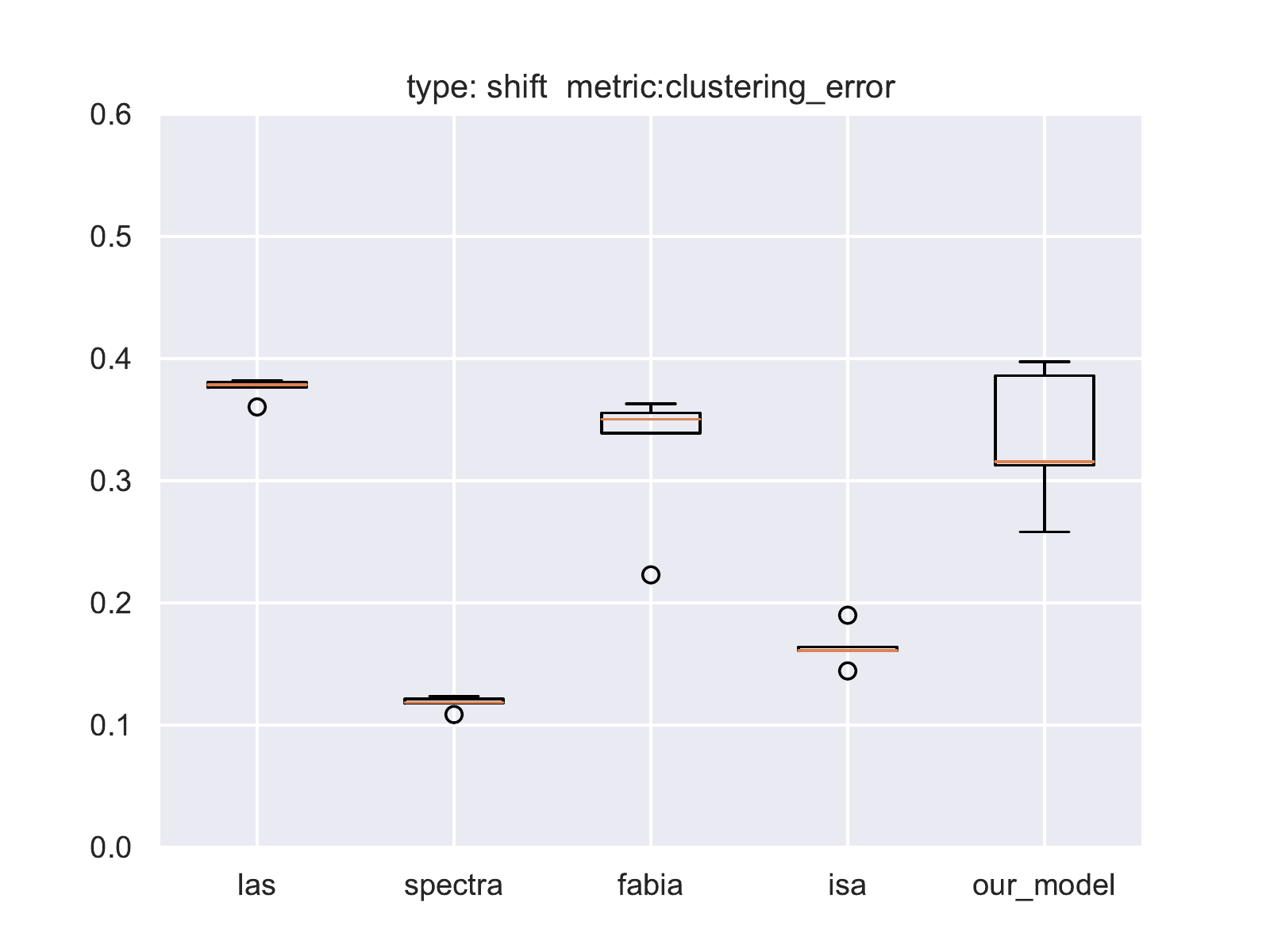}
		\label{st_5}
	}
	\subfigure{
		\includegraphics[scale=0.23]{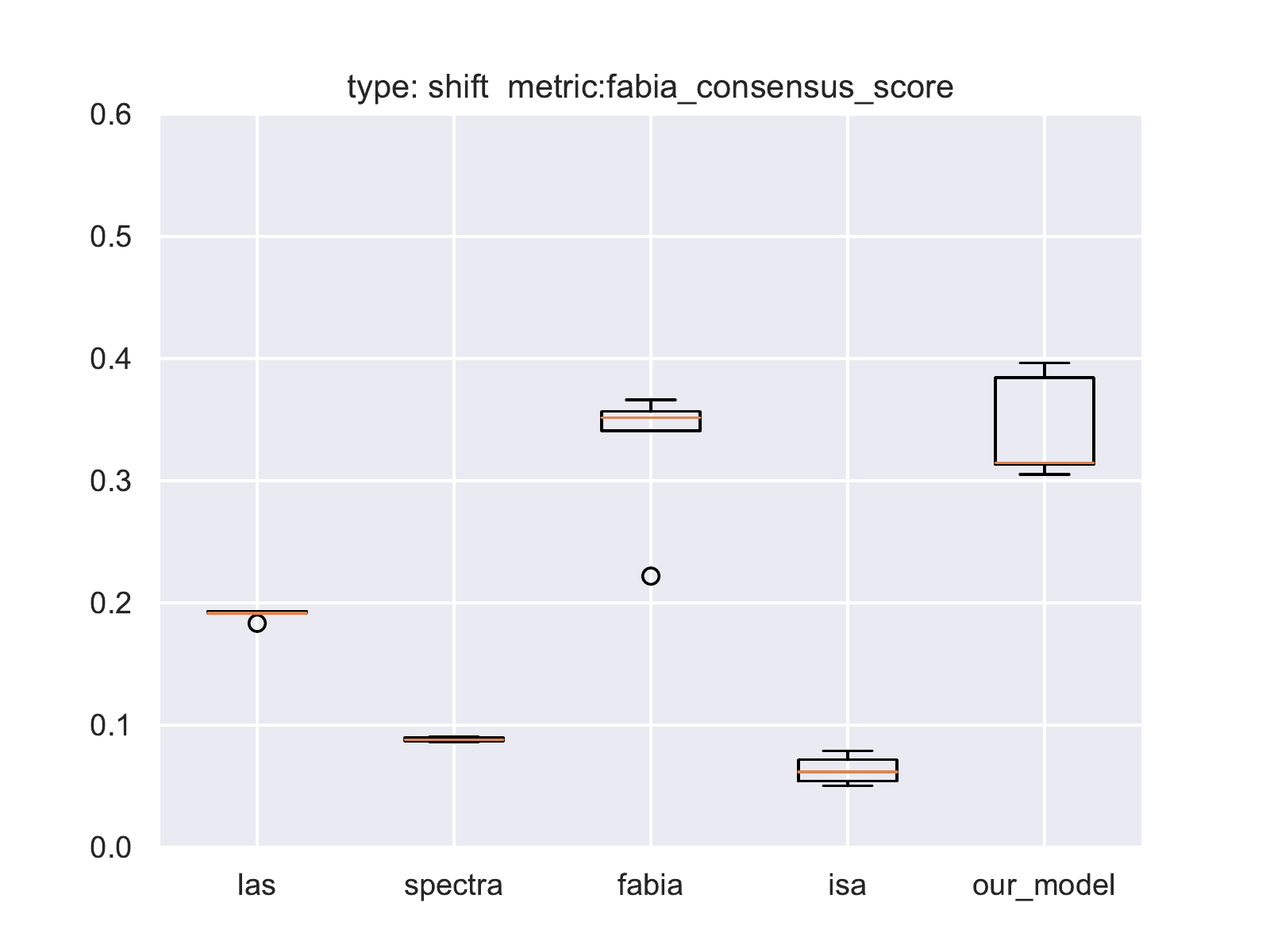}
		\label{st_6}
	}
	\quad
	\subfigure{
		\includegraphics[scale=0.23]{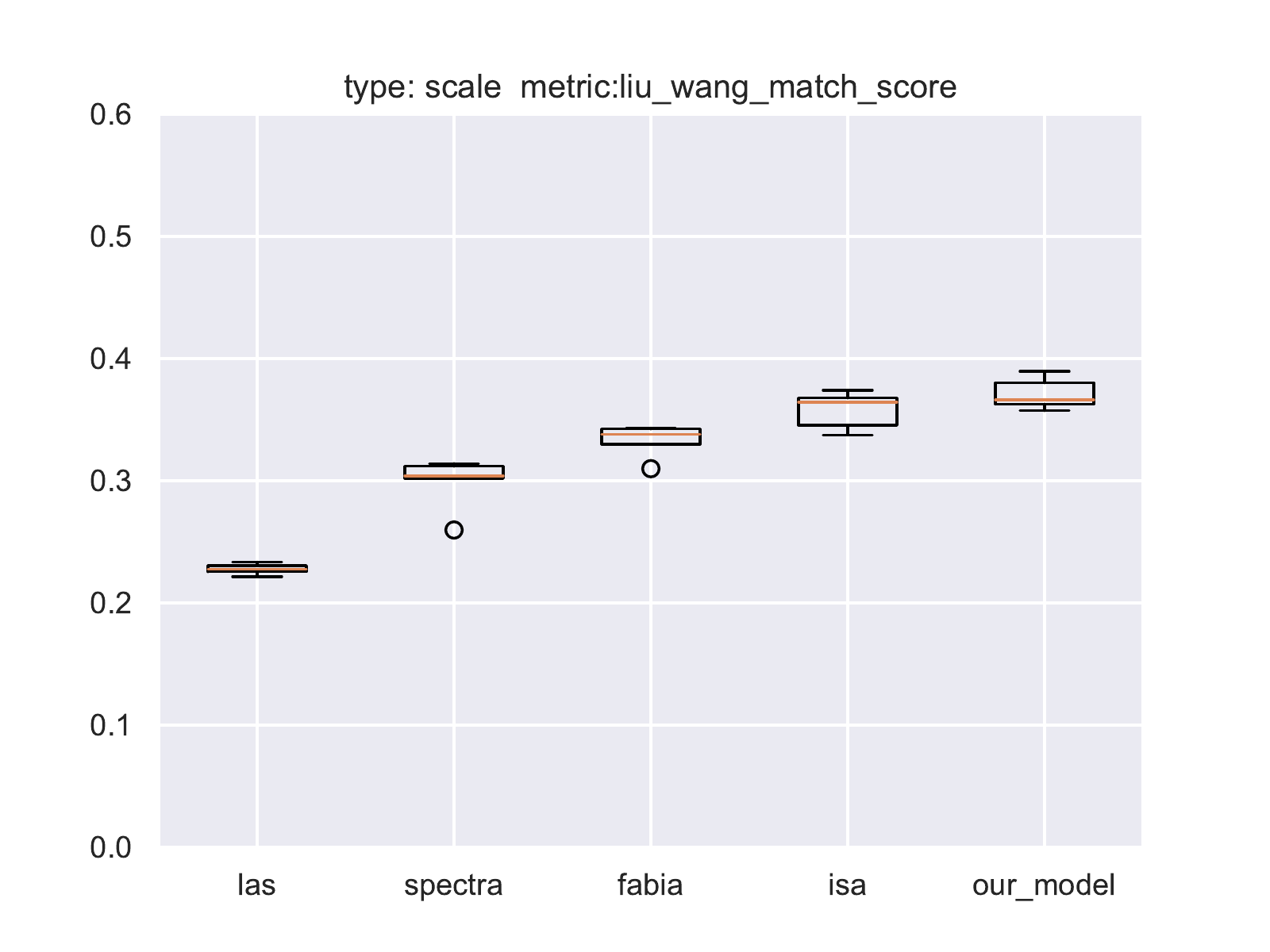}
		\label{scale0}
	}
	\subfigure{
		\includegraphics[scale=0.23]{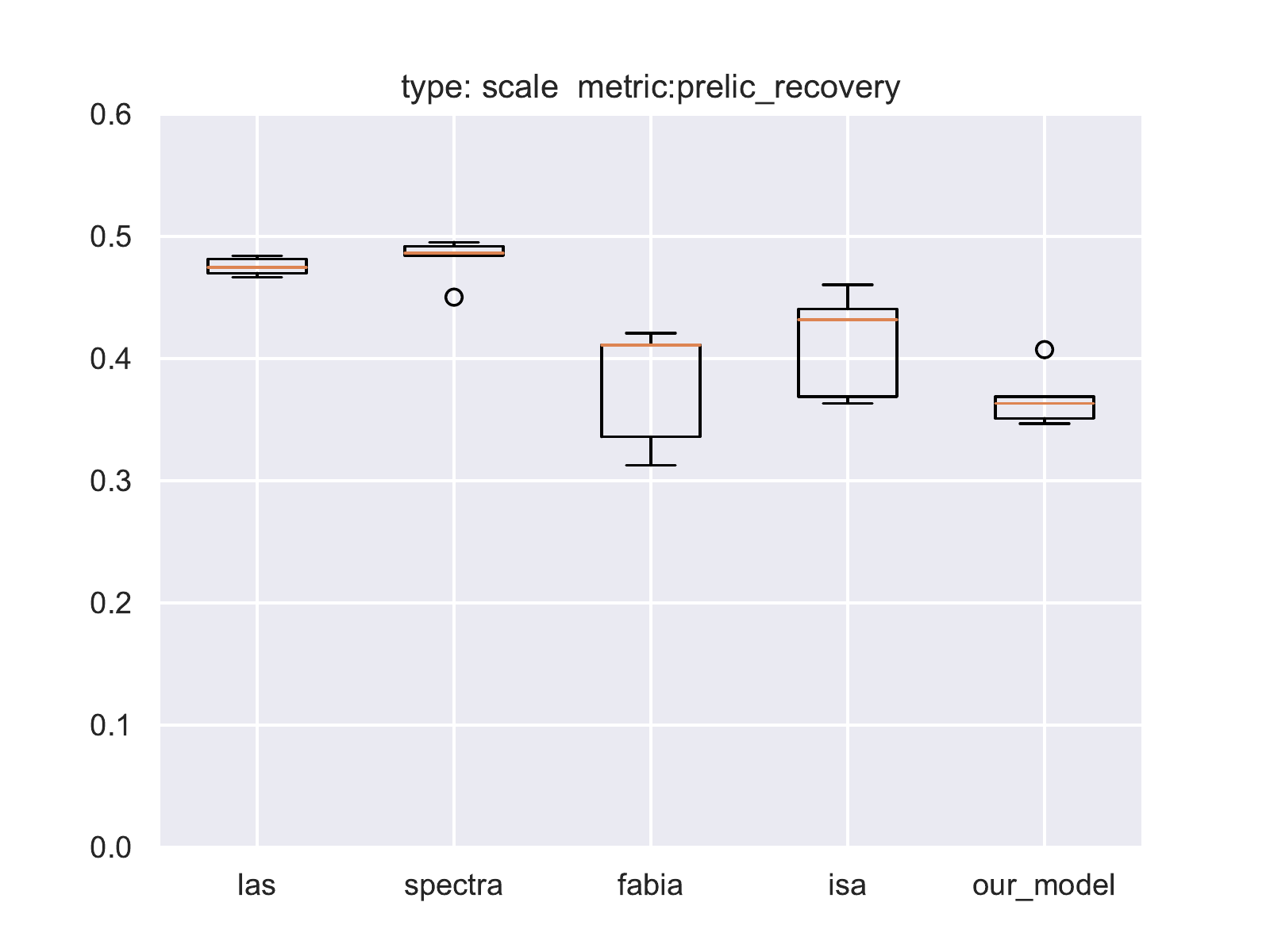}
		\label{scale1}
	}
	\subfigure{
		\includegraphics[scale=0.23]{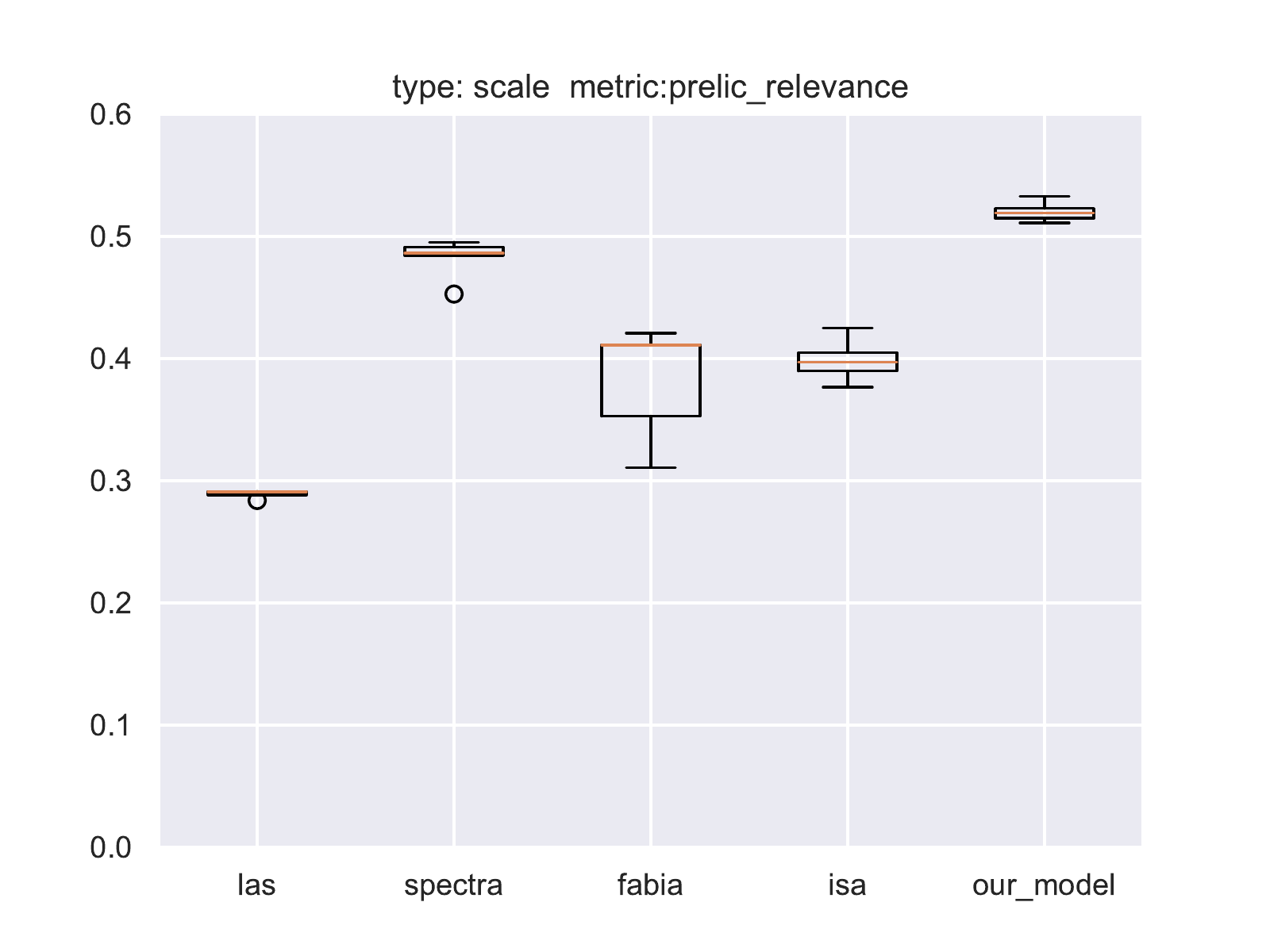}
		\label{scale2}
	}
	\subfigure{
		\includegraphics[scale=0.23]{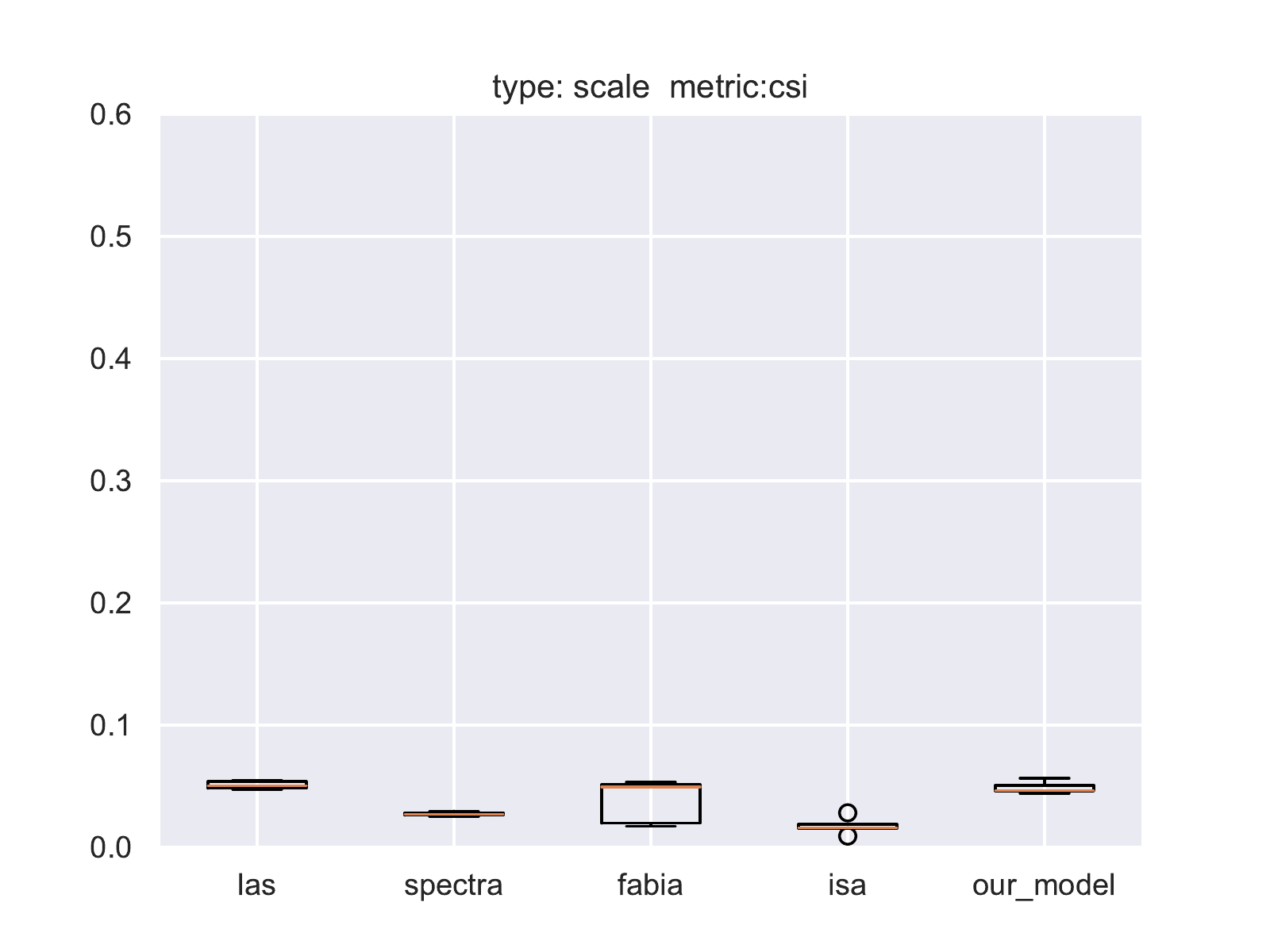}
		\label{scale3}
	}
	\quad
	\subfigure{
		\includegraphics[scale=0.23]{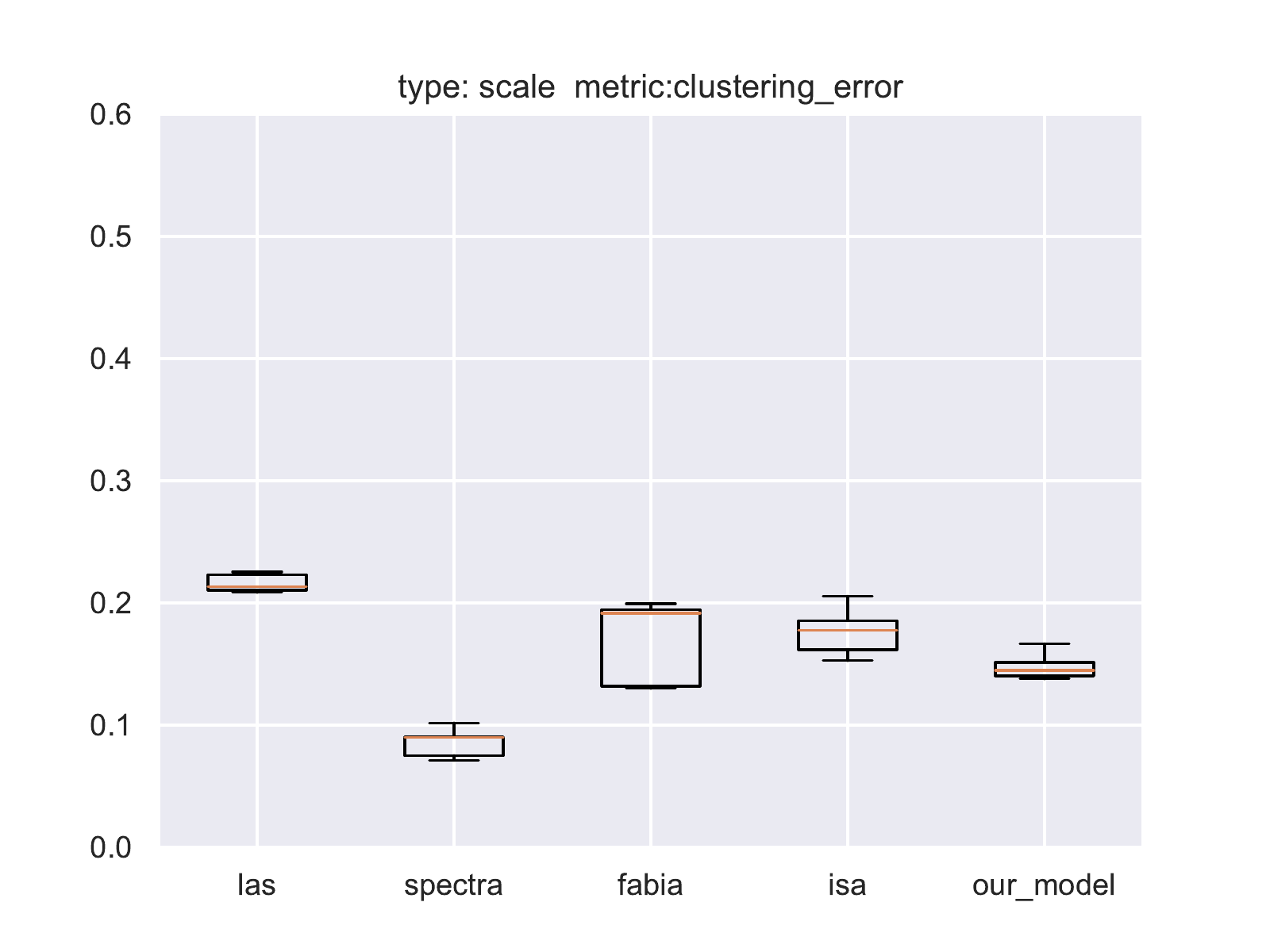}
		\label{scale4}
	}
	\subfigure{
		\includegraphics[scale=0.23]{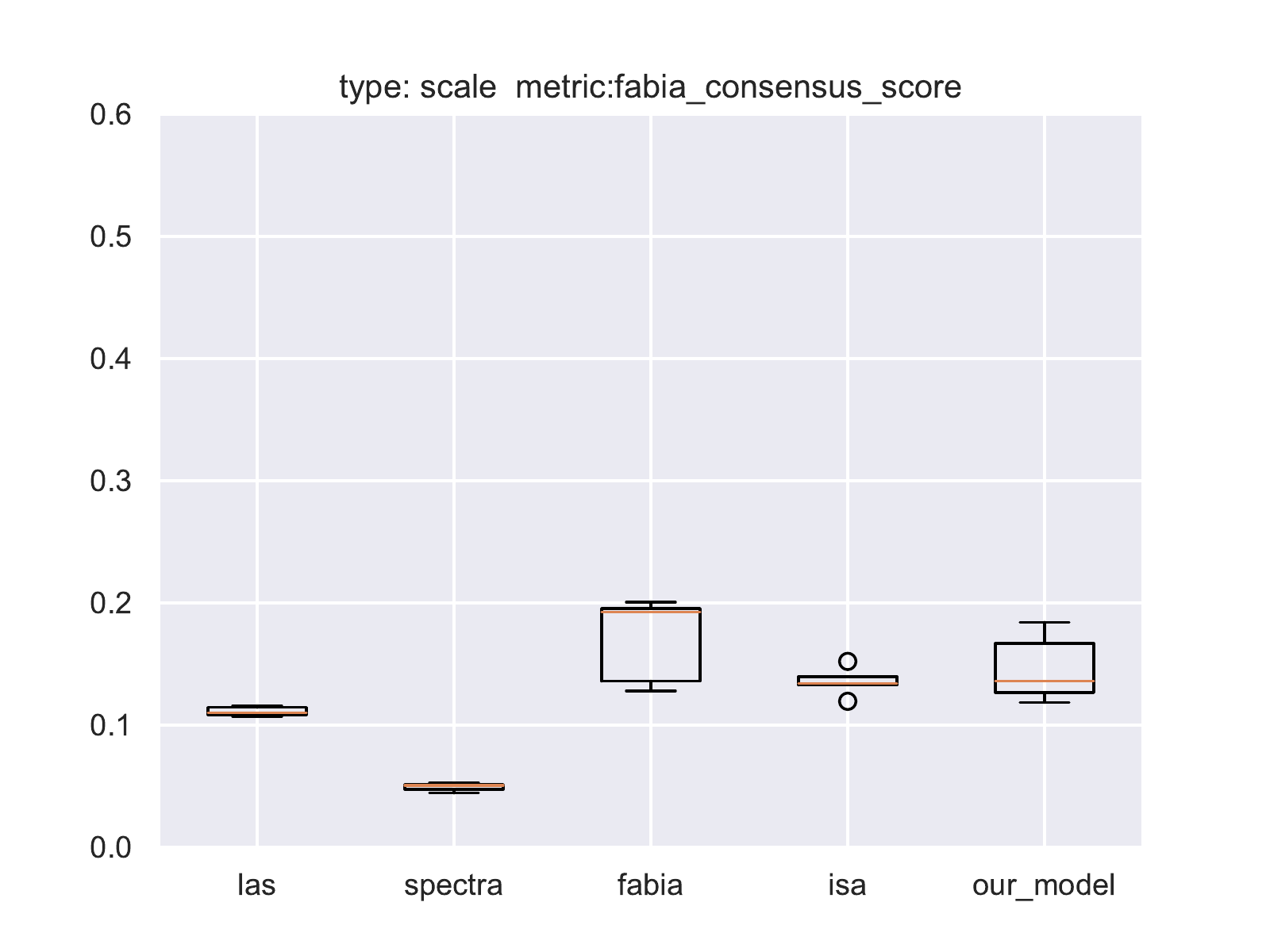}
		\label{scale5}
	}
	\subfigure{
		\includegraphics[scale=0.23]{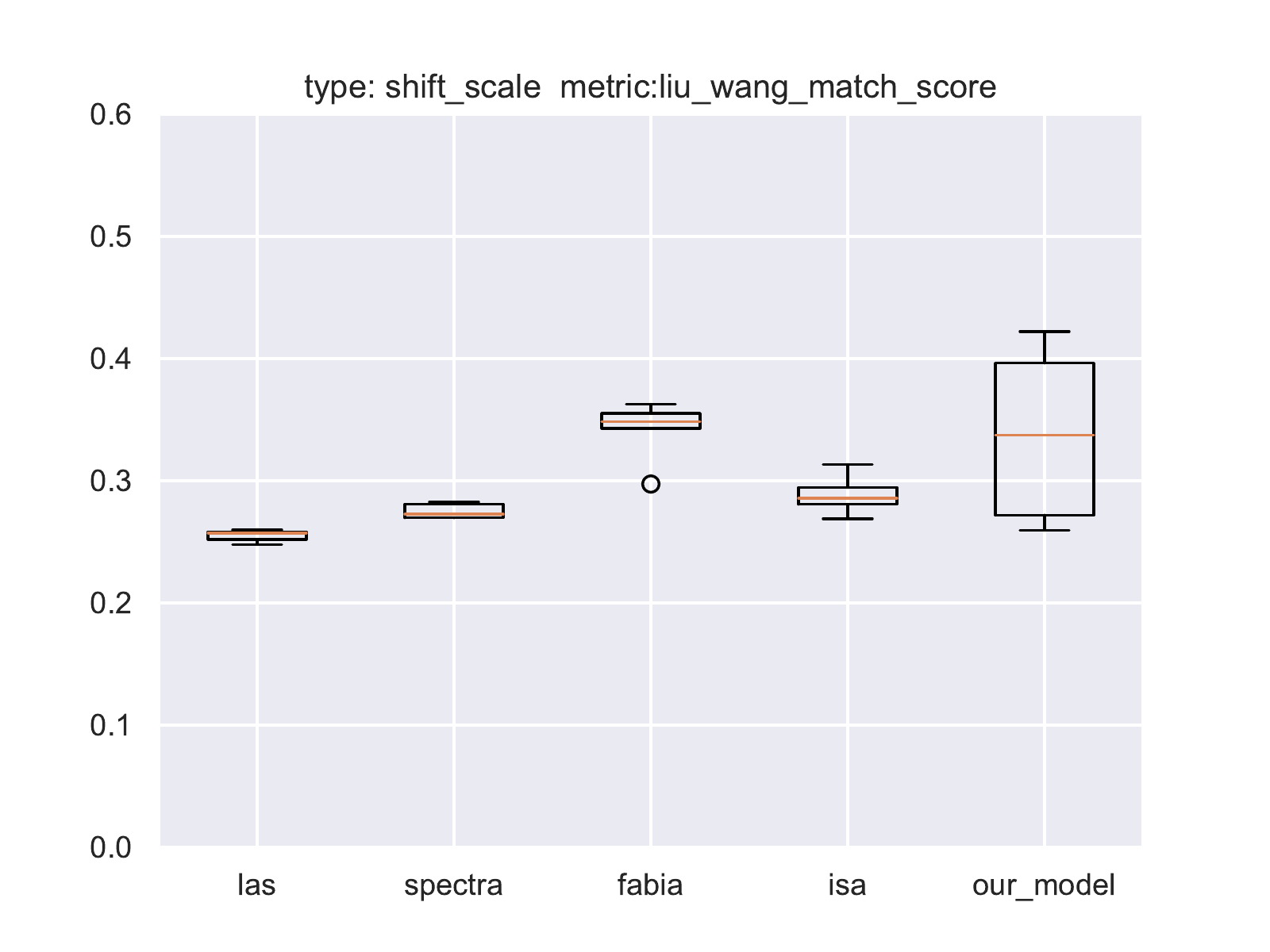}
		\label{shit_1}
	}
	\subfigure{
		\includegraphics[scale=0.23]{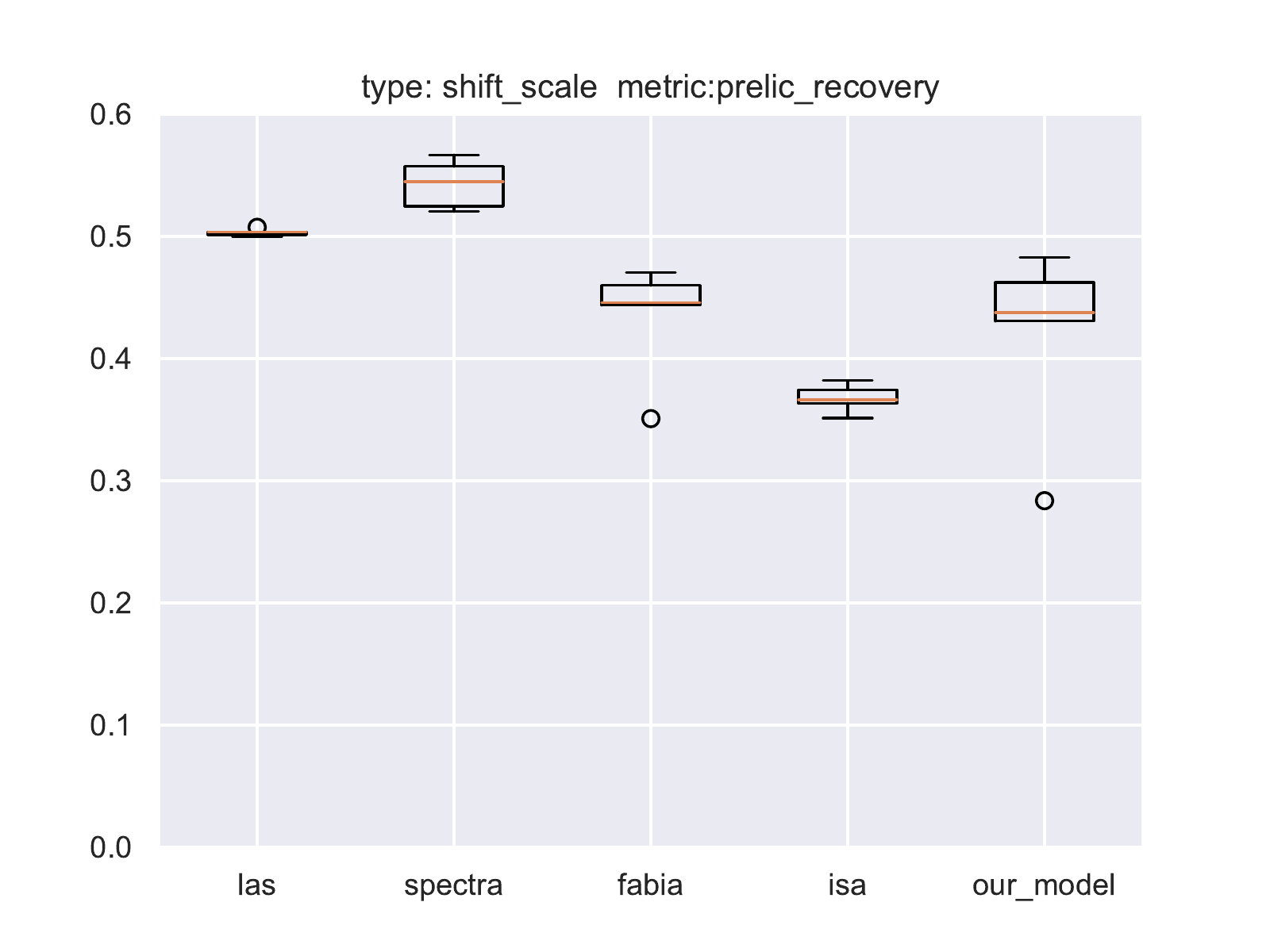}
		\label{shit_2}
	}
	\quad
	\subfigure{
		\includegraphics[scale=0.23]{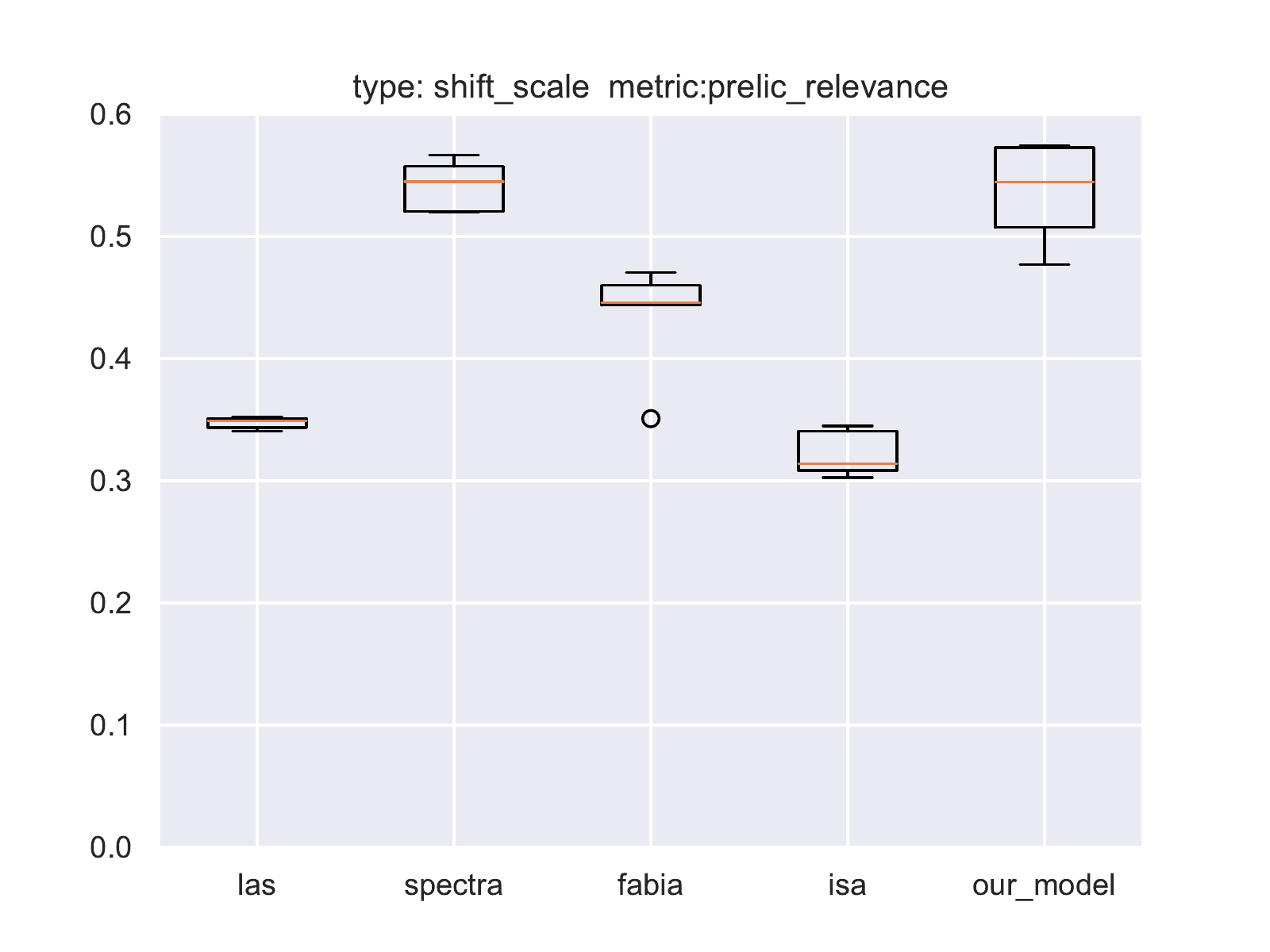}
		\label{shit_3}
	}
	\subfigure{
		\includegraphics[scale=0.23]{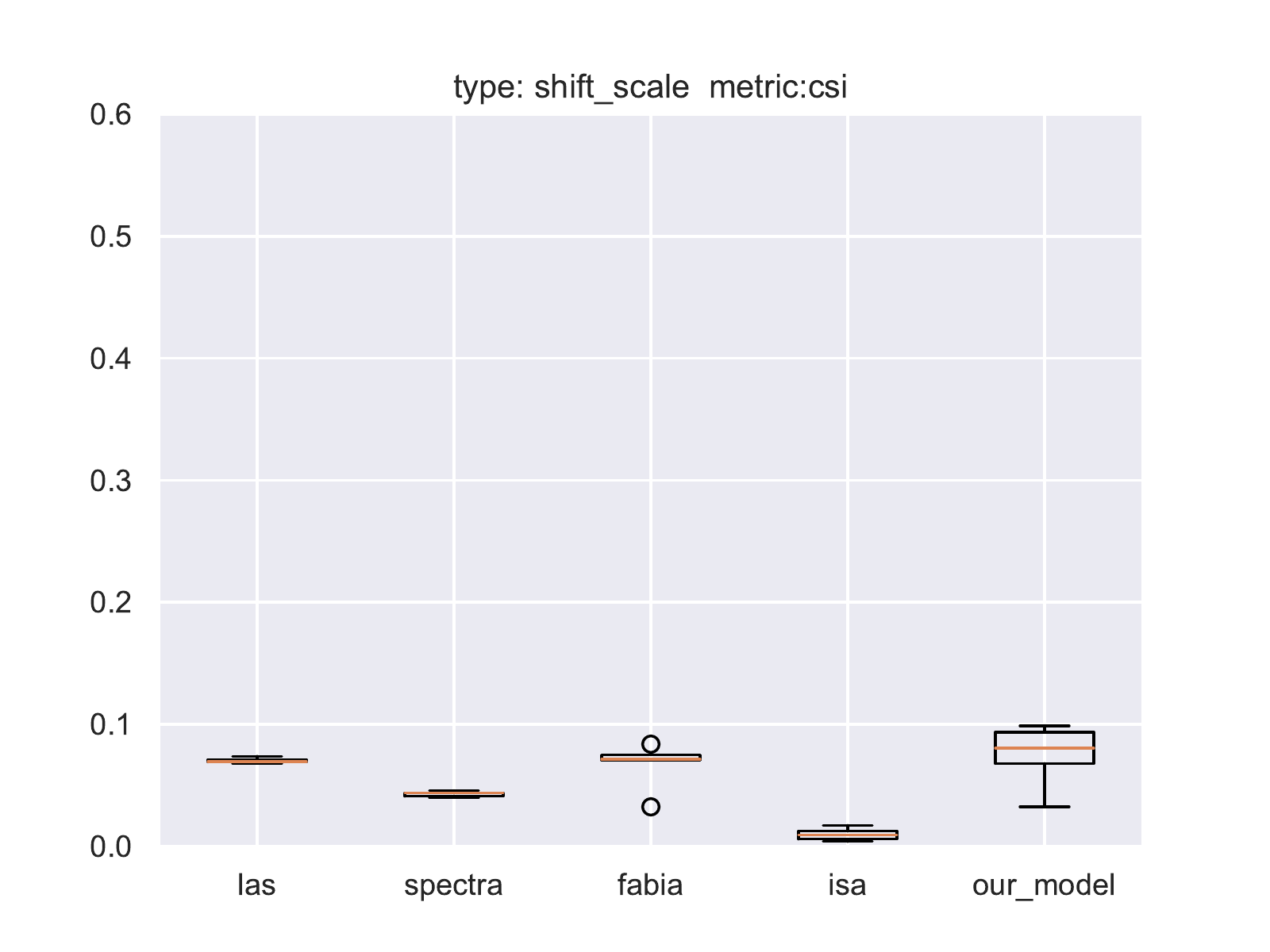}
		\label{shit_4}
	}
	\subfigure{
		\includegraphics[scale=0.23]{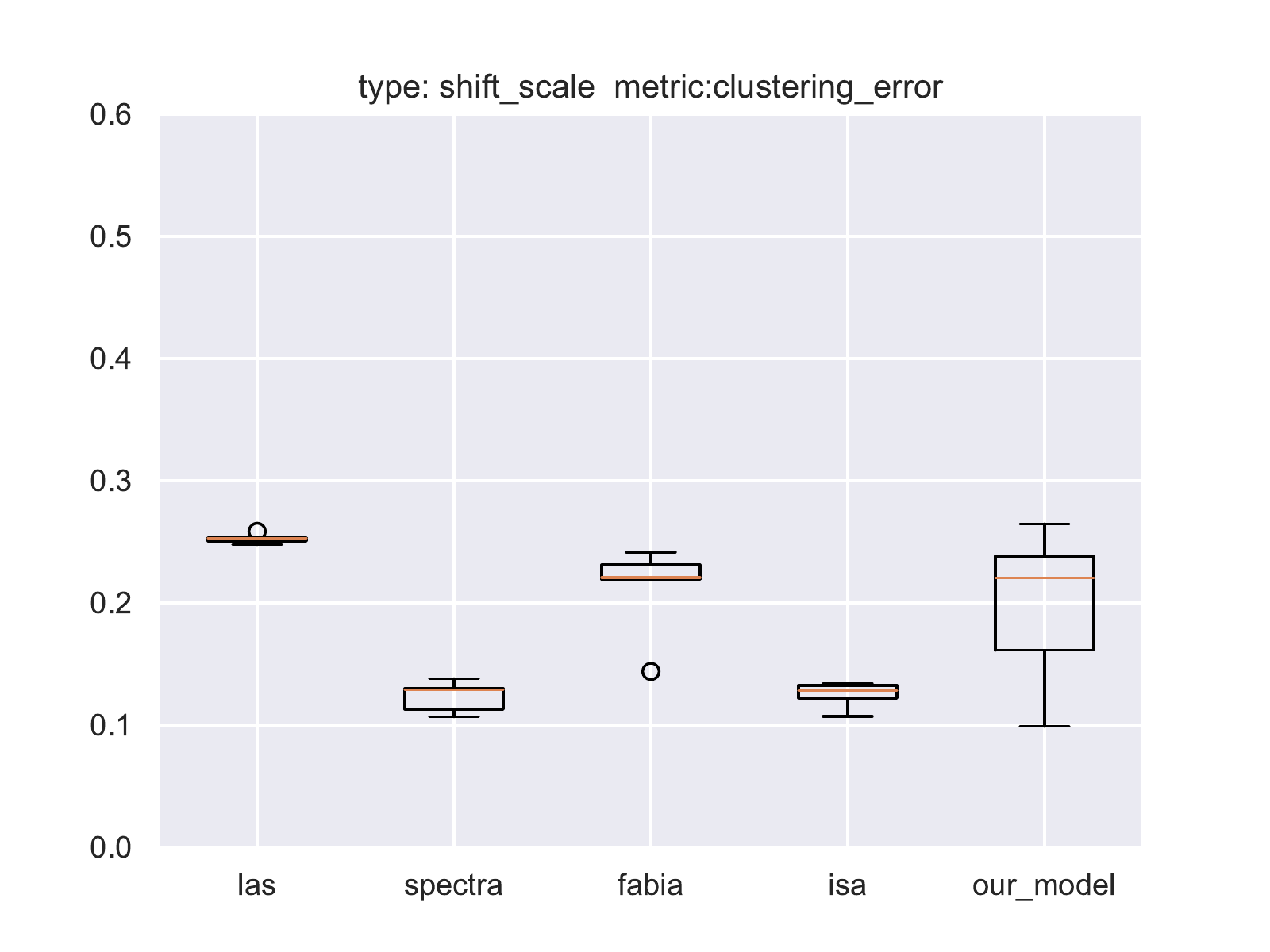}
		\label{shit_5}
	}
	\subfigure{
		\includegraphics[scale=0.23]{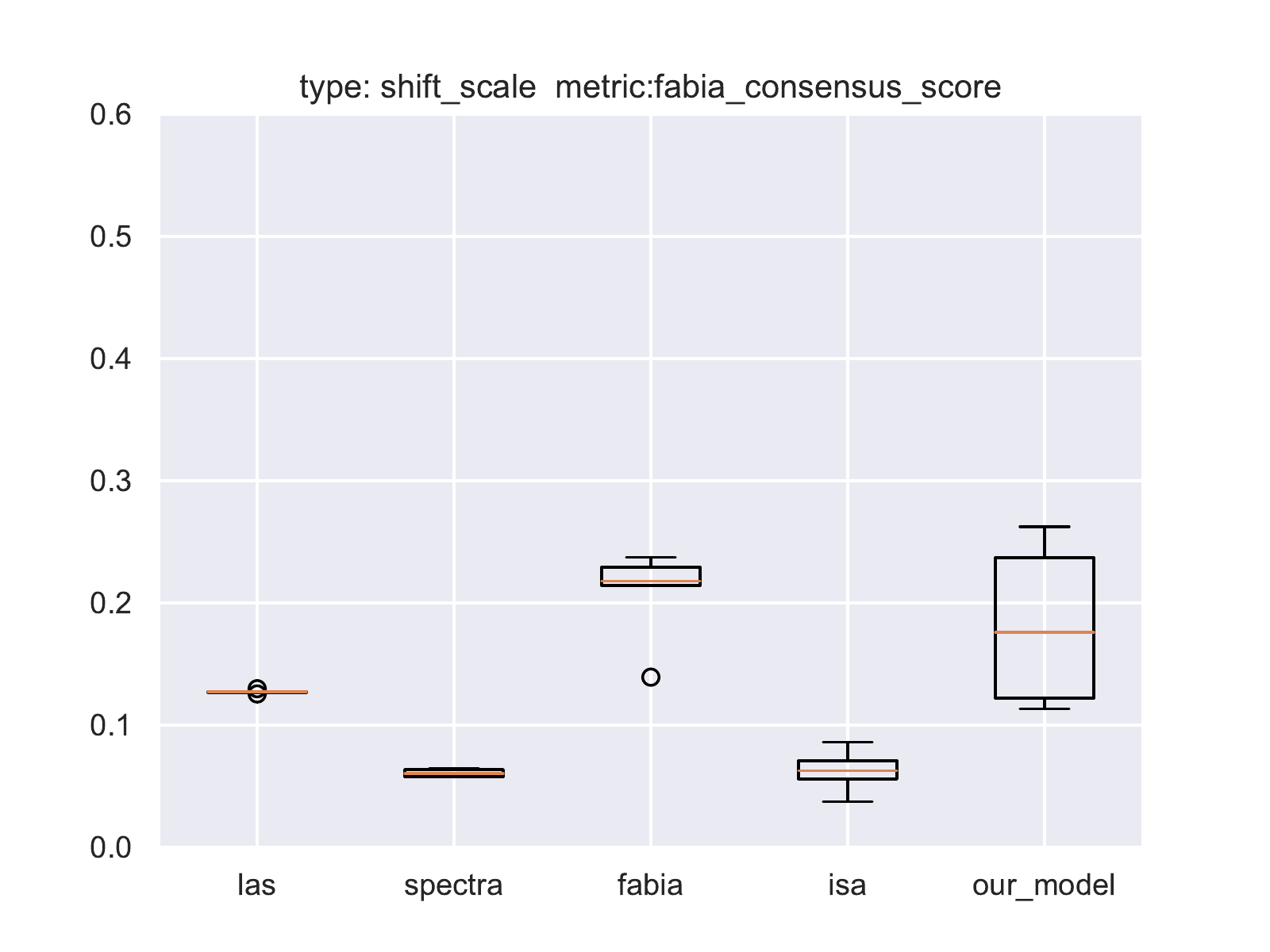}
		\label{shit_6}
	}
	\caption{The results of synthetic data experiments. We compare our framework with 4 typical biclustering algorithms on 4 synthetic data type using 6 metrics. }
	\label{fig5}
	
\end{figure*}

To be fair in the performance comparison, for each synthetic data type, we perform 5 independent runs to obtain an average test result for each method. The detailed results are presented in \refFig{fig5}. As \refFig{fig5} shows, there is no perfect algorithm that performs best on all the synthetic datasets. That is because each biclustering model is always designed to fit a specific data type. 

Specifically, for LAS method, it performs best on the constant biclusters, while it can not find shift and scale biclusters effectively. FABIA method can perform competitive with our method on shift and scale datasets. Howerver, it performs worse on constant datasets. Overall, our method can detect well biclusters on each dataset. 

\begin{figure}
	\centering
	\includegraphics[width=\linewidth]{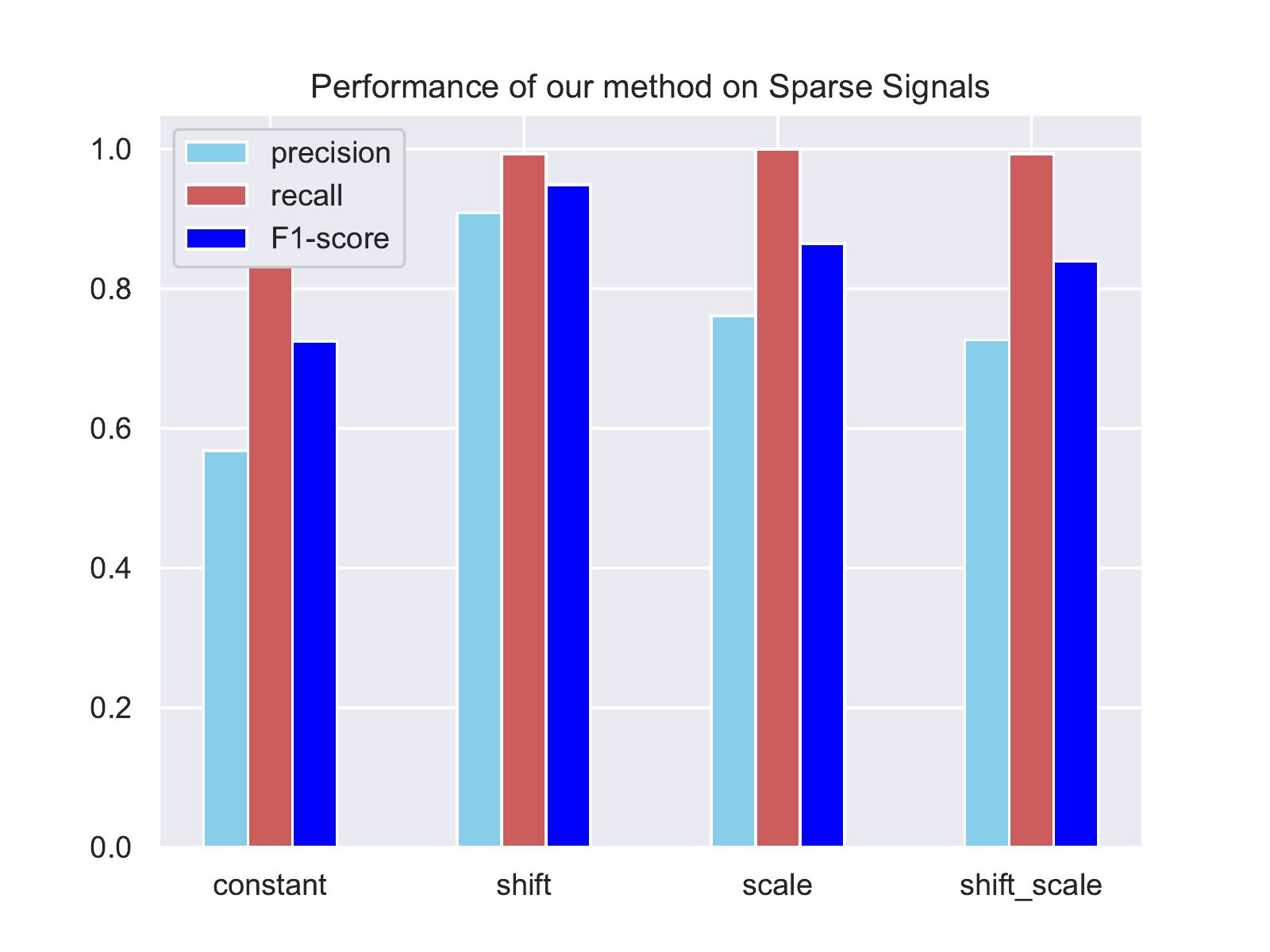}
	\caption{Performance of our framework on Sparse Signals.} 
	\label{fig4}
\end{figure}

We also test the ability of our method to identify sparse signals. \refFig{fig4} shows the performance of our method on 4 sythetic datasets. Our method can achieve great F1-scores. Note that the recall is almost 1 no matter what type the data is. It guarantees that every sparse signal will not be missed, meaning that each student's indivial characteristics will not be ignored in practice. 

\begin{figure}
	\centering
	\subfigure{
		\includegraphics [width=65pt]{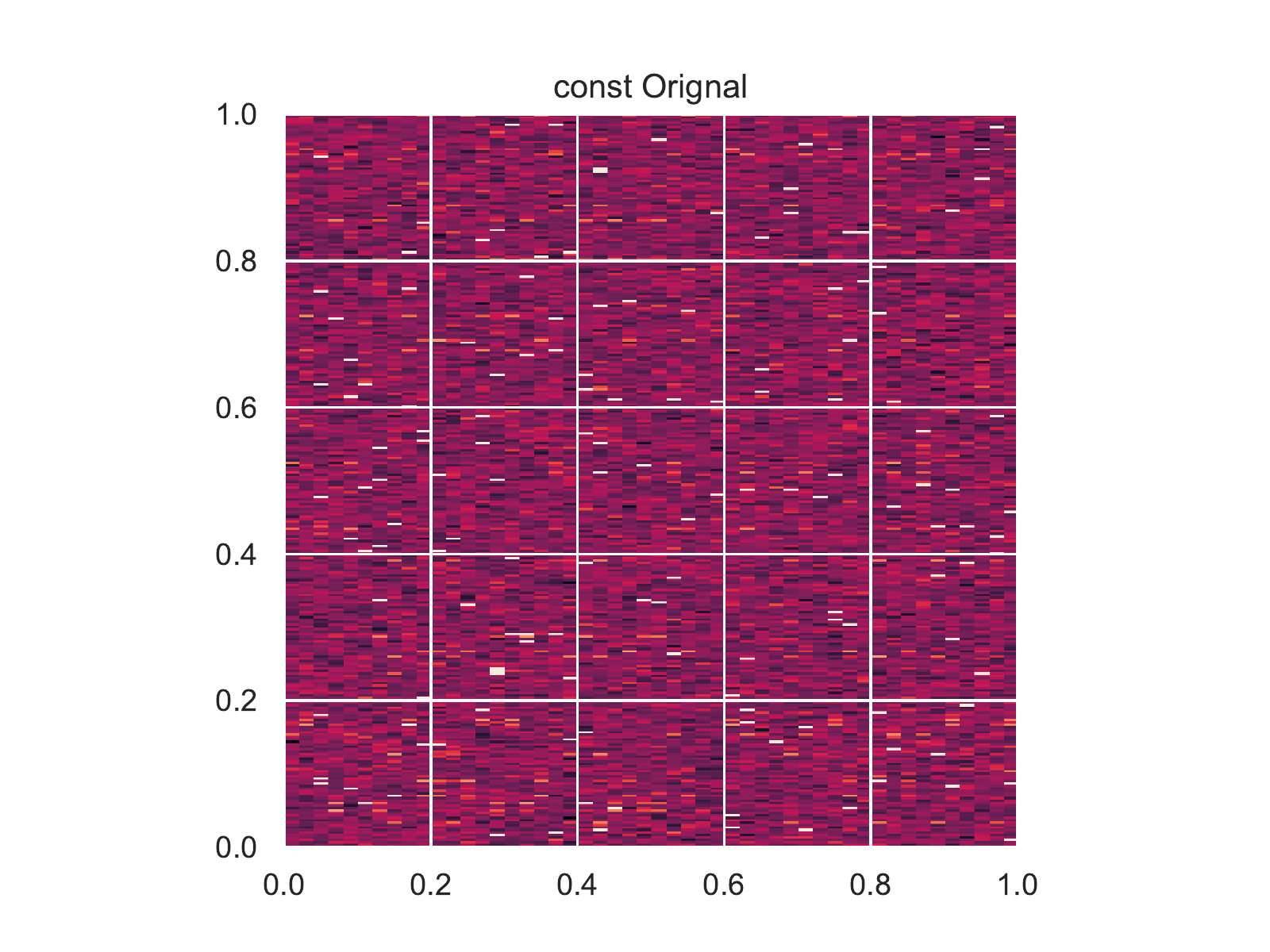}
		\label{const_1}
	}
	\subfigure{
		\includegraphics [width=65pt]{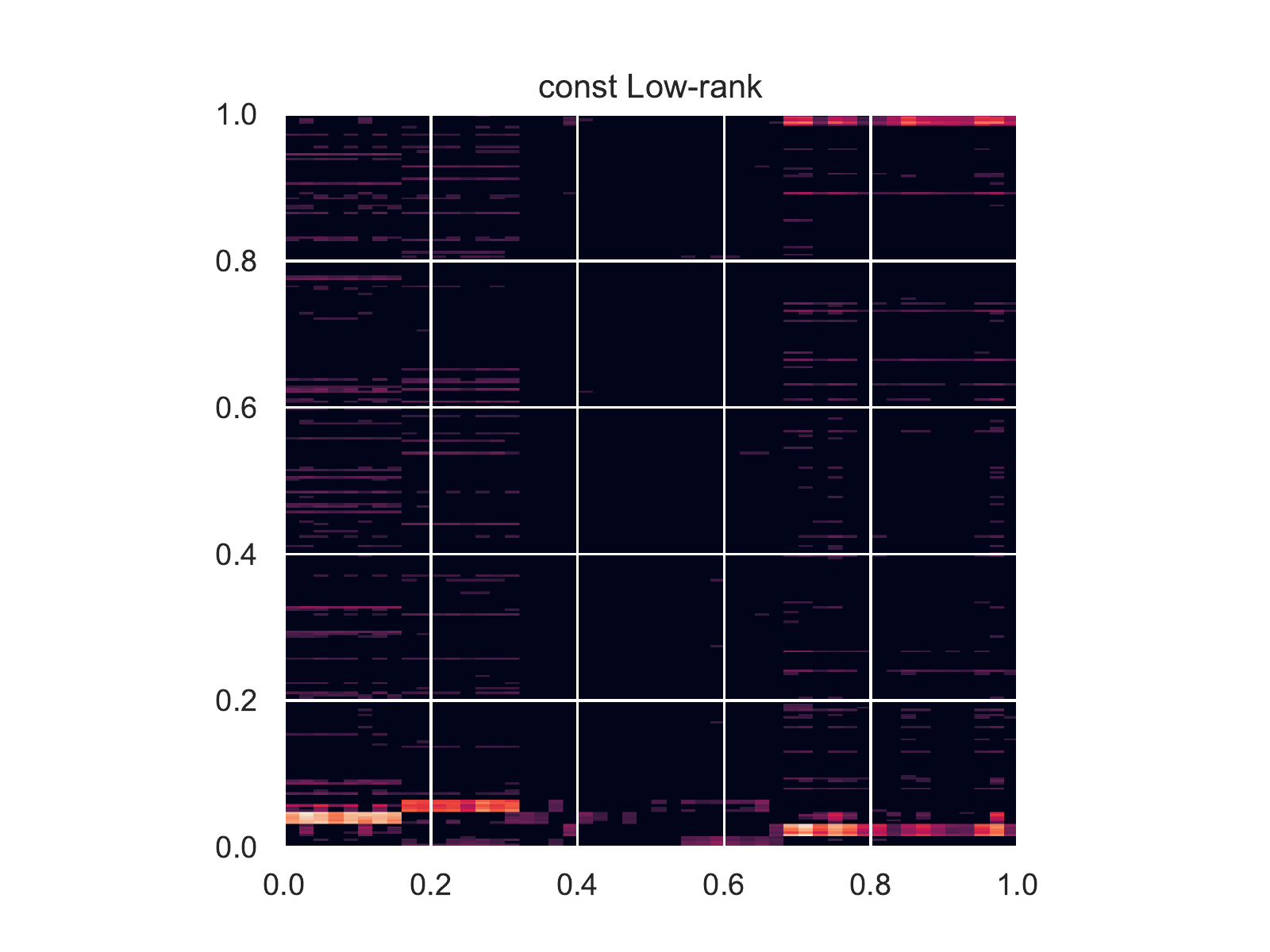}
		\label{const_2}
	}
	\subfigure{
		\includegraphics [width=65pt]{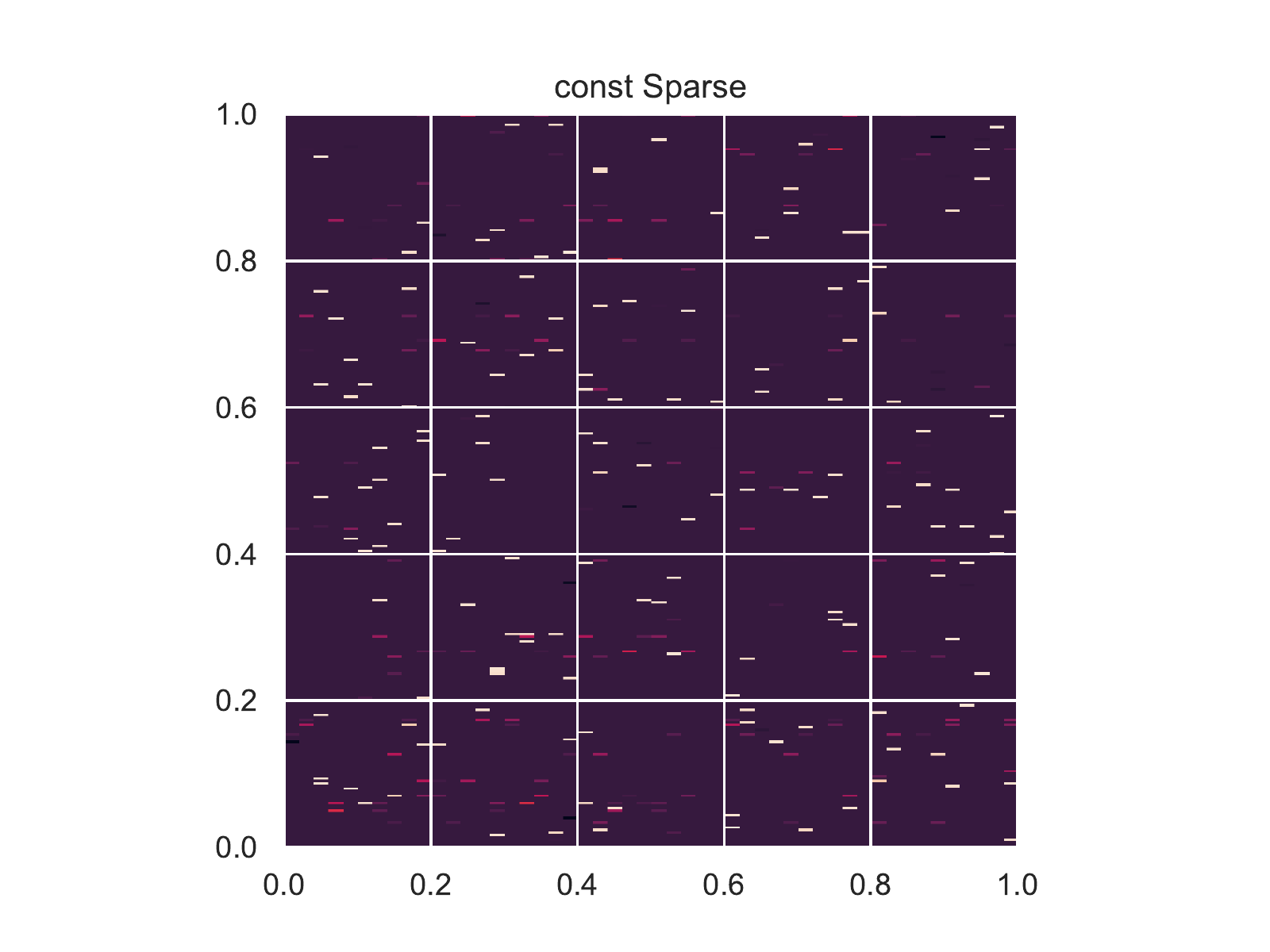}
		\label{const_3}
	}
	\quad
	\subfigure{
		\includegraphics[width=65pt]{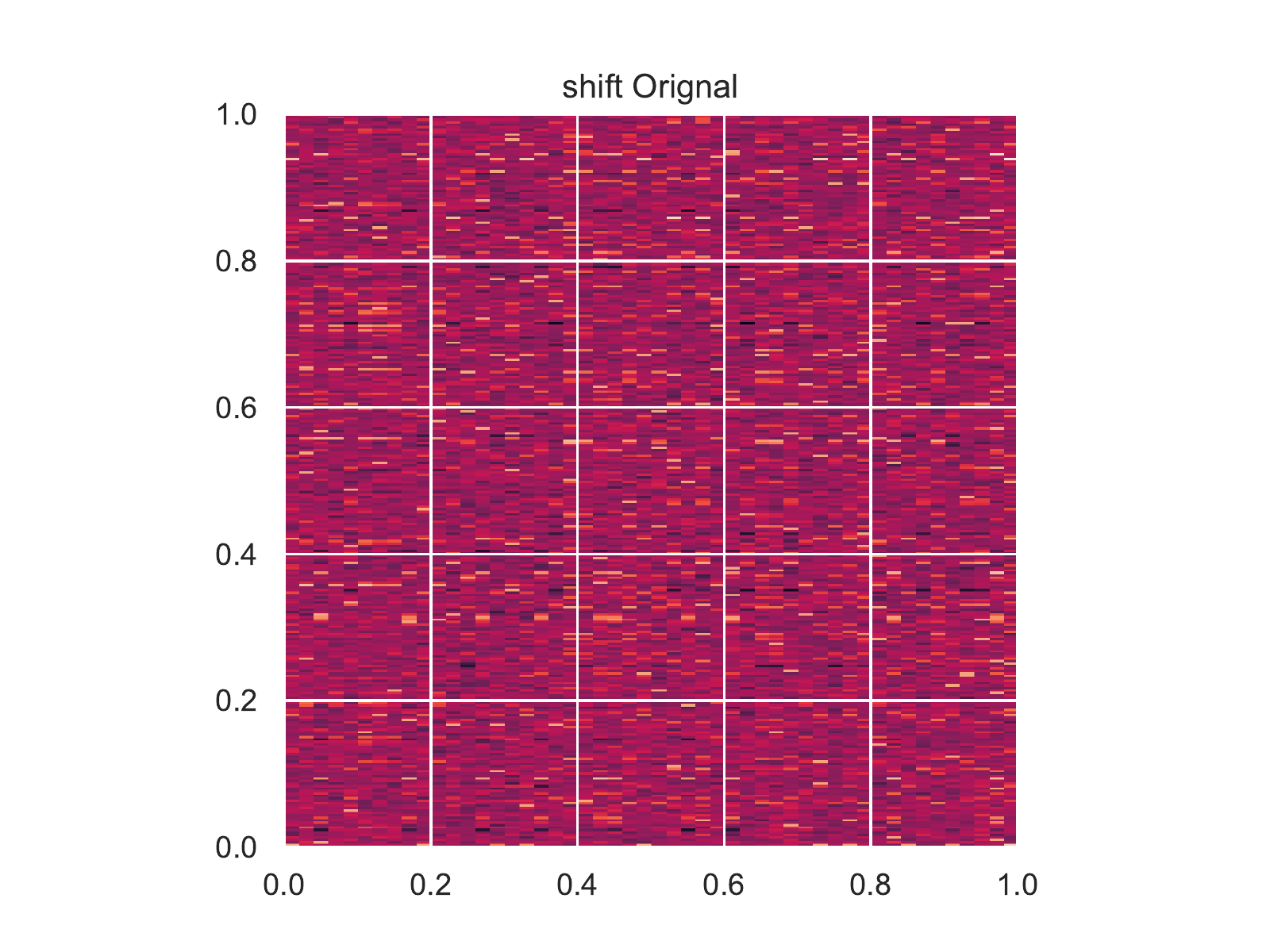}
		\label{const_1}
	}
	\subfigure{
		\includegraphics[width=65pt]{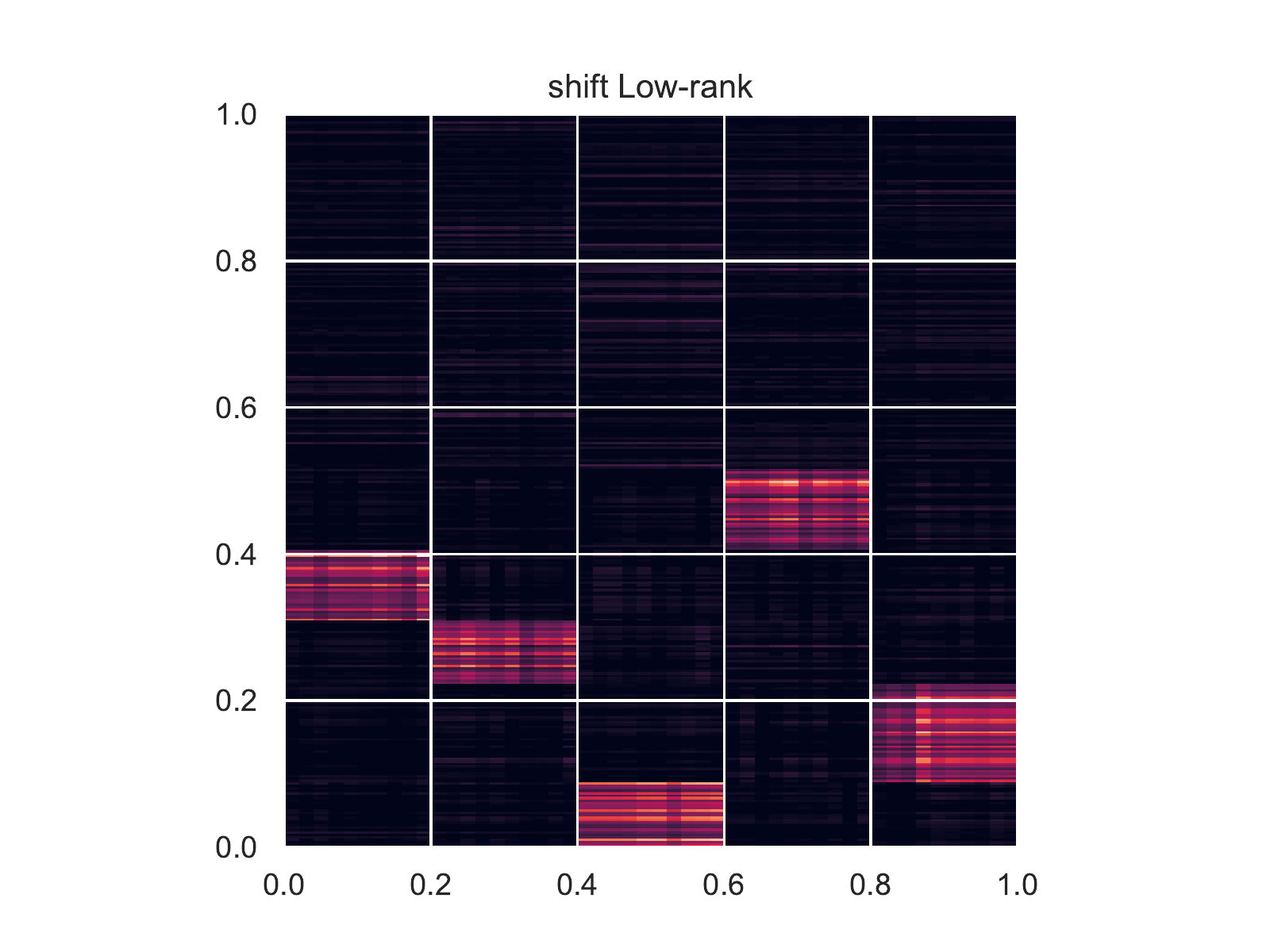}
		\label{const_2}
	}
	\subfigure{
		\includegraphics[width=65pt]{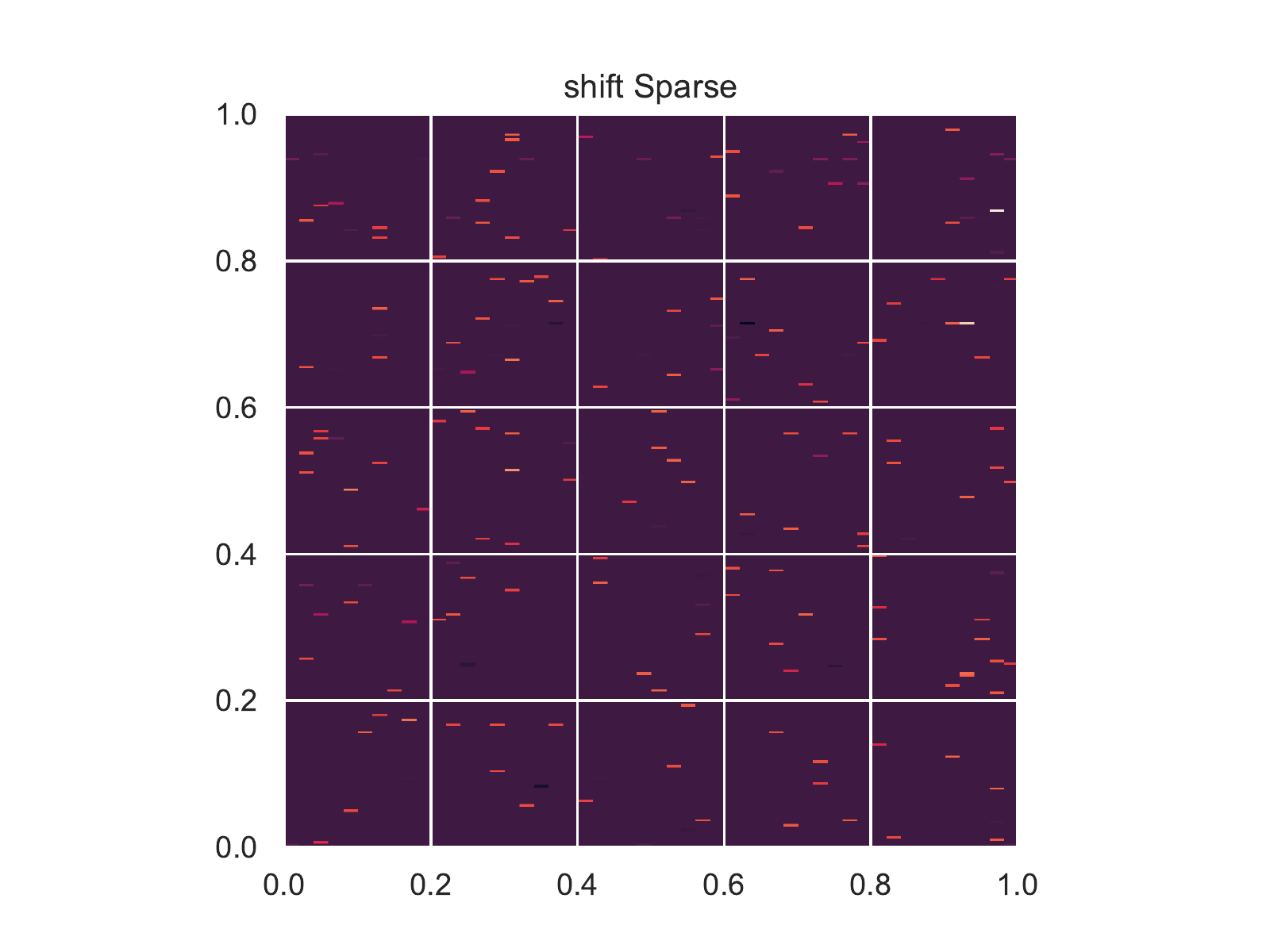}
		\label{const_3}
	}
	\quad
	\subfigure{
		\includegraphics[width=65pt]{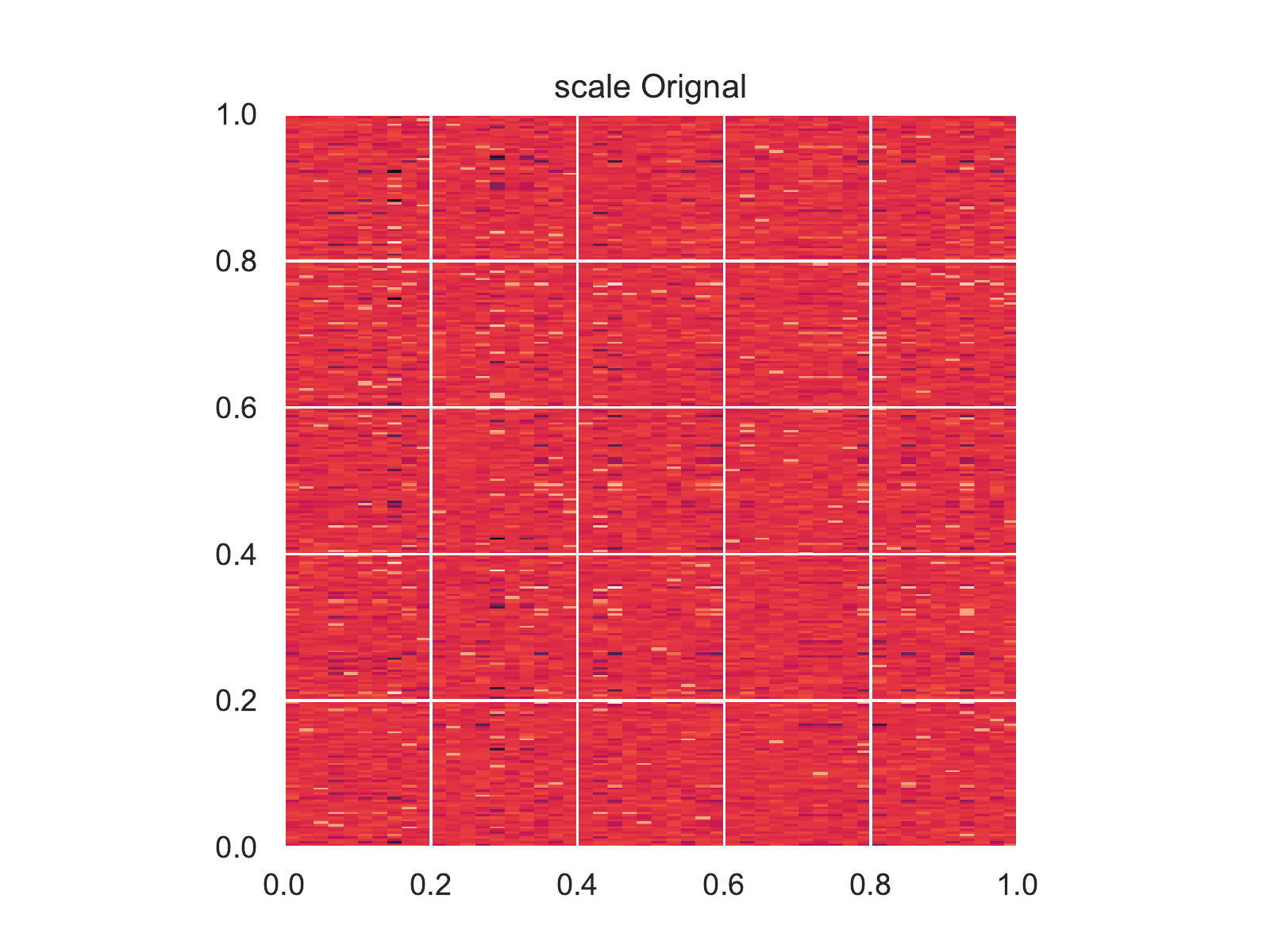}
		\label{const_1}
	}
	\subfigure{
		\includegraphics[width=65pt]{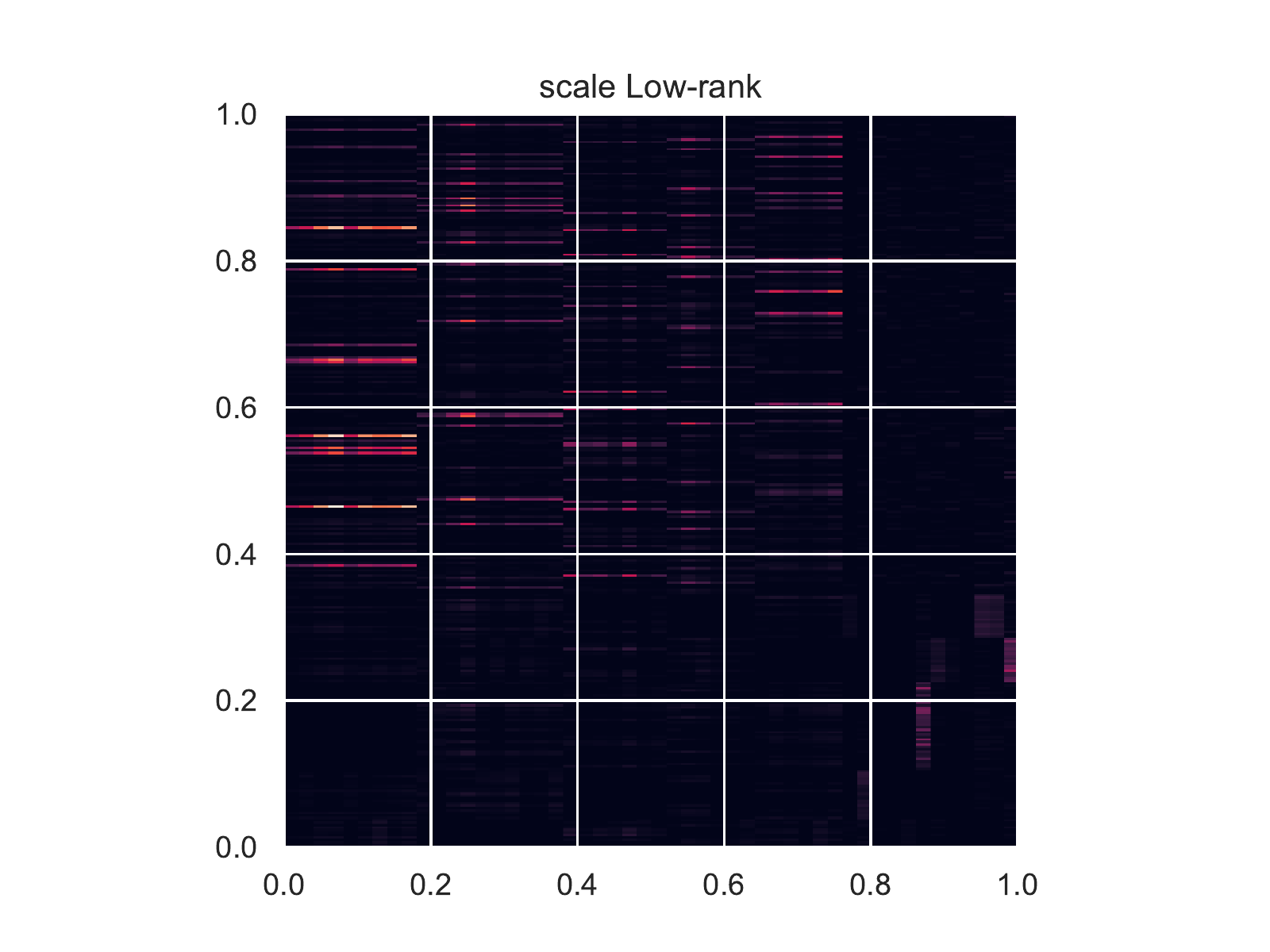}
		\label{const_2}
	}
	\subfigure{
		\includegraphics[width=65pt]{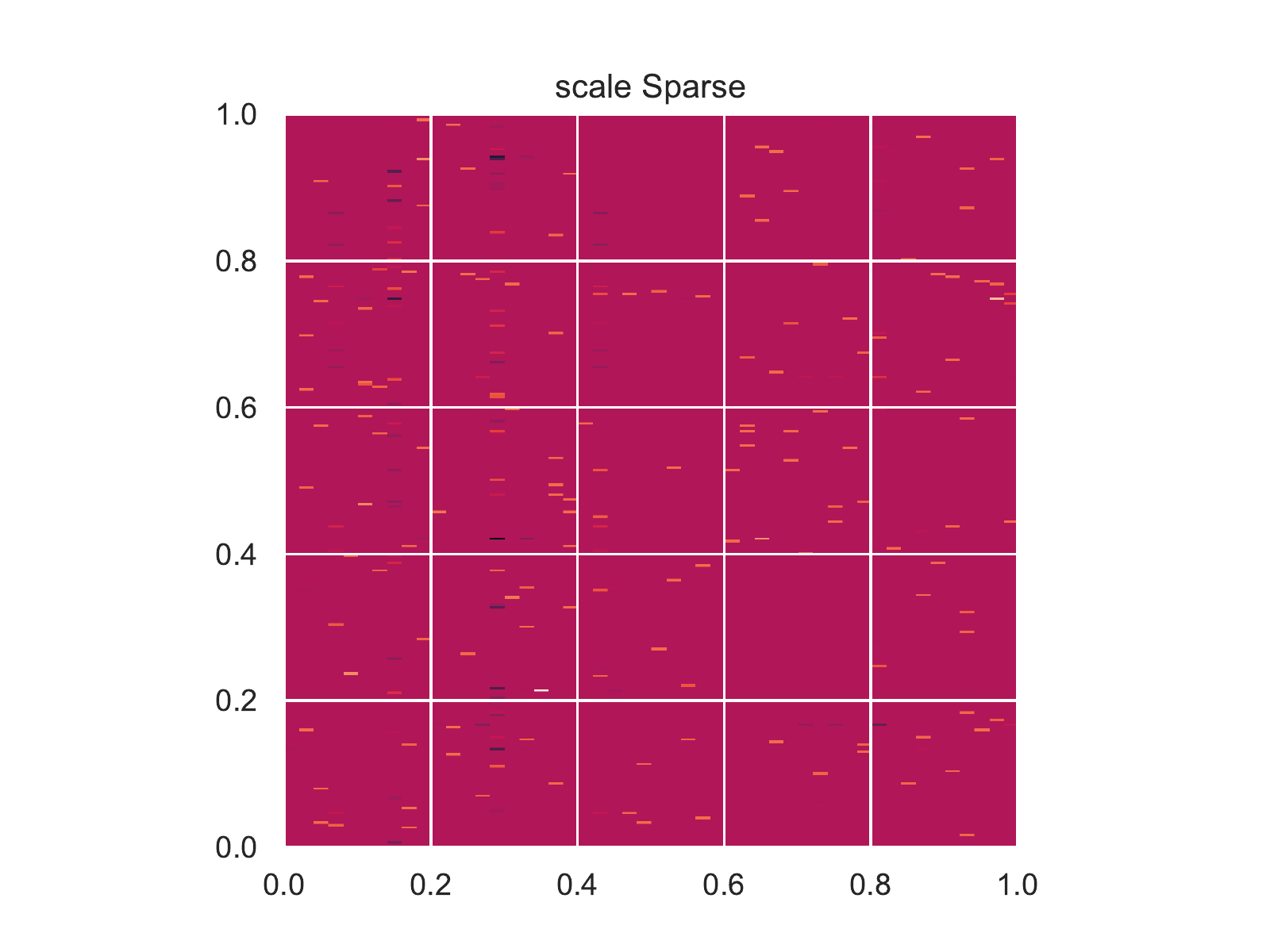}
		\label{const_3}
	}
	\quad
	\subfigure{
		\includegraphics[width=65pt]{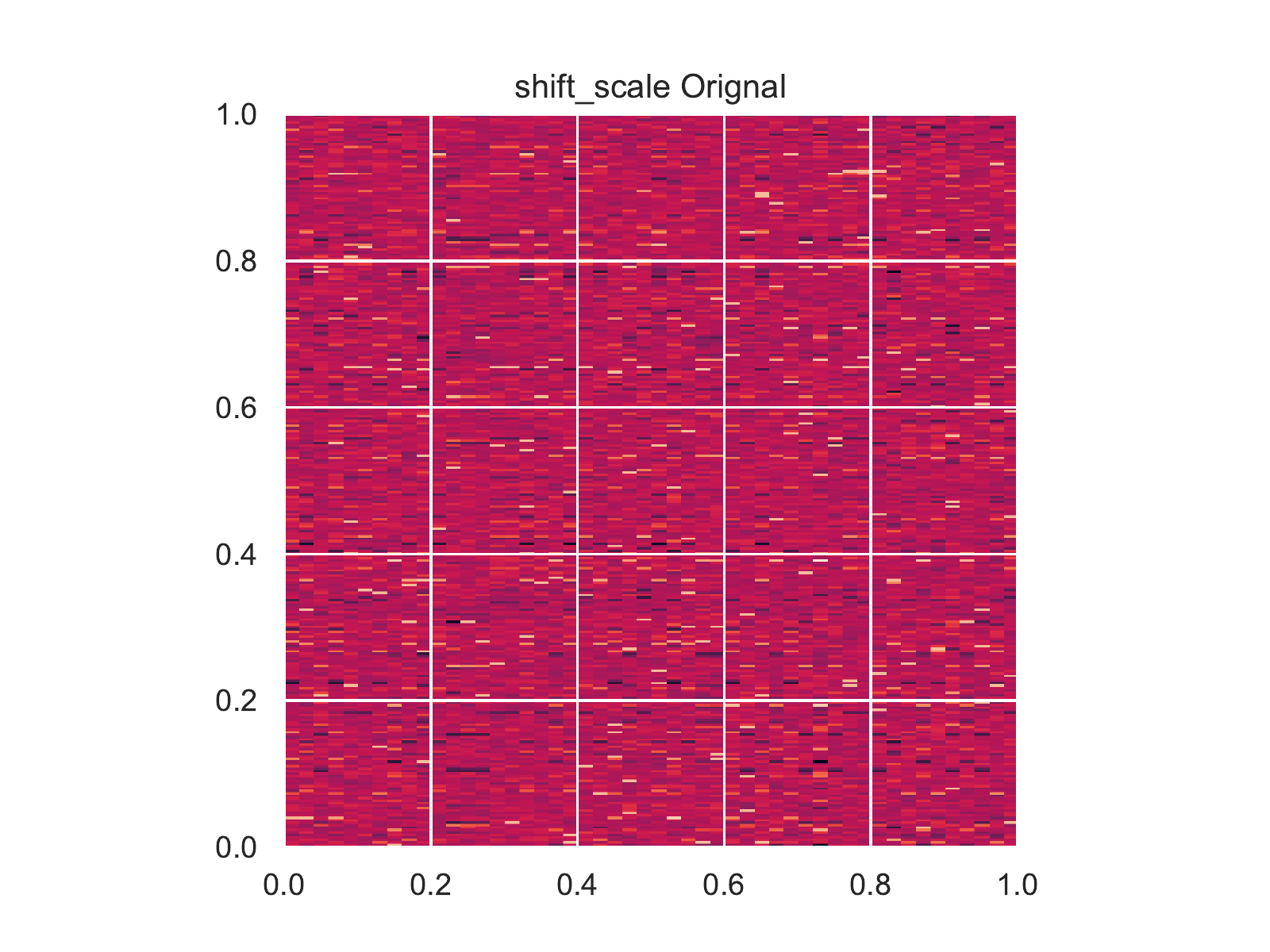}
		\label{const_1}
	}
	\subfigure{
		\includegraphics[width=65pt]{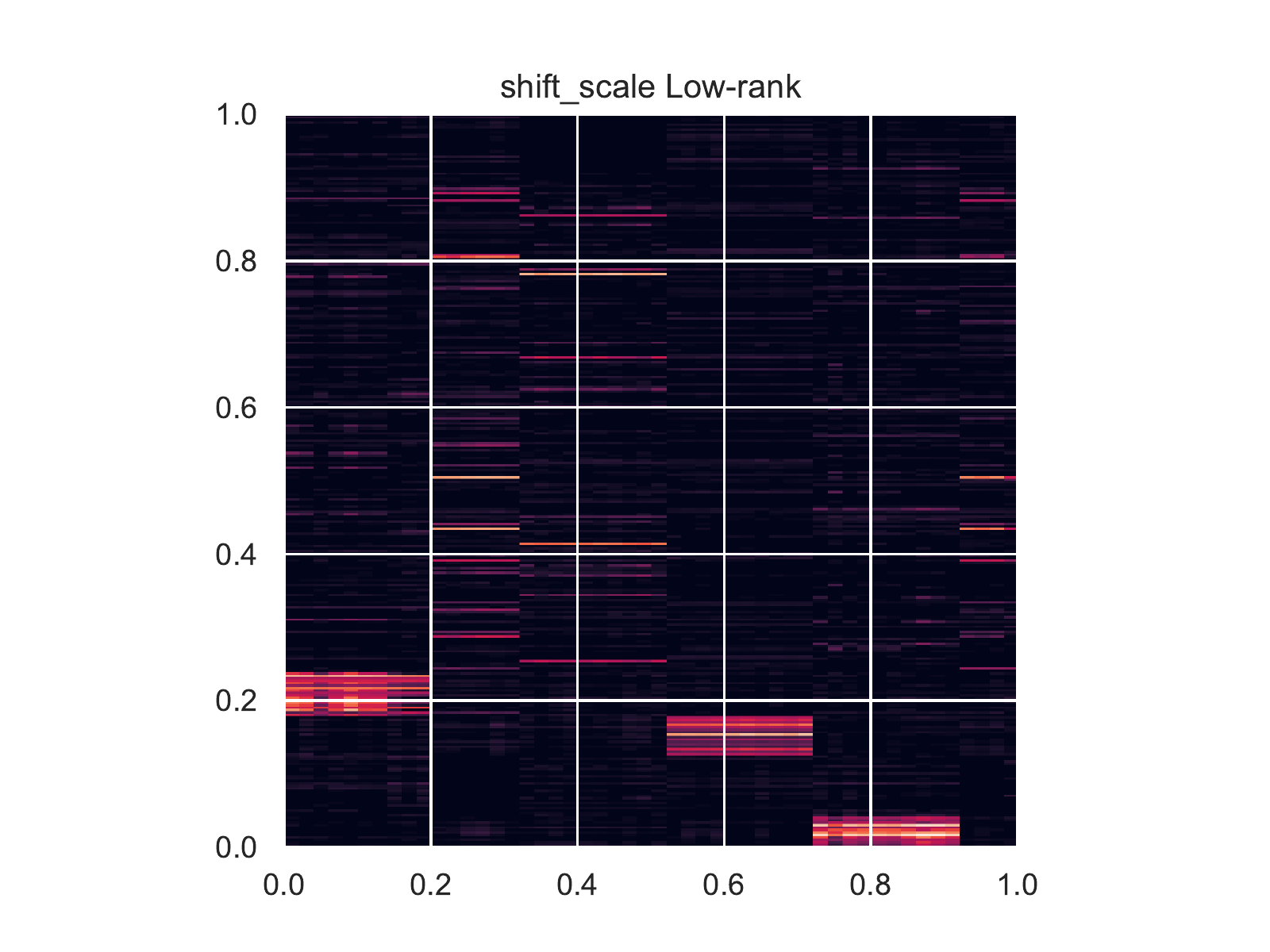}
		\label{const_2}
	}
	\subfigure{
		\includegraphics[width=65pt]{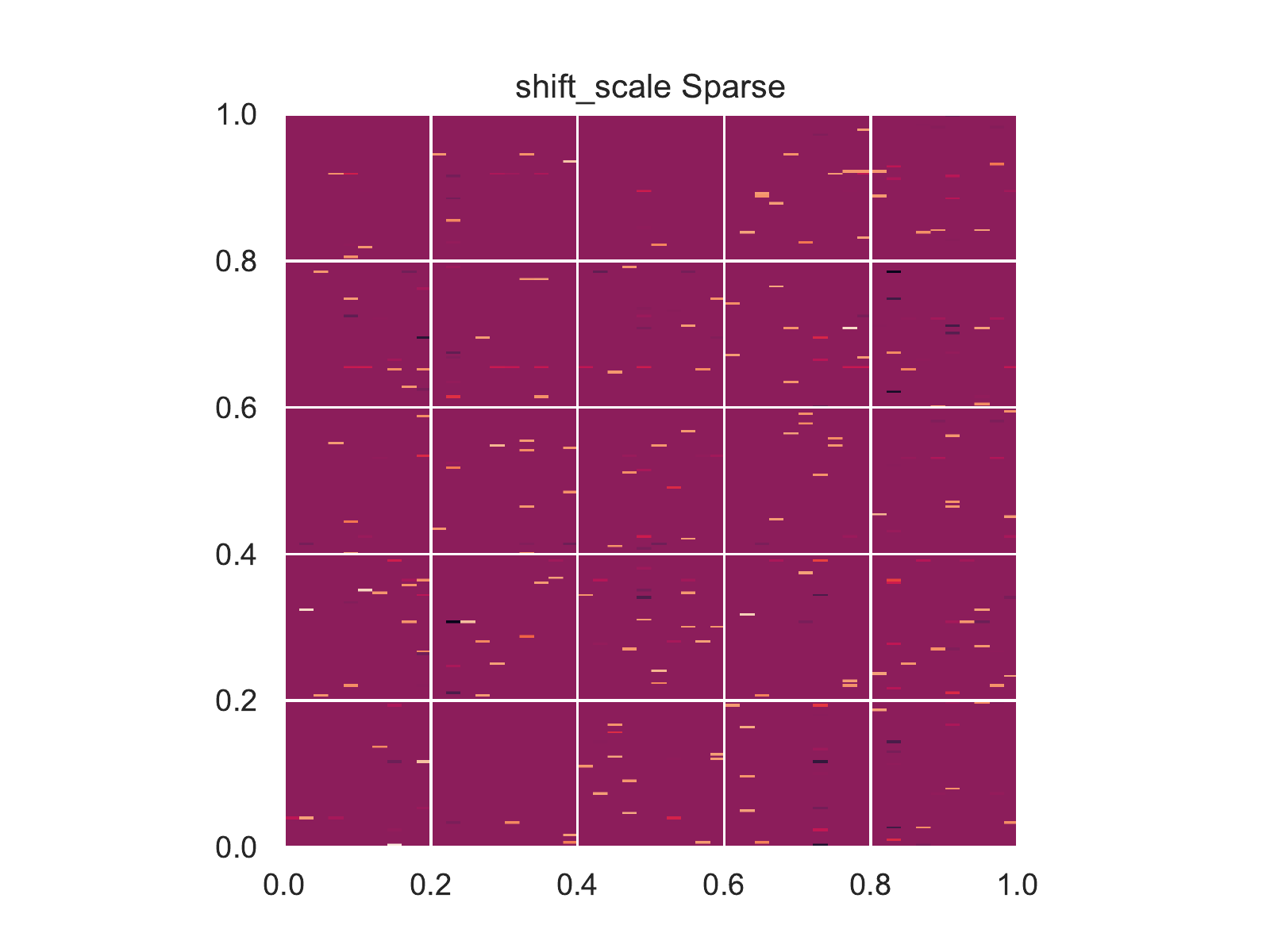}
		\label{const_3}
	}
	\caption{Illustrations of the results on synthetic data. The left ones are the original input matrix. The middle and the right ones are the results recovered by our framework.}
	\label{fig3}
	
\end{figure}

In \refFig{fig3}, we present illustrations of the orignal matrix, low-rank matrix and sparse matrix recovered by our method for each dataset. 

\subsection{Experiments with real data}
Since synthetic data can only reflect certain aspects of reality, we also perform our framework on two real datasets to
test the effectiveness. In this subsection, we will first descripe the detail of the two datasets and then present the performances of our method.

\subsubsection{Datasets and Preprocessing}
We collected two datasets of students' performance: ADS data \cite{DBLP:conf/its/HenriquesFC19} and MATH data.
The ADS dataset is provided by the Department of Informatics of the Pontifical Cathaolic University of Rio de Janero (PUC-Rio), which captures the performance of students along the topics of the Advanced Data Structures (ADS) course. The MATH dataset is collected from AIXUEXI Education Group LTD, which contains 76 students in Junior High School Grade 1 and 54 knowledge topics.

We do some preprocessings on the two datasets. For ADS data, as the goal is to detect the weaknesses of students, we first use 100 minus the performance scores and divide the resulted scores into 10 levels where 0 denotes an excelling grade and 9 
a low grade. For MATH data, we first use the DKT method \cite{NIPS2015_5654} to get the mastery matrix and then use 1 minus the elements in the matrix as the input data.

\subsubsection{Results}
\begin{figure}[htbp]
	\centering
	\includegraphics[width=78pt]{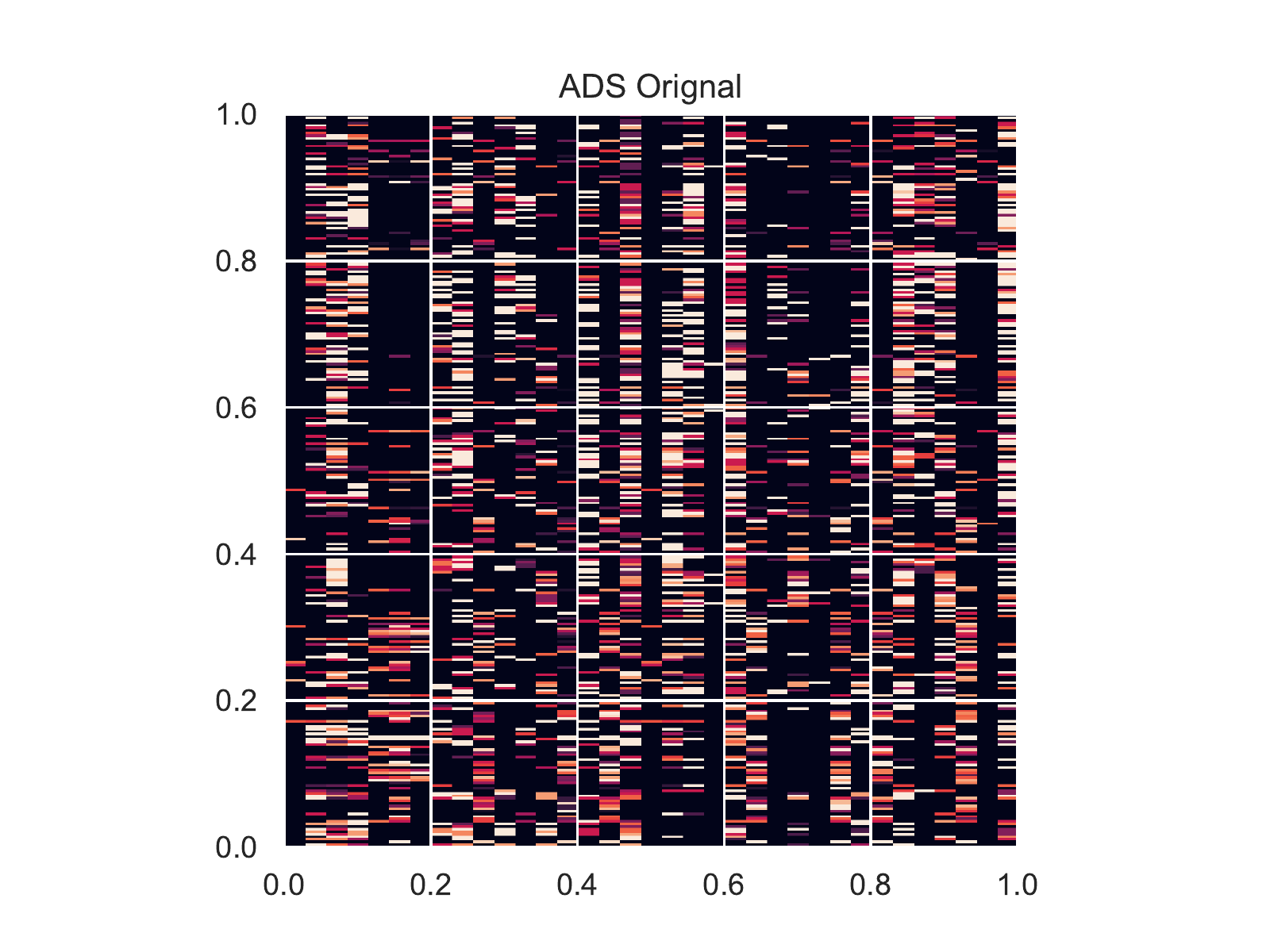}
	\includegraphics[width=78pt]{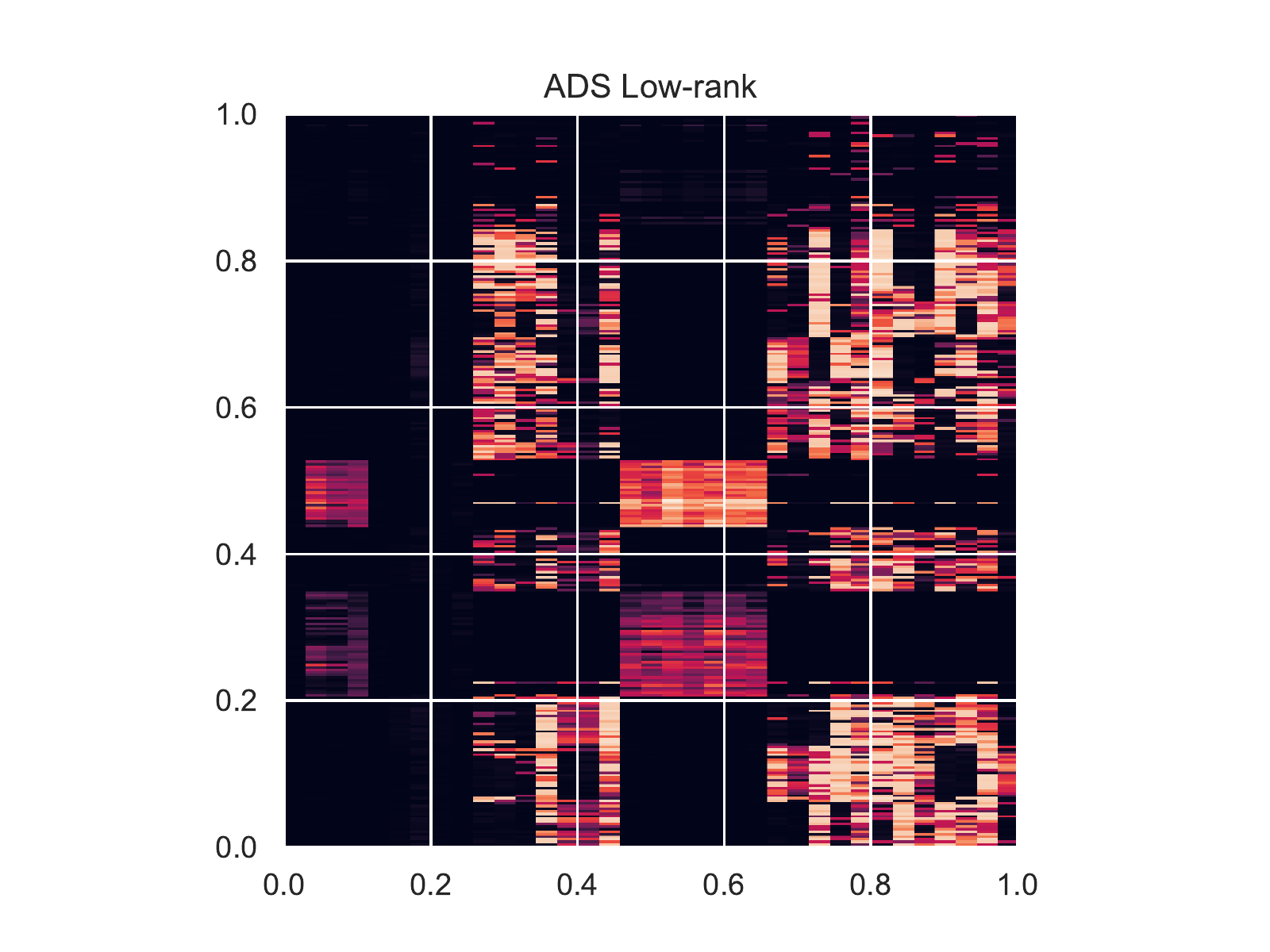}
	\includegraphics[width=78pt]{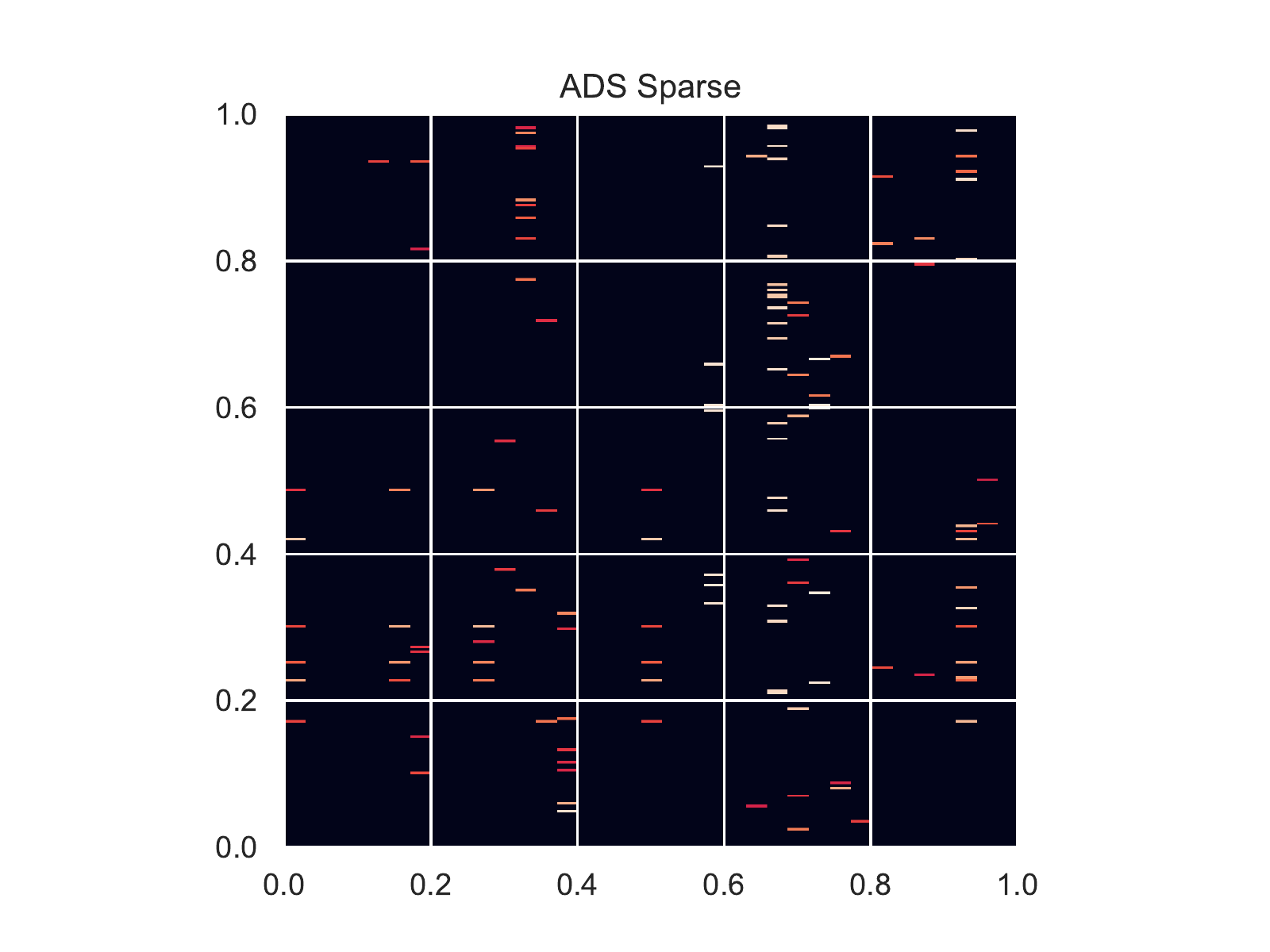}
	\caption{The results of ADS data experiments.}
	\label{fig6}
\end{figure}

\begin{figure}[htbp]
	\centering
	\includegraphics[width=78pt]{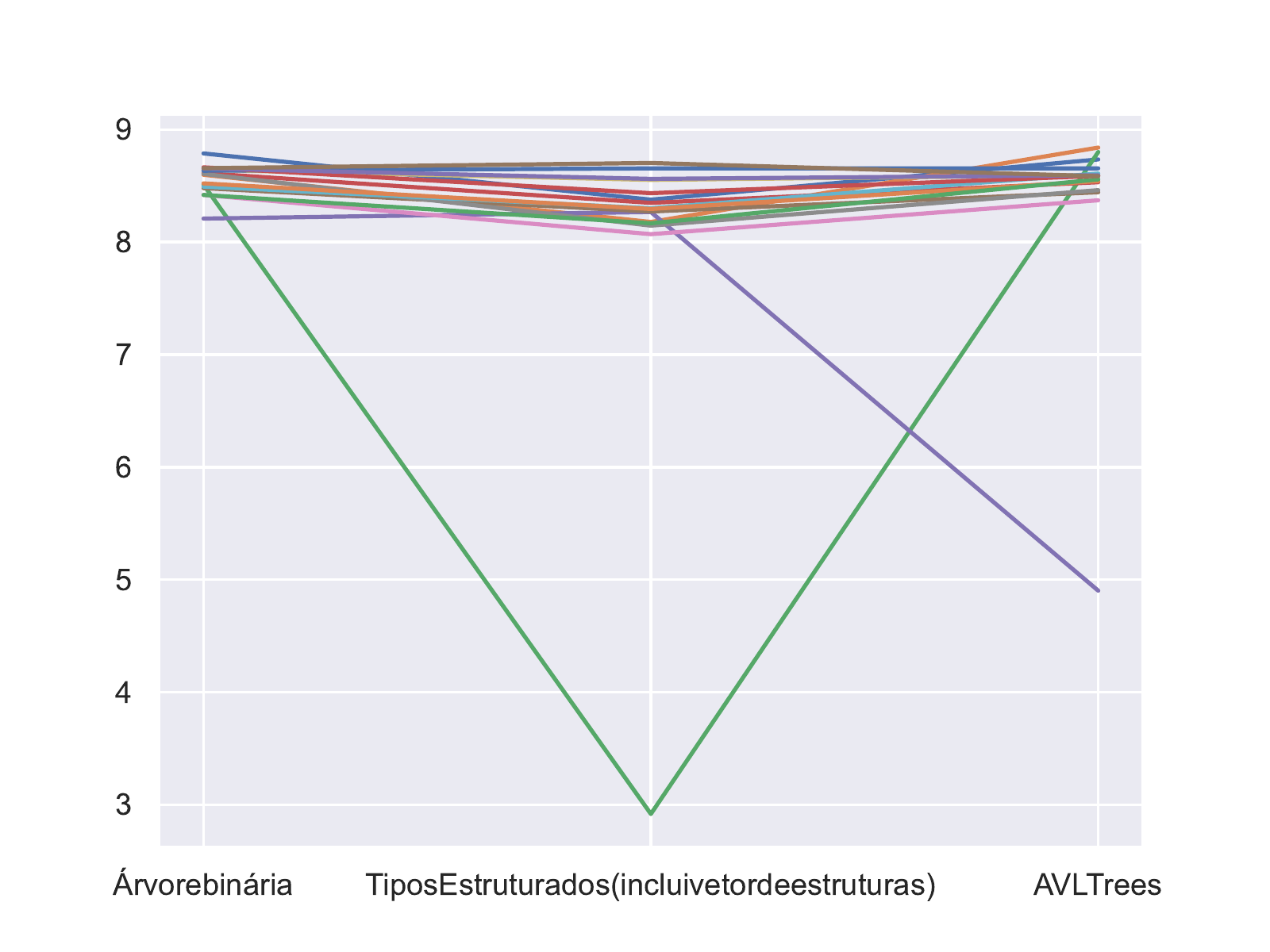}
	\includegraphics[width=78pt]{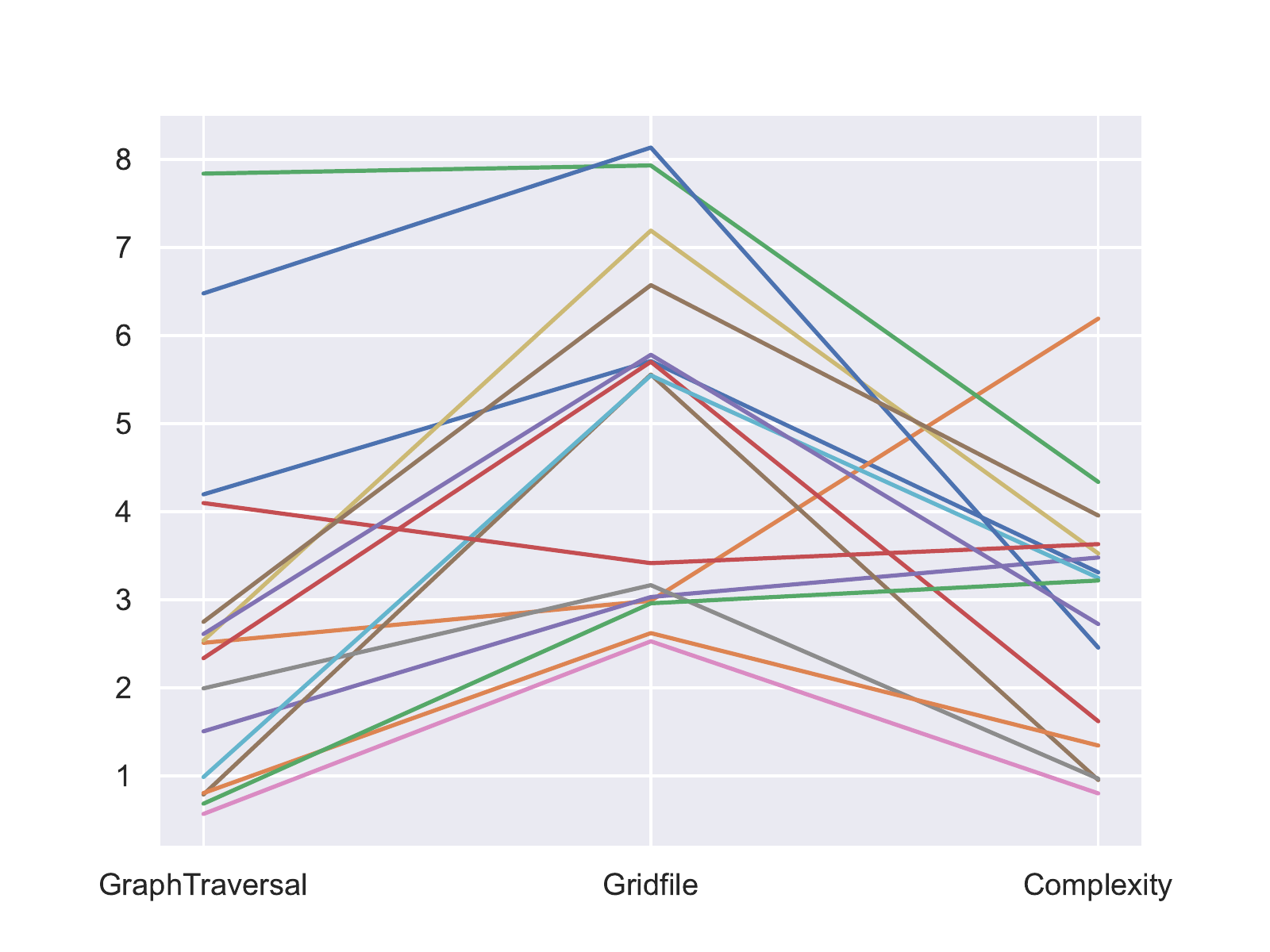}
	
	\includegraphics[width=78pt]{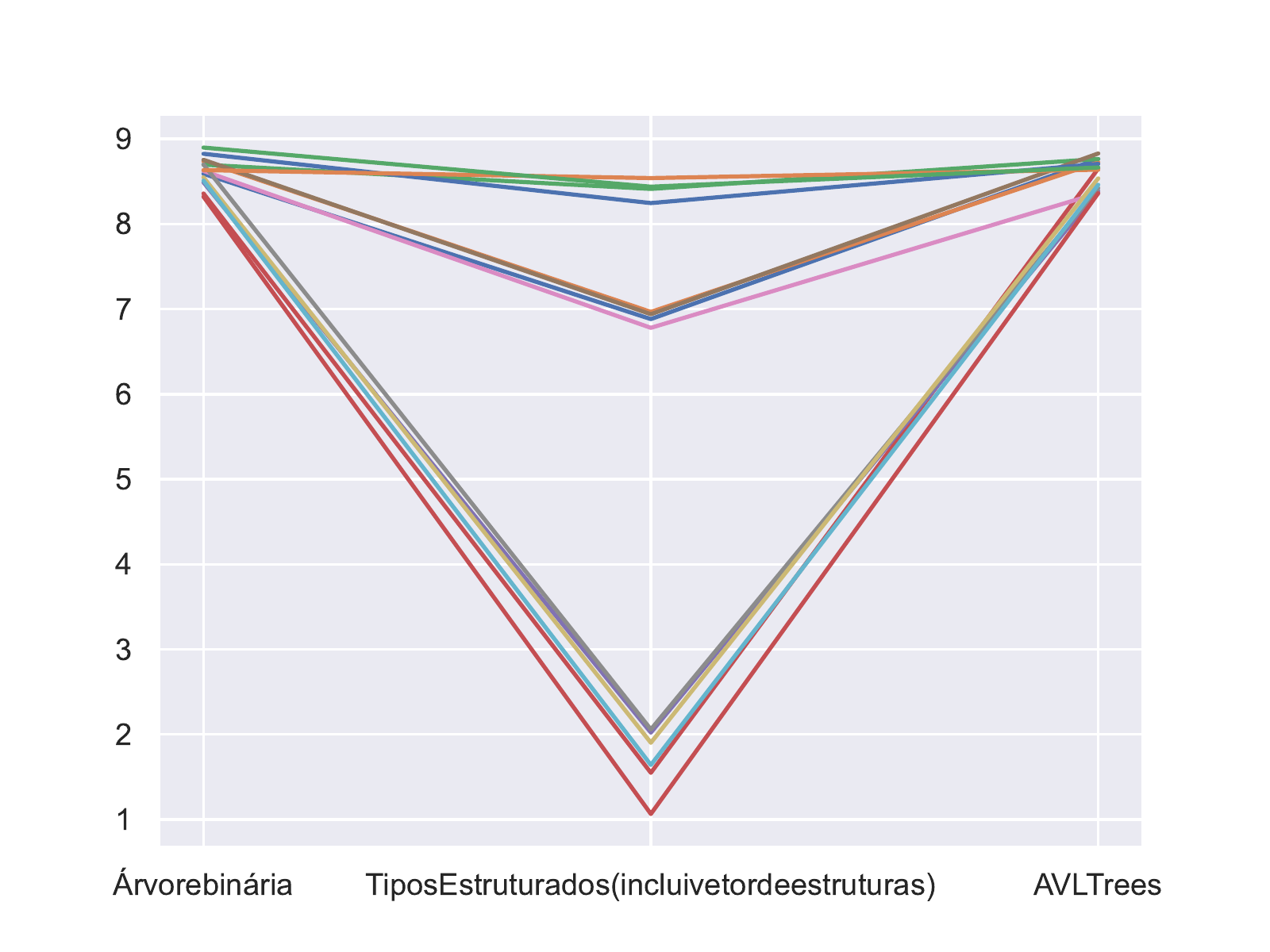}
	\includegraphics[width=78pt]{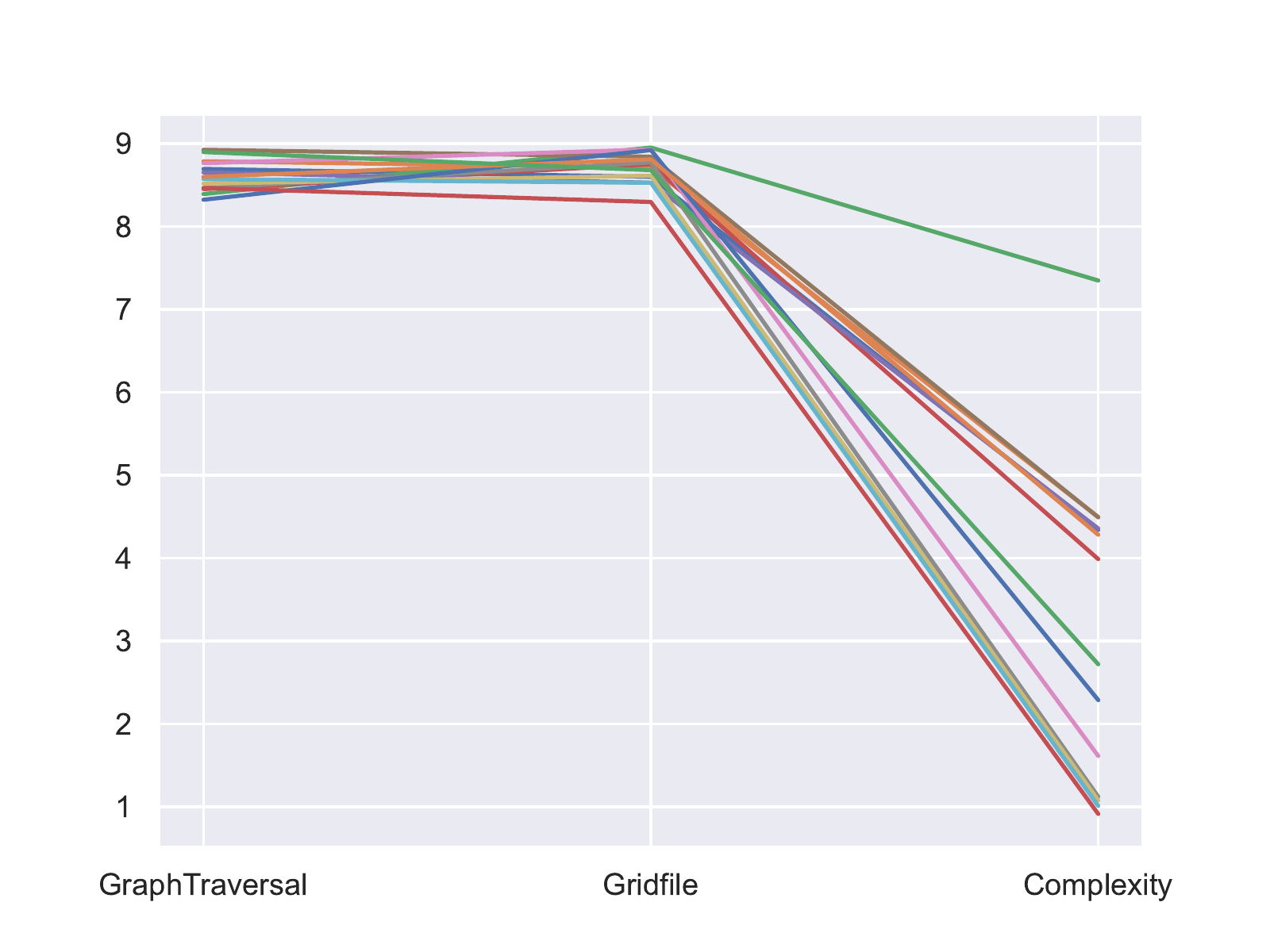}
	\caption{The patterns of four biclusters recovered in ADS dataset}
	\label{fig8}
\end{figure}

\begin{figure}[htbp]
	\centering
	\includegraphics[width=78pt]{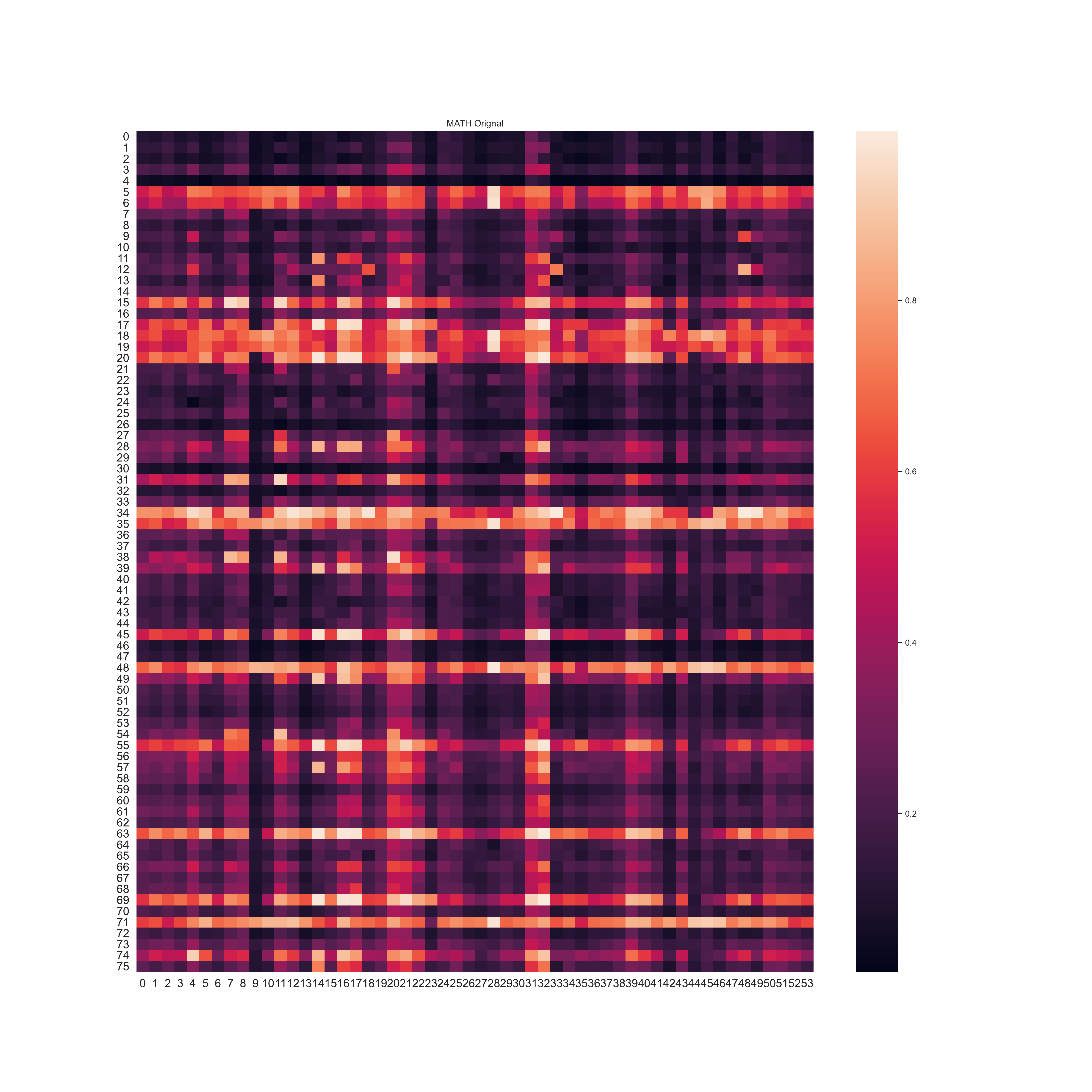}
	\includegraphics[width=78pt]{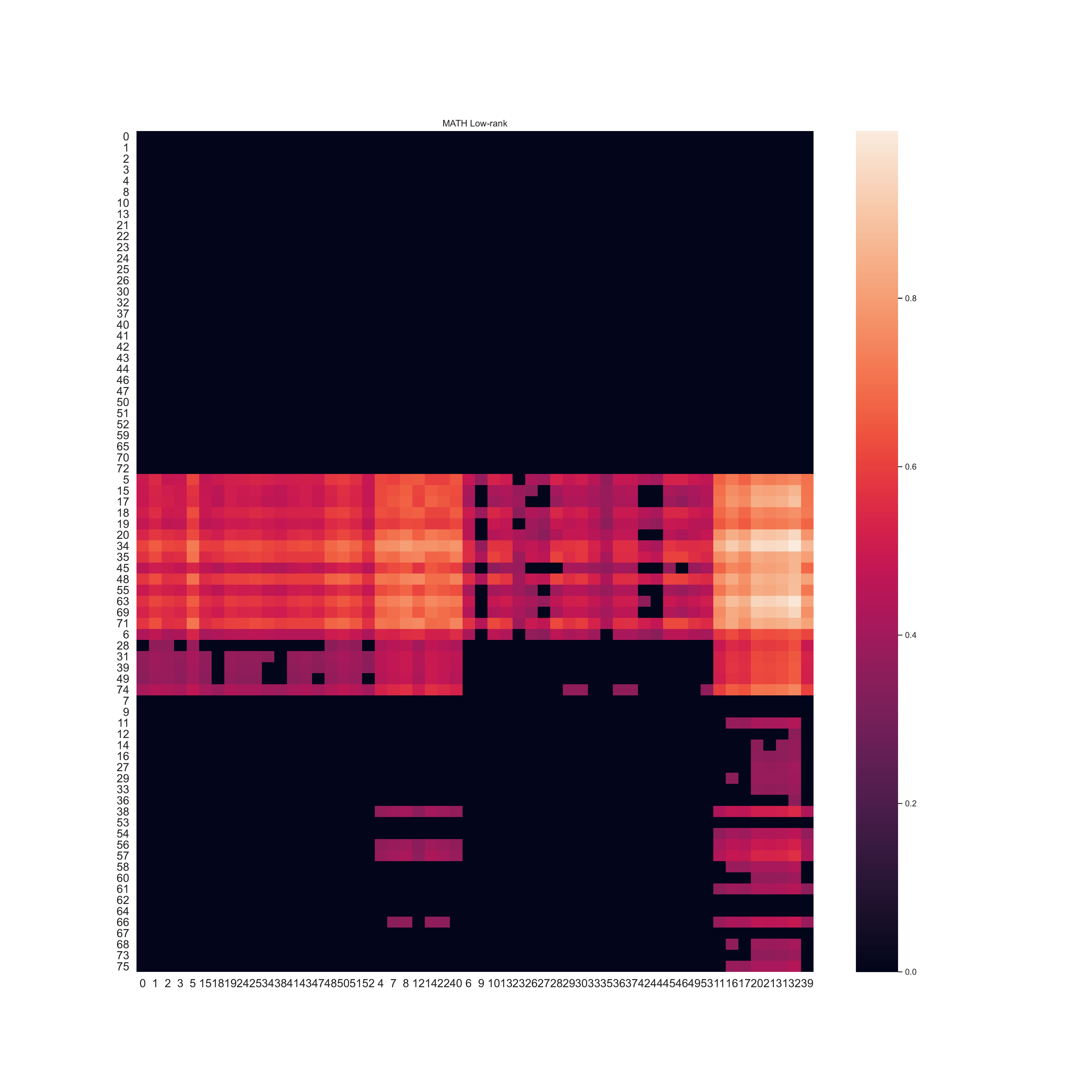}
	\includegraphics[width=78pt]{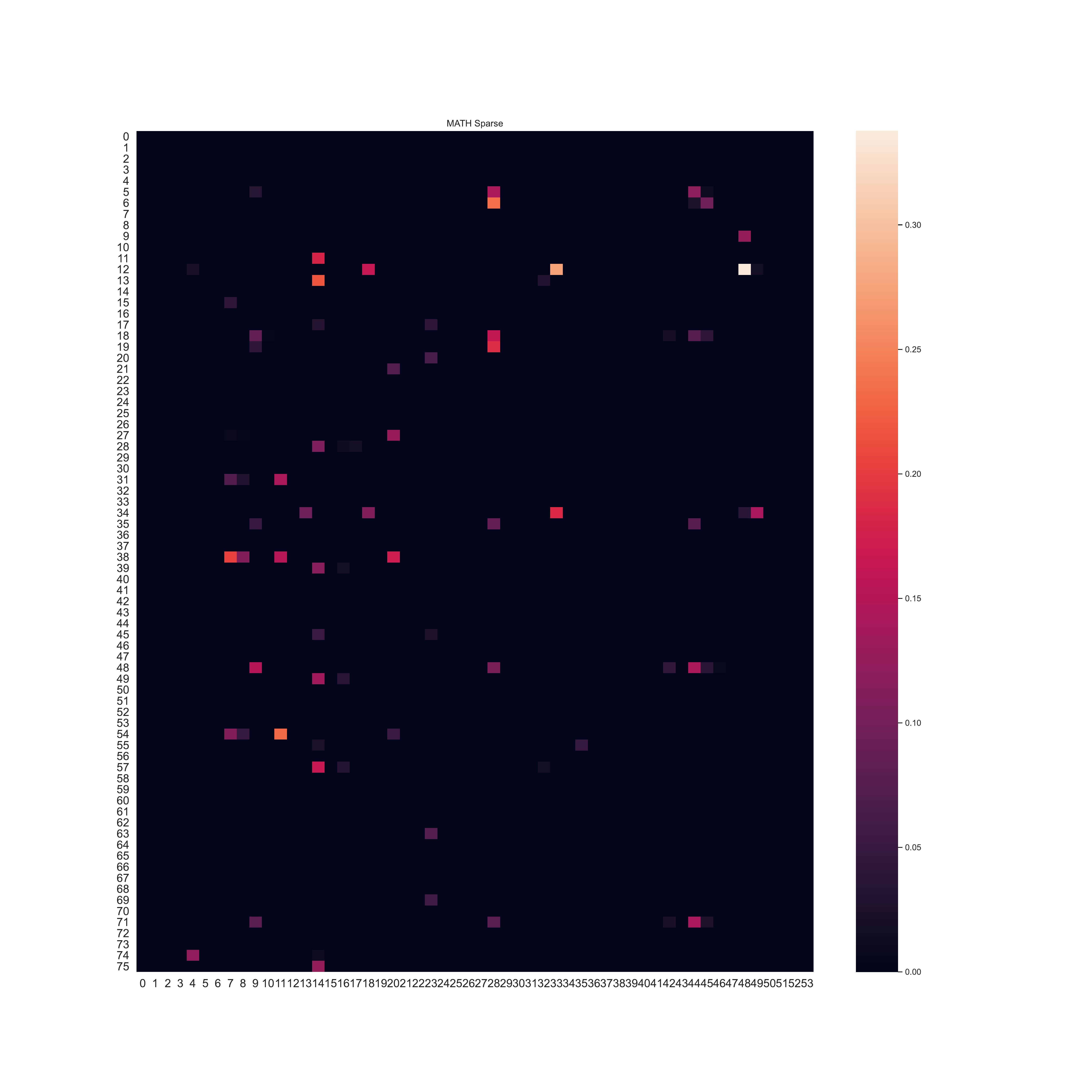}
	\caption{The results of MATH data experiments.}
	\label{fig7}
\end{figure}

The results of our method on ADS dataset is shown in \refFig{fig6}. For the resulted sparse component, the sparse rate is approximately 0.05. For the low-rank part, there are approximately 20 students and 5 knowledge topics per bicluster on average. We can find that the framework proposed in this article can efﬁciently and comprehensively detect student-topic biclusters. Table 2 gives the $P-$values of some biclusters found in ADS dataset by our method. The filter step of the framework ensures the statistic significant of the biclusters detected. \refFig{fig8} presents the patterns of four biclusters recovered in the low-rank matrix. Here, four patterns are divided into 2 knowledge topic sets: \{GraphTraversal, Gridfile, Complexity\} and \{Árvorebinária, TiposEstruturados(incluivetordeestruturas), AVLTrees\}. We can find that different biclusters follow different patterns. For the left two biclusters, students in the upper one maybe have difficulties in learning these three topics, while students in the lower one may be not good at \{Árvorebinária, AVLTrees\}, but for TiposEstruturados(incluivetordeestruturas), they are not the same as students in the upper one.
This phenomenon is more obvious in the right two biclusters. Thus, based on the different patterns, instructors can capture students' weak knowledge and give specific interventions to each group. We can get similar results from MATH dataset, as the patterns shown in \refFig{fig7}. 

\begin{figure}[htbp]
	\centering
	\includegraphics[width=200pt]{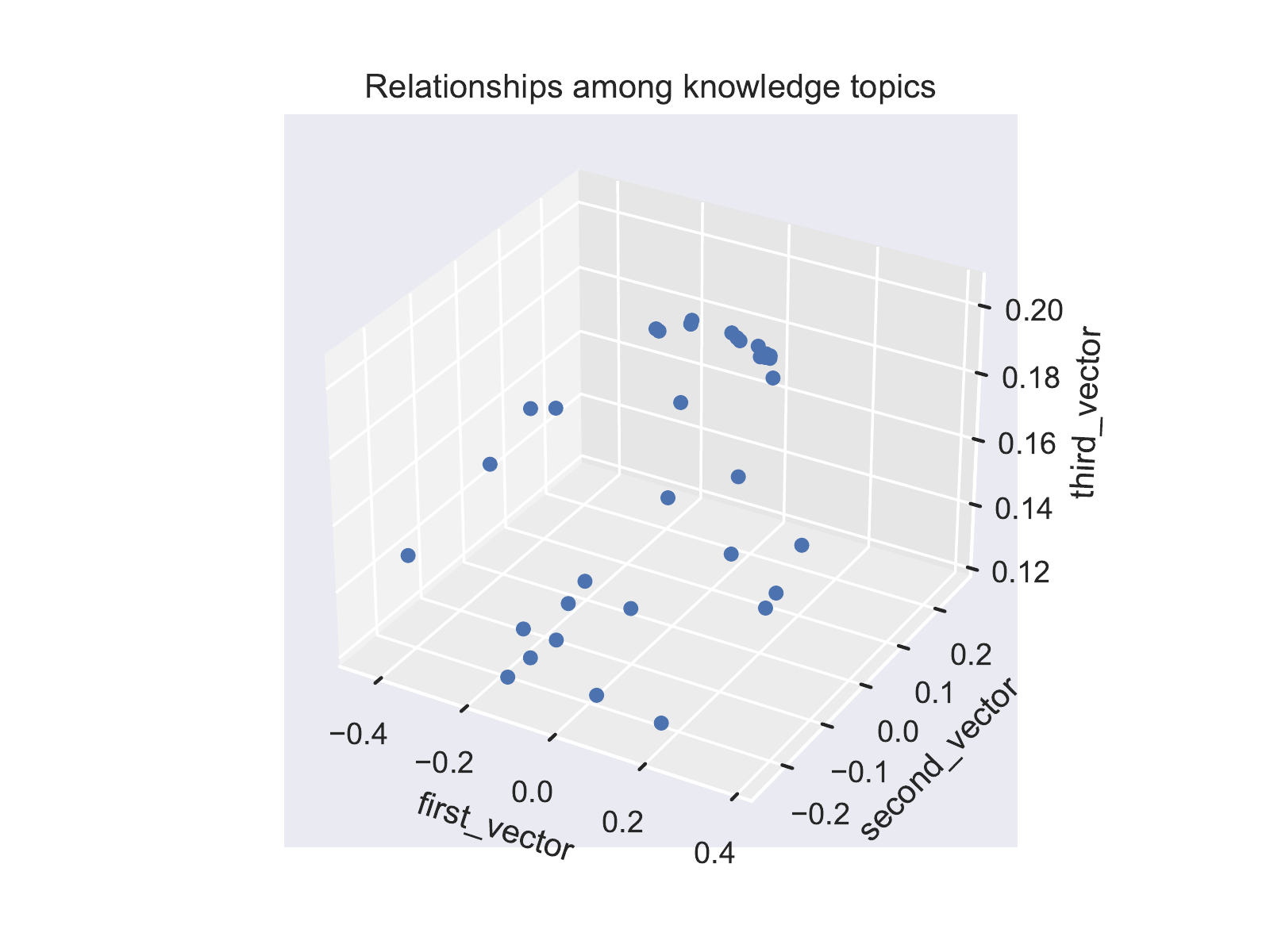}
	\caption{Analysis on the first three of the right singular vectors. We can regard the elements in the first three vectors as the important features of the corresponding topics. Clearly, some topics are grouped together based on these features.}
	\label{fig9}
\end{figure}

Furture more, we do some analysis on the recovered low-rank matrix. Inspired by Latent Semantic Analysis, we do Singular Value Decomposition on the recovered low-rank matrix. The right singular vectors can be regard as latent features for each topic. As shown in \refFig{fig9},  it is clear to see that some knowledge topics are clustered based on these features. This is in accord with intuition and practical. First, some topics may have the similar difficulties so that most students cannot master them easily. Secondly, for students, to master some knowledge topcis is highly relied on the mastery of same prior knowledge. Low proficiency on prior knowledge results in the fact that the following topics show a similar degree of mastery. This result gives us a new perspective to find the relationships among topics and enlightens us that well-defined relationships among knowledge topics could be useful in the detection of meaningful groups.

\begin{table}
	\centering
	\caption[\textbf{Some biclusters found in ADS dataset.} ]{$P-$values of some biclusters found in ADS dataset.}
	\begin{tabular}{p{2.5cm}|p{3cm}|p{2.2cm}}
		\hline
		%\multirow{1}{}{}
		\textbf{Students' ID} &\textbf{Knowledge Topics} &\textbf{P-value} \\ \hline
		16 students: [34  74  82  ... 168 174 175]  & ListasEncadeadas, Bitvector, GenericTrees, B-TreesInsertion & 7.1E-28$|$3.8E-5\\ \hline
		34 students: [12  20  31  ... 251 254 261]  & Árvorebinária, TiposEstruturados, AVLTrees & 4.8E-5$|$4.0E-4\\ \hline
		14 students: [3  19  30 ... 117 124 144]  & GraphTraversal, Gridfile, Complexity & 1.1E-13$|$2.8E-5\\ \hline
		12 students: [9  86 128 ... 244 245 281]  & ListasEncadeadas, Bitvector, GenericTrees, B-TreesInsertion & 1.6E-22$|$3.8E-5\\ \hline
	\end{tabular}
\end{table}

\section{Conclusion}
Balancing group teaching and individual mentoring is an important topic in education area. In this article, we propose a matrix recovery based method for detecting common characteristics for a group of students and identifying individual ones for each student. In addition, statistical evaluation method is applied to filter the spurious biclusters. The experiment results show that the method can provide more stable biclusters and accurate sparse signals. As shown in our experimental results, some relationships among knowledge topics are detected. Inspired by this, we can integrate information from knowledge graph to identify more meaningful groups in the future work.

%\section{Appendices}
%\subsection{Evaluation Metrics}
%\begin{enumerate}
%	\item {\textbf{liu wang match score.}} Factor analysis for bicluster acquisition models the data matrix as the sum of some biclusters plus additive noise, where each bicluster is the outer product of two sparse vectors. 
%	\item {\textbf{prelic recover score.}} Large Average Submatrices is an iterative procedure that searches for submatrices in data by locally maximizing a Bonferroni based significance score.
%	\item {\textbf{prelic relevance score.}} Iterative Signature Algorithm starts with randomly selected samples and features, evaluating and updating them through iterative steps until convergence.
%	\item {\textbf{csi.}} Spectral biclustering method uses singular value decomposition to find a checkerboard pattern in the data in which each bicluster is up- or downregulated.
%	\item {\textbf{cluster error.}} Spectral biclustering method uses singular value decomposition to find a checkerboard pattern in the data in which each bicluster is up- or downregulated.
%	\item {\textbf{fabia consensus score.}} Spectral biclustering method uses singular value decomposition to find a checkerboard pattern in the data in which each bicluster is up- or downregulated.
%\end{enumerate}

\bibliography{sample-base}
\bibliographystyle{aaai21}

\end{document}